\documentclass{magicstyle/magic}

\usepackage{amsmath,amssymb,amsfonts}
\usepackage{xspace}
\usepackage{float}
\usepackage{enumitem}
\usepackage{makecell}
\usepackage{pifont}
\usepackage{todonotes}
\setlist[itemize]{leftmargin=*,itemsep=0em,parsep=0.3em,topsep=0.3em}

\newcommand{\github}{\raisebox{-1.5pt}{\includegraphics[height=1.05em]{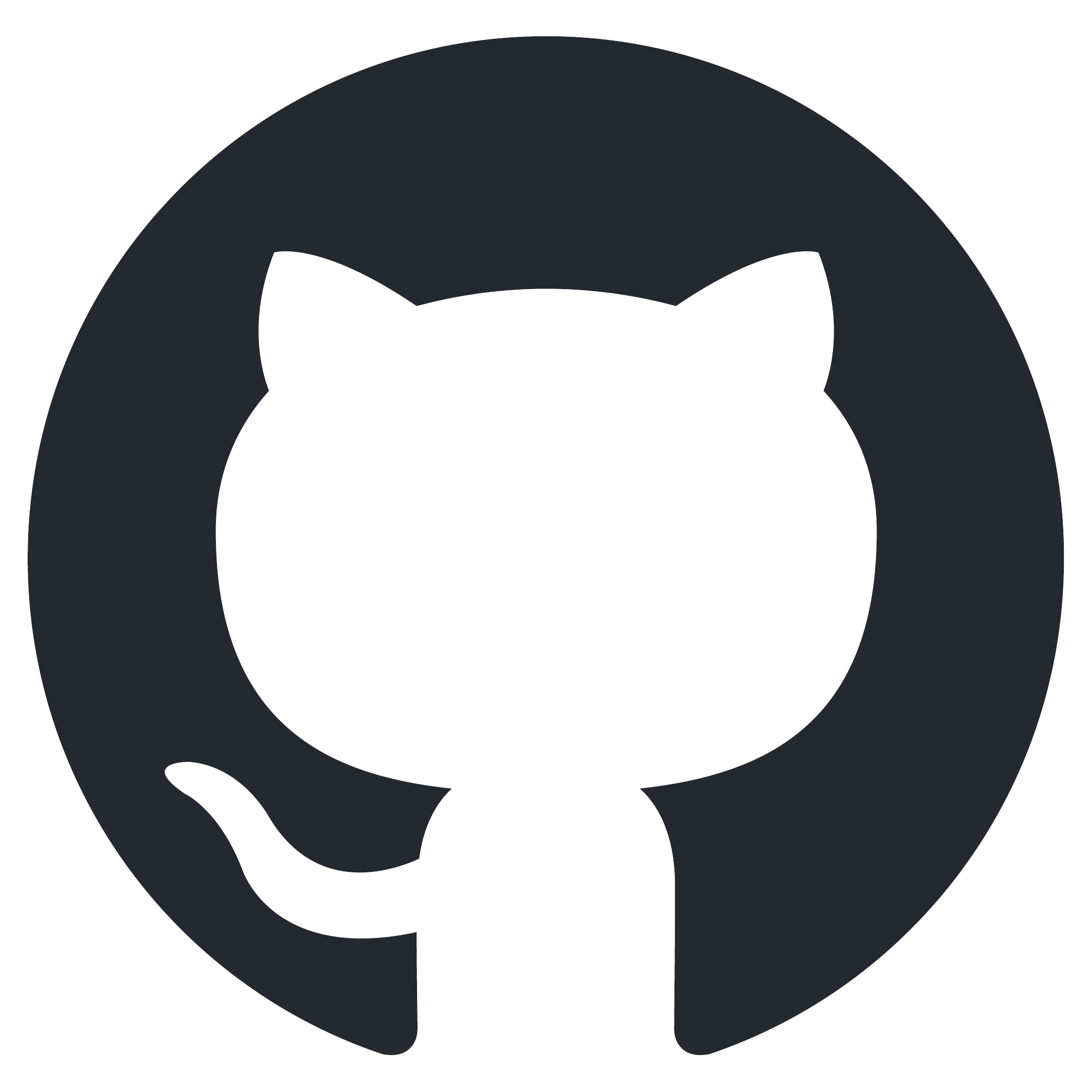}}\xspace}
\newcommand{\huggingface}{\raisebox{-1.5pt}{\includegraphics[height=1.05em]{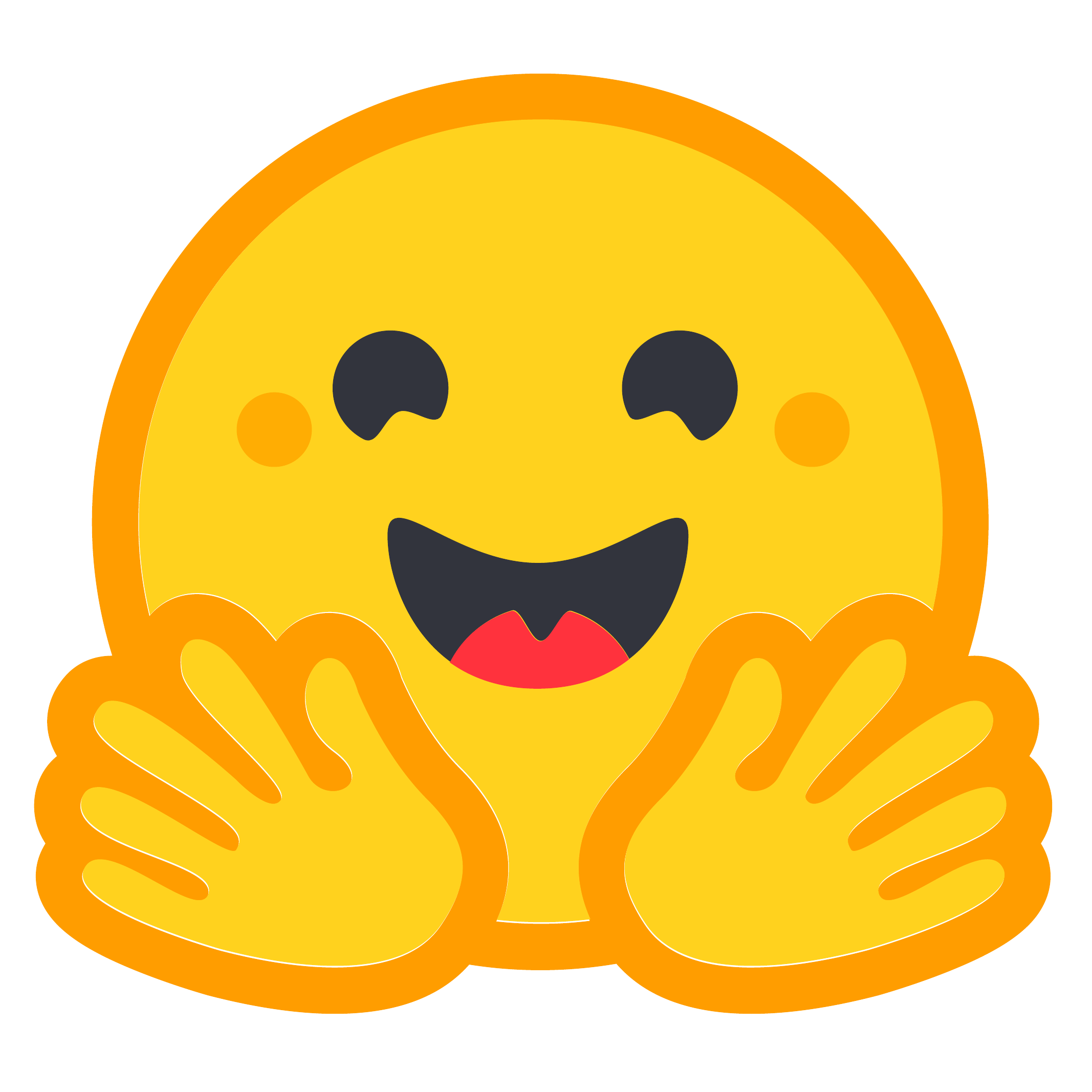}}\xspace}

\newcommand{\emailLogo}{\raisebox{-1.5pt}{\includegraphics[height=1.05em]{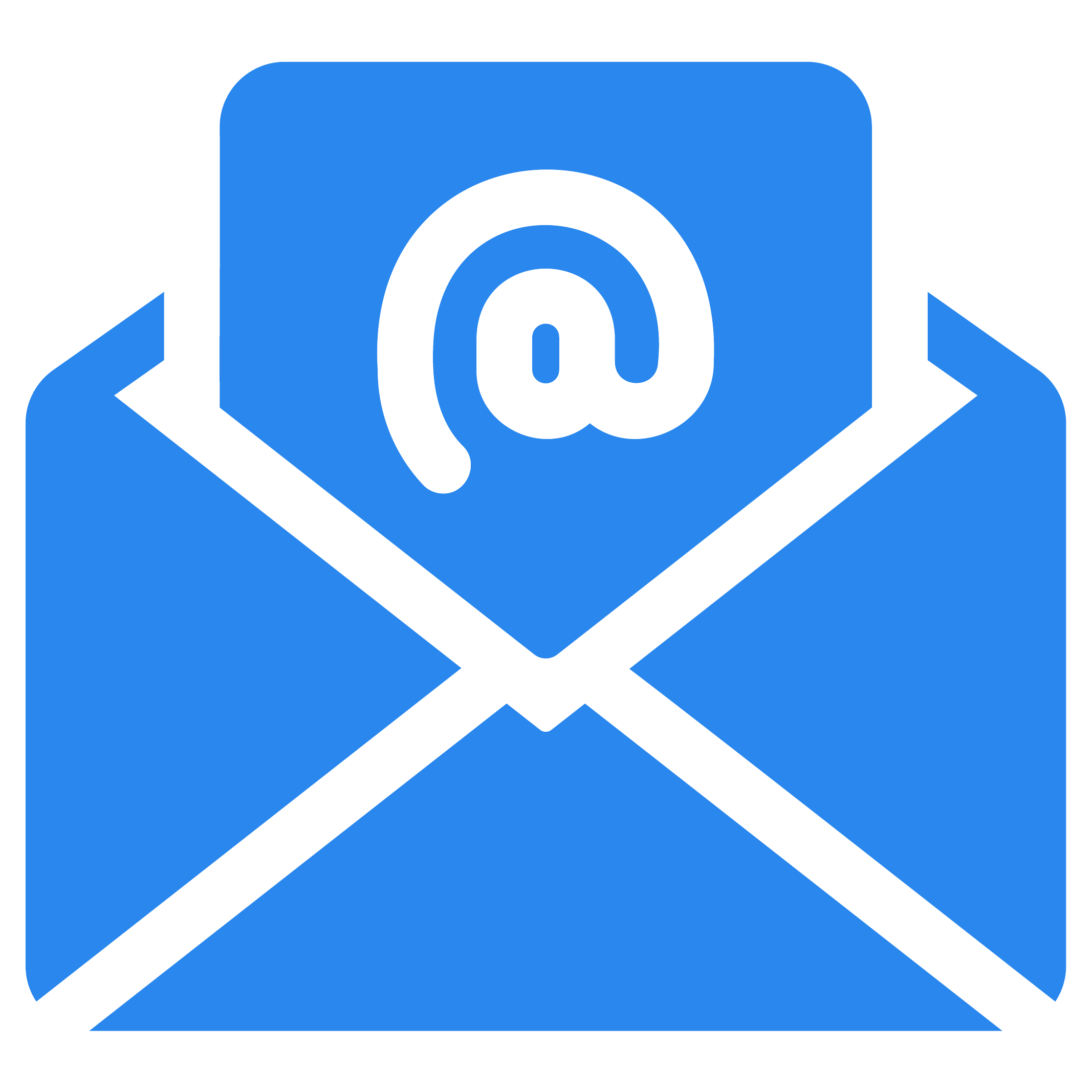}}\xspace}
\newcommand{\websiteLogo}{\raisebox{-1.5pt}{\includegraphics[height=1.05em]{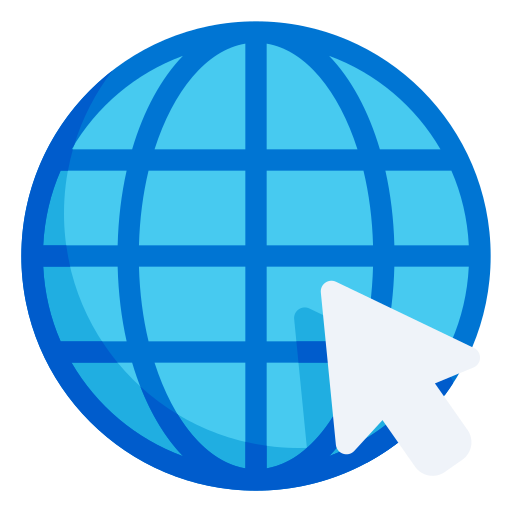}}\xspace}

\newtcolorbox{takeaway}[1][Takeaway]{enhanced, breakable, frame hidden,
    colback=magicmist, arc=3.5pt, left=0.35cm, right=0.35cm, top=0.25cm, bottom=0.25cm,
    before skip=0.4cm, after skip=0.4cm,
    borderline west={2.5pt}{0pt}{magicblue},
    fontupper=\small,
    before upper={{\sffamily\bfseries\color{magicnavy} #1.}\enskip}}

\definecolor{RTBlue}{HTML}{1c6ef3}
\definecolor{RTBlueLight}{HTML}{d2e3fc}

\newcommand{\rtbest}[1]{\textbf{#1}}
\newcommand{\rtsecond}[1]{\underline{#1}}

\title{\textcolor{magicblue}{RoboColosseum}\\{\fontsize{17pt}{12pt}\selectfont \textcolor{magicblue}{Evaluating Physical Intelligence at Scale In-The-Wild}}}

\author[1]{Tan Sang Nguyen\textcolor{magicblue}{\ding{170}}$^{\dagger}$}
\author[2]{Yi Ru Wang\textcolor{magicblue}{\ding{170}}$^{\dagger}$}
\author[1]{Shailesh Shailesh\textcolor{magicblue}{\ding{170}}$^{\dagger}$}
\author[1]{Gang Yang\textcolor{magicblue}{\ding{170}}}
\author[1]{Jiamo Hu\textcolor{magicblue}{\ding{170}}}
\author[1]{Jianuo Jiang\textcolor{magicblue}{\ding{170}}}
\author[3]{Haoquan Fang\textcolor{magicblue}{\ding{170}}}
\author[4,5]{Zhongzheng Ren\textcolor{magicblue}{\ding{170}}}
\author[1]{Zhongyi Luo\textcolor{magicblue}{\ding{170}}}
\author[8]{Ian Chong\textcolor{magicblue}{\ding{170}}} 
\author[8]{Aaron Tung\textcolor{magicblue}{\ding{170}}}
\author[8]{Santiago Pravisani\textcolor{magicblue}{\ding{170}}}
\author[1]{Minghui Lin}
\author[8]{Yao Mo}
\author[6]{Ishneet Singh}
\author[1]{Benjamin Tee}
\author[1]{Harold Soh}
\author[2]{Siddhartha Srinivasa}
\author[2,5]{Dieter Fox}
\author[1]{David Hsu}
\author[7]{Haoming Song}
\author[7]{Maxime Ellerbach}
\author[7]{Thomas Wolf}
\author[7]{Pepijn Kooijmans\textcolor{magicblue}{\ding{170}}}
\author[1,9]{Ruijie He\textcolor{magicblue}{\ding{170}}}
\author[8]{Michael Cho\textcolor{magicblue}{\ding{170}}}
\author[1,9]{Jiafei Duan\textcolor{magicblue}{\ding{170}}*}
\affiliation[1]{National University of Singapore}
\affiliation[2]{University of Washington}
\affiliation[3]{Stanford University}
\affiliation[4]{University of North Carolina at Chapel Hill}
\affiliation[5]{Allen Institute for AI}
\affiliation[6]{University of Oxford}
\affiliation[7]{Hugging Face}
\affiliation[8]{BitRobot}
\affiliation[9]{Singapore Embodied AI Centre}

\contribution[$^{\dagger}$]{Co-first}
\contribution[\textcolor{magicblue}{\ding{170}}]{Core Contributor}
\contribution[*]{Corresponding Author}

\abstract{General-purpose robot policies must work across robots, tasks, and environments, yet the variation that makes physical trials informative also makes them difficult to reproduce. We introduce \textsc{RoboColosseum}, a distributed evaluation framework that combines unified policy serving and standardized protocols across a network of locally operated robots in diverse environments. Our initial study evaluates five vision-language-action (VLA) policies and three frontier AI agents across four embodiments (SO-101, Franka, Bimanual-YAM, and G1 Humanoid). The \texttt{Open Track} compares models without task-specific adaptation in blinded, pairwise trials under matched local conditions. The \texttt{Fine-tune Track} measures how well VLA policies adapt to 20 tasks when each policy receives the same 100 demonstrations per task. Across 5,699 Open Track comparisons at 124 sites, frontier agents rival specialized VLA policies, though all systems struggle to complete open-world tasks reliably. After fine-tuning, GR00T N1.7 leads on SO-101 and $\pi_{0.5}$ leads on Franka and Bimanual-YAM. Policy rankings show weak agreement across embodiment-specific task suites (Kendall's $W=0.24$). Open Track rankings poorly predict relative performance after fine-tuning, with substantial rank reversals on all three embodiments shared between tracks. Measuring partial progress reveals differences that task success alone misses. On the G1 Humanoid, no policy completes any of the five tasks, yet average progress ranges from 13.3\% to 23.2\% across policies. Open Track rankings stabilize as trials accumulate, and active matchmaking achieves comparable ranking agreement with substantially fewer physical trials than uniform sampling. These results show how distributed physical evaluation can reveal where robot policies generalize, how well they adapt to new tasks, and how far they progress on tasks they fail to complete.}

\metadata[\vspace{.3em}\websiteLogo Webpage:]{\href{https://robocolosseum.ai/}{\texttt{robocolosseum.ai}}}
\metadata[\vspace{.3em}\github Code:]{\href{https://github.com/RoboColosseum}{\texttt{github.com/RoboColosseum}}}
\metadata[\vspace{.3em}\huggingface Models \& Datasets:]{\href{https://huggingface.co/RoboColosseum}{\texttt{huggingface.co/RoboColosseum}}}
\metadata[\vspace{.3em}\emailLogo Contact:]{\email{duanj1@nus.edu.sg}}

\begin{document}

\maketitle

\section{Introduction}
\label{sec:intro}

\begin{figure*}[t]
\centering
\includegraphics[width=\textwidth]{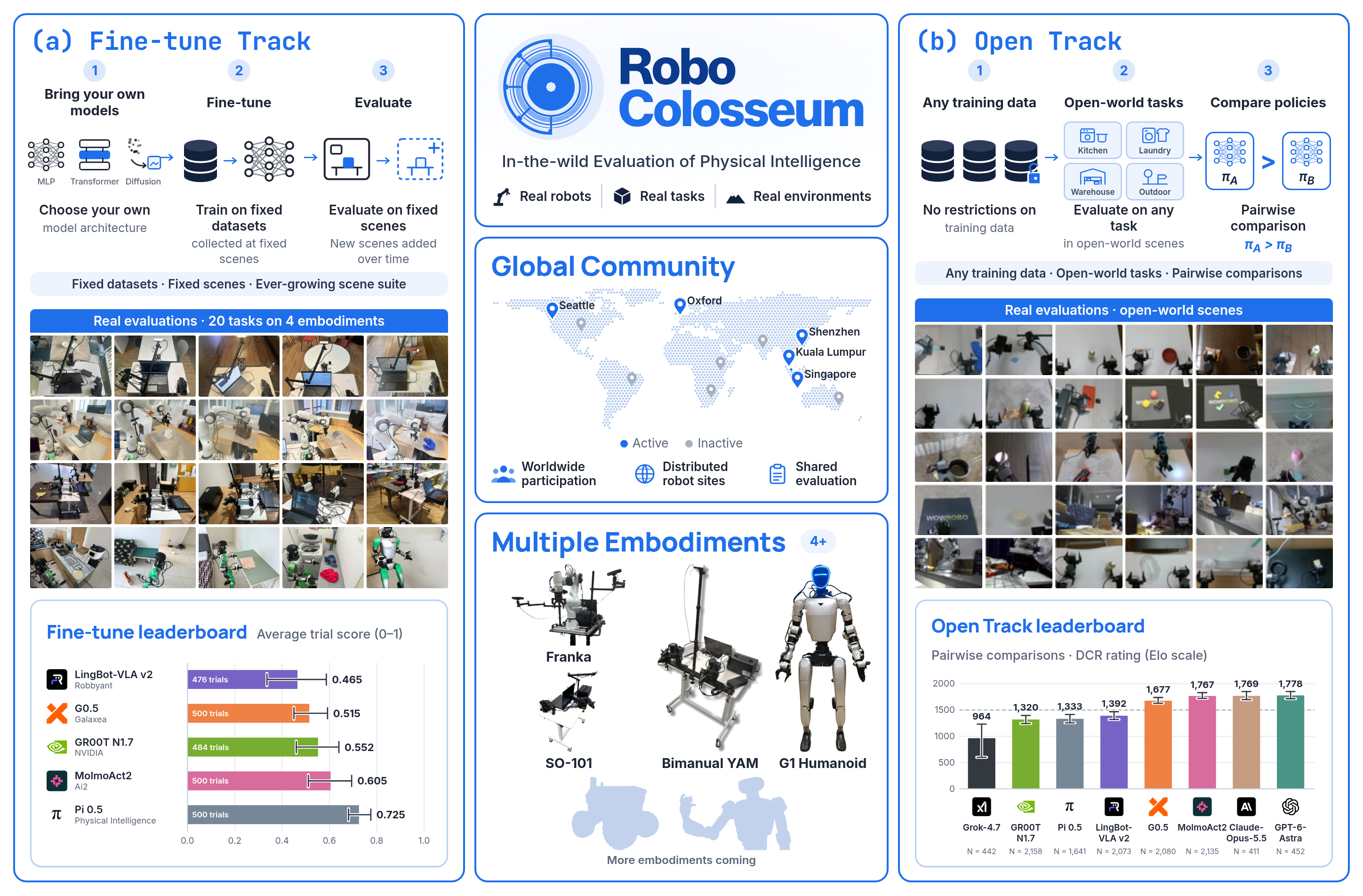}
\caption{\textbf{RoboColosseum overview.}
RoboColosseum connects distributed evaluation sites through shared robot interfaces, policy-serving infrastructure, and two complementary evaluation tracks.
In the \textbf{Fine-tune Track}, policies receive the same 100 demonstrations per task and are evaluated on predefined tasks using common success and progress criteria.
In the \textbf{Open Track}, policies are evaluated without additional task-specific adaptation through blinded, pairwise comparisons under matched local conditions, while tasks and environments vary across comparisons. The results shown for both tracks are based on evaluations we conducted from 17 September to 28 October 2026. The project spans SO-101, Franka in the DROID configuration, Bimanual-YAM, and Unitree G1, with participation determined by track and model compatibility.
The lower panels illustrate track-specific evaluation outputs and examples of physical evaluation settings.
The global network depicts the broader project vision; the scope of completed experiments is reported separately.}
\label{fig:framework}
\vspace{-2pt}
\end{figure*}

\begin{quote}
   \textit{You don't know what you don't know.}\\[2pt]
  --- Socrates, \emph{Philosopher}
\end{quote}

Recent advances in robot learning are shifting the field from policies tailored to individual tasks and platforms toward general-purpose models~\citep{fang2026molmoact2,team2025gemini,intelligence2025pi05visionlanguageactionmodelopenworld,gr00tn1_2025,lee2025molmoactactionreasoningmodels} trained across diverse tasks, environments, and embodiments~\citep{10611477,khazatsky2024droid}. Although compelling demonstrations from leading robotics companies highlight the potential of these models~\citep{dyna2026dyna2,sunday2025act1,skild2025roboticbrain,generalist2026gen1}, establishing their reliability under real-world deployment conditions remains a fundamental challenge. In language and vision, large-scale benchmarks incorporating human preferences on real-world tasks have helped drive progress by providing a shared basis for assessing model capabilities relevant to end users~\citep{zheng2023judging,chiang2024chatbot,huang2024vbench}. Robot policies, by contrast, operate through closed-loop interactions with the physical world, where each action shapes the conditions for what follows. Evaluating them reliably is therefore difficult, especially at scale and in the varied, uncontrolled settings of the real world. Measuring progress requires understanding what robots can consistently accomplish across these conditions.

Such assessments are difficult to standardize because outcomes depend jointly on the policy, robot setup, initial conditions, and surrounding environment~\citep{vincent2024generalizable}. Even nominally identical tasks can differ in object placement, camera configuration, lighting, and physical properties, making it difficult to attribute performance differences to policy capability. Physical evaluation must therefore reconcile two requirements: sufficient control to support meaningful comparisons and sufficient environmental diversity to assess robustness beyond a particular setup. Simulation supports scalable evaluation through repeatable resets and controlled variation in tasks and environments~\citep{9001253,tao2025maniskill3}. However, differences in sensing, calibration, dynamics, and visual appearance can limit how well simulated performance predicts real-world outcomes, making physical validation essential~\citep{li24simpler,pumacay2024colosseum,kim2026molmospaces}. Achieving comparable scale with physical robots requires substantial labor for setup, resets, hardware maintenance, and trial execution. Consequently, real-world studies often rely on relatively few rollouts, leaving performance estimates sensitive to trial selection and experimental variability~\citep{kress2024robot,vincent2024generalizable}.

Distributing evaluation across sites broadens environmental coverage~\citep{khazatsky2024droid,atreya2025roboarena}, but efforts that depend on students and volunteers are difficult to sustain and scale. Running physical trials requires continued staffing, equipment maintenance, and coordination across sites, often alongside competing research commitments. Sustained evaluation therefore requires dedicated operators and shared physical infrastructure, supported by protocols for repeating trials, matching local conditions, and verifying whether measured performance can be reproduced. 


We introduce \textsc{RoboColosseum}, a distributed, multi-embodiment benchmark for evaluating general-purpose robot policies in the wild. It connects geographically distributed sites through unified evaluation and policy-serving infrastructure, with standardized hardware configurations, task definitions, and evaluation protocols. The benchmark initially supports four embodiments Bimanual-YAM, SO-101, Franka, and Unitree G1 and evaluates eight foundation models: five vision-language-action (VLA) policies and three frontier LLMs.

Two complementary tracks assess adaptation and generalization. The \texttt{Fine-tune Track} measures performance on predefined semantic tasks after adaptation using a fixed budget of task-specific demonstrations. The \texttt{Open Track} evaluates zero-shot generalization across diverse real-world tasks and environments without task-specific adaptation. Together, they enable systematic assessment of physical intelligence across heterogeneous embodiments. With a growing fleet of over 100 robots across the United States, United Kingdom, Singapore, Malaysia, and China, operated by full-time evaluators, \textsc{RoboColosseum} aims to make real-world evaluation scalable and sustainable. This shared infrastructure provides a lens into the capabilities and limitations of robot foundation models, enabling the research community to investigate fundamental scientific questions through systematic physical trials. To broaden access, we pledge 10,000 hours of free evaluation to academic labs and partners. 

Our initial study includes 5,699 \texttt{Open Track} pairwise comparisons across 124 sites and \texttt{Fine-tune Track} evaluations on 20 tasks across four embodiments. Frontier AI agents achieve manipulation performance competitive with specialized VLA policies, while reliable task completion remains challenging as task difficulty increases. With the same 100-demonstration adaptation budget, policies exhibit strongly task- and embodiment-dependent performance, with weak agreement in rankings across embodiments ($W=0.24$) and substantial rank reversals between the Open and Fine-tune Tracks. Finally, leaderboard stability improves with evaluation scale, while active matchmaking reaches approximately 90\% agreement with the full-data ranking after 400--500 evaluations, compared with roughly 1,000 under uniform sampling. These findings demonstrate the importance of evaluating physical intelligence across diverse tasks, embodiments, and environments, while showing that large-scale real-world evaluation can be made statistically reliable and sample-efficient.

\section{Related Work}
\label{sec:related}

\subsection{Simulated and Real-World Benchmarking}
\label{sec:robot_benchmarks}

Real-world robot benchmarks aim to make physical policy evaluation more reproducible despite variation in hardware, calibration, object placement, and scene reset. These factors can substantially affect measured performance, motivating standardized hardware, task definitions, initial conditions, and success criteria~\citep{kress2024robot}. Early benchmarks such as FurnitureBench improve comparability by prescribing a common robot platform, workspace, objects, and task protocols~\citep{furniturebench2025}, while SceneReplica further studies reproducibility by reconstructing physical scenes using reference images and calibrated setups~\citep{SceneReplica2024}. Remote testbeds such as TOTO and the Real Robot Challenge instead centralize physical hardware and allow researchers to submit policies for evaluation on a shared robot platform, reducing variation caused by independently constructed setups~\citep{toto2023, pmlr-v176-bauer22a}.

More recent benchmarks extend these ideas toward broader policy and embodiment coverage~\citep{pumacay2024colosseum,wang2025roboeval,kim2026molmospaces,wang2026roboplaygrounddemocratizingroboticevaluation}. RoboDojo~\citep{chen2026robodojo} evaluates 18 real-world tasks across three bimanual robot platforms using standardized workspaces, reset procedures, and sub-step scoring, while VLA-REPLICA provides a low-cost SO-101 setup designed to be reconstructed at different locations~\citep{chen2026robodojo, huang2026vlareplica}. \texttt{RoboColosseum} builds on this line of work while targeting broader physical coverage across multiple embodiments and sites under shared evaluation protocols.

\subsection{Generalist Robot Policies}
\label{sec:rfm}

Generalist robot policies seek to transfer learned skills across tasks, environments, and embodiments. RT-X models demonstrate that pooling demonstrations from different robots can improve performance through transfer across platforms~\citep{10611477}. Octo provides a pretrained policy that can be adapted to new observation and action spaces~\citep{octomodel2024}. Vision-language-action models such as OpenVLA combine pretrained vision-language representations with robot demonstrations to predict actions from visual observations and language instructions, while supporting fine-tuning on demonstrations from a target setting~\citep{kim2024openvla}.

Recent models extend these capabilities through diverse training data and different approaches to action generation. $\pi_{0.5}$ combines data from multiple robots with web data and semantic supervision to support manipulation in previously unseen homes~\citep{black25p05}. GR00T combines vision-language processing with a dedicated action-generation module~\citep{bjorck2025gr00t}, while MolmoAct2 connects a backbone trained for embodied reasoning to a continuous-action expert~\citep{fang2026molmoact2}. These models provide different approaches to combining semantic knowledge with physical control.

General-purpose language models can also guide robot behavior through skill selection or program generation. SayCan selects learned robot skills using both language-model predictions and estimates of execution success~\citep{ahn2022can}, while Code as Policies generates programs that process perception outputs and call robot control APIs~\citep{liang2022code}. \texttt{RoboColosseum} compares VLA policies and frontier AI agents across physical environments and embodiments without task-specific adaptation, and separately measures how VLA policies adapt using a common budget of task demonstrations.

\subsection{Distributed Robotics Evaluation}
\label{sec:arena_style_eval}
Distributed evaluation provides an alternative to tightly centralized robot benchmarks by allowing policies to be tested across heterogeneous physical environments while aggregating results into a common comparison. RB2 is an early example, distributing manipulation evaluation across multiple laboratories while controlling task definitions and aggregating site-specific rankings rather than directly comparing raw scores across different physical setups~\citep{dasari2021rb}. More recently, RoboArena extends this idea to generalist robot policies through a decentralized evaluation network spanning seven academic institutions. Evaluators freely select tasks and environments, while each local comparison evaluates two anonymously assigned policies back-to-back under matched scene conditions. Across 612 pairwise comparisons, RoboArena showed that rankings derived from distributed evaluations correlate more strongly with an exhaustive oracle ranking than conventional fixed single-site evaluation~\citep{atreya2025roboarena}. However, RoboArena is instantiated on a single DROID/Franka embodiment, and its open-ended decentralized design makes controlled single-factor experiments difficult.

Pairwise comparison has also been widely used for arena-style evaluation. Chatbot Arena ranks anonymized models from human pairwise preferences using Bradley--Terry-based estimation \citep{chiang2024chatbotarena}, while active-ranking methods show that comparisons can be allocated toward more informative model pairs rather than sampled uniformly \citep{heckel2018approximate}. These ideas provide methodological precedent for \texttt{RoboColosseum’s Open Track}, where pairwise physical evaluations are aggregated into a global ranking and costly robot trials can be prioritized toward uncertain model pairs.

\section{RoboColosseum Benchmark}
\label{sec:robocolosseum_benchmark}

\begin{figure*}[t]
\centering
\includegraphics[width=\textwidth]{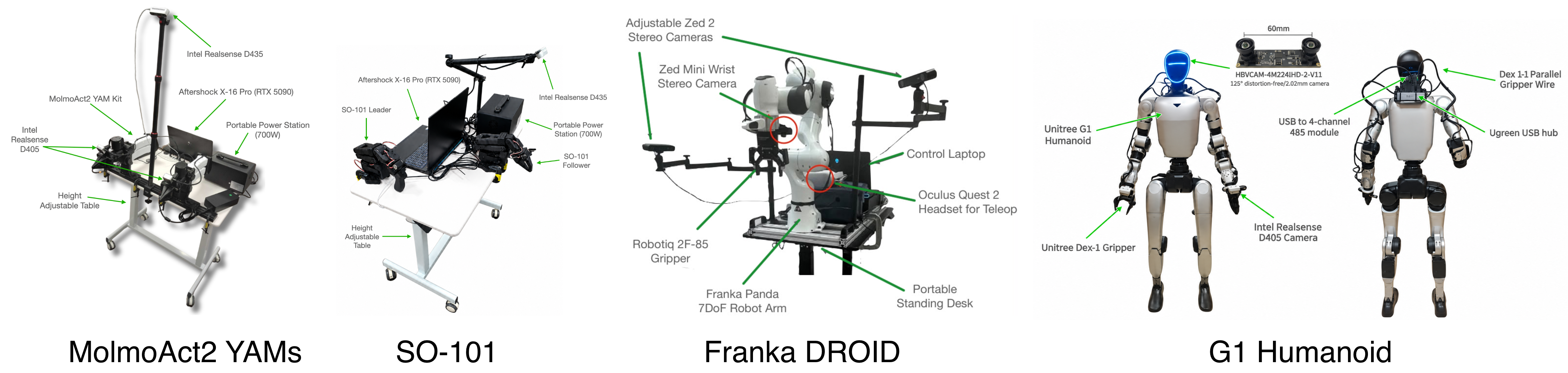}
\caption{\textbf{Robot platforms and hardware configurations.}
From left to right: Bimanual-YAM, SO-101, Franka in the DROID configuration, and Unitree G1, shown from the front and rear. Annotations identify cameras, grippers, computing equipment, and supporting hardware. These platforms span bimanual manipulation, low-cost and research-grade single-arm manipulation, and humanoid manipulation with locomotion. Observation and action specifications are standardized within each embodiment to support compatible policies across evaluation sites.}
\label{fig:robot_platforms}
\vspace{-2pt}
\end{figure*}

We present \texttt{RoboColosseum}, a distributed real-world benchmarking framework for systematically evaluating the generalization and adaptation capabilities of general-purpose robot policies across heterogeneous embodiments. A central contribution is the design of the physical infrastructure and operational protocols required to sustain large-scale, distributed robot evaluation. As illustrated in Figure~\ref{fig:framework}, the framework integrates geographically distributed robot evaluation sites, shared policy-serving infrastructure, and standardized evaluation protocols into a unified benchmarking system. \texttt{RoboColosseum} currently supports four robot embodiments and eight AI systems, comprising five vision-language-action (VLA) policies and three API-accessible AI agents. The benchmark comprises two complementary tracks: an \texttt{Open Track} for evaluation without task-specific adaptation and a \texttt{Fine-tune Track} for evaluation following controlled task-specific adaptation. Together, these tracks assess out-of-the-box capabilities and the ability to acquire new manipulation skills from demonstrations.

\subsection{Physical Infrastructure and Operational Protocols}
\label{sec:setup}
\texttt{RoboColosseum} currently supports four robot embodiments spanning diverse costs, morphologies, and control complexities: a low-cost single arm, a bimanual system, a research-grade arm, and a walking humanoid (\Cref{tab:embodiments}). Evaluating the same model families across these embodiments tests whether findings generalize across hardware platforms, a question single-embodiment benchmarks cannot address. To sustain reliable daily evaluation at scale, we establish standardized hardware configurations, setup procedures, and operational protocols across sites.

\begin{table}[t]
\centering
\caption{\textbf{Embodiments in \texttt{RoboColosseum}.} Each embodiment has a fixed observation and action specification shared by all policies and sites. \texttt{Open Track} checkpoints are trained on the listed embodiment dataset; \texttt{Fine-tune Track} checkpoints use only the task demonstrations (\Cref{sec:finetuning_track}). Action dimensions include one gripper dimension per arm: 5 arm joints for SO-101, 7 for the Franka, and 6 per arm for Bimanual-YAM.}
\label{tab:embodiments}
\small
\renewcommand{\arraystretch}{1.15}
\begin{tabular}{@{}llll@{}}
\toprule
\textbf{Embodiment} & \textbf{Action space} & \textbf{Cameras} & \textbf{Dataset} \\
\midrule
SO-101 & 6-D joint position & 1 external + 1 wrist & MolmoAct2-SO101~\citep{fang2026molmoact2} \\
DROID (Franka) & 8-D joint position & 2 external + 1 wrist & DROID dataset ~\citep{khazatsky2024droid} \\
Bimanual-YAM & 14-D joint position & 1 top + 2 wrist & MolmoAct2-YAM~\citep{fang2026molmoact2} \\
\bottomrule
\end{tabular}
\end{table}

\paragraph{Embodiments and Compute.} \texttt{RoboColosseum} supports SO-101, Bimanual YAM, Franka, and the G1 humanoid, with standardized hardware configurations adapted for deployment in unstructured environments. SO-101 uses a single external camera following common community configurations, while Bimanual YAM adopts the MolmoAct2 setup; both systems are mounted on mobile trolleys. The Franka platform follows the DROID~\citep{khazatsky2024droid} configuration, and G1 is equipped with parallel grippers, all of which are shown in Figure \ref{fig:robot_platforms} Policy inference runs either locally on a laptop equipped with an RTX 5090 GPU or remotely through a server-hosted endpoint in which both are operating by a remote routing server. We measure inference and communication latency to verify that both configurations support low-latency execution.

\paragraph{Sites and Operation}
Academic robot evaluation typically relies on controlled laboratory tasks and environments. \texttt{RoboColosseum} conducts all evaluations at real-world sites, including offices, libraries, parks, dormitories, homes, food courts, and factories. Our evaluation network spans campuses and industry buildings in Singapore, homes in Malaysia, and factories in China, supporting the evaluation of diverse embodiments and deployment conditions. To sustain this effort, we hire up to 30 full-time robot operators execute evaluation protocols, reset task environments, and assess policy rollouts, with staffing expanding alongside evaluation demand. A dedicated data-processing pipeline applies face blurring and identity anonymization to evaluation recordings to protect participant privacy.


\subsection{Benchmark Design}
\label{sec:benchmark_design}
\texttt{RoboColosseum} is designed to support evaluation in different real-world settings while keeping model comparisons fair. We do not require every site to use the same environment, objects, or background. Instead, the benchmark uses two evaluation tracks with different controls. In the \texttt{Fine-tune Track}, models are evaluated on predefined tasks using a common adaptation and evaluation protocol. In the \texttt{Open Track}, models are tested on diverse tasks and environments, but the two models in each comparison are evaluated under the same local conditions. 

For each track, we clearly define what is fixed and what can vary. In the \texttt{Fine-tune Track}, models receive the same task-specific demonstration data and are evaluated using the same task definition, success criteria, progress criteria, and evaluation procedure. The training method can differ across models because different policy architectures require different fine-tuning procedures. This follows the controlled benchmark design used in DataComp, where selected factors are fixed so that different submissions can be compared under the same evaluation setting~\citep{gadre2023datacompl}. In the \texttt{Open Track}, models cannot use task-specific adaptation, while the task and environment can vary across evaluations. Within each pairwise comparison, the robot, task instruction, objects, initial setup, and timeout are kept the same.

\subsubsection{Fine-tune Track}
\label{sec:finetuning_track}
The \texttt{Fine-tune Track} evaluates how efficiently pretrained policies adapt to new tasks through rapid fine-tuning on task-specific demonstrations. By holding the adaptation data and evaluation protocol fixed, this track compares models’ ability to learn from the same demonstrations under consistent conditions, enabling reproducible evaluation and fair comparison.

\paragraph{Tasks.}
For each embodiment, \texttt{RoboColosseum} maintains an extensible registry of manipulation tasks, of which five per embodiment, 20 in total, are included in the current evaluation (\Cref{app:finetuning_track_tasks}). Each task is specified by a task card that defines the natural-language instruction, the objects, the initial setup, the success criteria, the timeout, the evaluation site, and a set of task-specific progress criteria. The tasks span pick-and-place, stacking (bowls, cubes, and cups), object reorientation (standing a fallen bottle upright), articulated objects (drawers, a laptop lid, and a microwave door), deformable objects (towels and clothing), liquid pouring, tool use (sweeping with a dustpan and brush), bimanual coordination tasks with the bimanual YAMs, and loco-manipulation with the G1 Humanoid.

\paragraph{Data Collection.} For each task, we collect 100 high-quality teleoperated demonstrations using embodiment-specific interfaces: GELLO~\citep{wu2024gello} for Bimanual-YAM, a leader--follower arm setup for SO-101 joint-space control, the DROID Quest teleoperation system for Franka DROID, and a Pico VR headset with an off-the-shelf whole-body controller for the G1 humanoid. Demonstrations are collected at the evaluation site using the same robot and environment to match deployment conditions. All policies are fine-tuned on the same demonstrations, which we publicly release.

\paragraph{Fine-tuning protocol.}
Each evaluated model is fully fine-tuned from its base pretrained checkpoint using the same 100 demonstrations, which constitute the only embodiment- and task-specific data used during adaptation. Each policy follows its official fine-tuning recipe until convergence, allowing optimization schedules and training durations to vary across architectures. We thus control the adaptation data and evaluation protocol while allowing model-specific training procedures. The full protocol, base checkpoints, and per-model hyperparameters are provided in \Cref{app:finetuning_protocol}. This design complements DataComp~\citep{gadre2023datacompl}: we hold the data fixed and vary the training procedure, rather than the reverse.

\paragraph{Evaluation and metrics.}
After fine-tuning, each policy is evaluated for $N = 25$ trials per task using its original inference implementation. The task goal remains fixed, while object placements and environmental conditions may vary within the task specification. Each task is evaluated using $K$ task-specific progress criteria. For each rollout $n$, let $s_n \in \{0, 1, \ldots, K\}$ be the number of criteria satisfied; the rollout's normalized progress is the fraction $s_n/K$, so each criterion is weighted equally. Criteria need not be satisfied in a fixed order. Success requires satisfying the task's full success requirements. Evaluators may terminate rollouts that risk damage to the robot or scene objects; these rollouts receive $s_n = 0$ and are counted as failures. For each policy--task pair, the reported progress score (PS) is the mean of this fraction across evaluation trials, and the success rate (SR) separately measures the fraction of trials satisfying the full success requirements:
\begin{equation}
\mathrm{PS} = \frac{1}{N}\sum_{n=1}^{N} \frac{s_n}{K},
\qquad
\mathrm{SR} = \frac{1}{N}\sum_{n=1}^{N} \mathbf{1}\left[\text{rollout } n \text{ succeeds}\right].
\label{eq:ps_sr}
\end{equation}
SR measures task completion, while PS captures how much of the task specification a policy satisfies, even when the task is not completed. Normalization by $K$ places PS on a common scale across tasks with different numbers of criteria. Per-task results are reported in \Cref{tab:finetuning_track_results}.

\subsubsection{Open Track}
\label{sec:open_track}
The \texttt{Open Track} evaluates models without task-specific adaptation on in-the-wild manipulation tasks proposed by participating evaluation sites.

\paragraph{Evaluation protocol and active matchmaking.}
In the \texttt{Open Track}, an evaluator defines a feasible task using locally available objects and environments and provides a natural-language instruction. \texttt{RoboColosseum} assigns two eligible models, denoted $A$ and $B$, whose identities remain hidden from the evaluator throughout evaluation. The models are executed consecutively on the same robot and task configuration. The evaluator records a pairwise preference ($A$ wins, $B$ wins, or a tie), task-progress scores, natural-language feedback, and a perceived task difficulty rating from 1 to 10.

We use active matchmaking to allocate evaluations across model pairs. After collecting initial pairwise comparisons to establish a preliminary ranking, we adopt the active sampling strategy of Chatbot Arena~\citep{chiang2024chatbotarena}, prioritizing pairs with the largest estimated reduction in uncertainty:
\begin{equation}
P_t(a) \propto
\sqrt{\frac{\hat{\Sigma}_{t,a,a}}{n_a}}
-
\sqrt{\frac{\hat{\Sigma}_{t,a,a}}{n_a+1}},
\label{eq:active_sampling}
\end{equation}
where $a$ denotes a candidate model pair, $n_a$ is its prior comparison count, and $\hat{\Sigma}_{t,a,a}$ is its estimated variance at iteration $t$. This strategy concentrates evaluations on uncertain or sparsely evaluated pairs while limiting redundant comparisons.



\paragraph{Evaluation Metrics}
\texttt{Open Track} evaluations vary substantially in task difficulty, making a raw win rate or task-agnostic pairwise ranking potentially sensitive to which tasks a model happens to encounter. We therefore rank policies using a \emph{Difficulty-Calibrated Rasch} (DCR) model that jointly estimates policy ability and task difficulty from physical success outcomes.

Let $p$ denote a policy and $t$ an \texttt{Open Track} task. Each policy has a latent ability $\theta_p$, and each task has a latent difficulty $\beta_t$. The probability that policy $p$ succeeds on task $t$ is modeled using a Rasch/1PL formulation~\citep{rasch1960probabilistic}:
\begin{equation}
q_{p,t}
=
P(y_{p,t}=1 \mid \theta_p,\beta_t)
=
\sigma(\theta_p-\beta_t),
\label{eq:dcr_success}
\end{equation}
where $y_{p,t}\in\{0,1\}$ denotes task success and $\sigma(\cdot)$ is the logistic function. Higher $\theta_p$ therefore indicates greater policy capability, while higher $\beta_t$ indicates a more difficult task.

Each \texttt{Open Track} task is additionally assigned a human difficulty rating
$d_t\in\{1,\ldots,10\}$. Because most in-the-wild tasks receive only a small number of physical evaluations, we do not treat this rating as a fixed
numerical weight. Rather, we use it as prior information over latent task difficulty, following prior work on incorporating judgmental difficulty
estimates into item-response models~\citep{yi2009irt}.

Let $\mu_k$ denote the typical latent difficulty of tasks assigned rating $k$, constrained such that
\begin{equation}
\mu_1 < \mu_2 < \cdots < \mu_{10}.
\end{equation}
For a task with $d_t=k$, we use
\begin{equation}
\beta_t \mid d_t=k
\sim
\mathcal{N}(\mu_k,\tau^2).
\label{eq:dcr_difficulty_prior}
\end{equation}
Physical outcomes can therefore move the inferred difficulty of an individual task away from its human-rated category mean.

For each matched A/B evaluation, both policies are executed on the same task and their individual success outcomes are retained. Consequently, the model distinguishes cases in which both policies succeed from cases in which both fail: the former provides evidence that the task is easier, while the latter provides evidence that it is harder. Policy abilities, task difficulties, and the human-difficulty calibration parameters are inferred jointly from all \texttt{Open Track} evaluations.

We rank policies by posterior mean ability,
\begin{equation}
s_p = \mathbb{E}[\theta_p \mid \mathcal{D}],
\end{equation}
where $\mathcal{D}$ denotes the collected \texttt{Open Track} evaluations. For easier interpretation on the public leaderboard, we report the same learned ability on an Elo-style scale:
\begin{equation}
R_p
=
1500
+
\frac{400}{\ln 10}\,
s_p.
\label{eq:dcr_rating}
\end{equation}
This transformation is linear and therefore does not change the model ordering, uncertainty, or difficulty-aware estimation. We refer to $R_p$ as the \emph{DCR Rating (Elo scale)}; it should not be interpreted as an Elo rating obtained through sequential Elo updates.

For ranking robustness analysis, we additionally compare DCR against the Bradley--Terry--Davidson (BTD) model~\citep{bradley1952rank,davidson1970extending}, which aggregates pairwise win/loss/tie outcomes without explicitly modeling task difficulty.

\section{Experiments}
\label{sec:experiments}

We conducted in-the-wild evaluations with \texttt{RoboColosseum} over a duration of 41 days, from 17 September 2026 to 28 October 2026. We evaluated eight models across the Fine-Tune and Open Tracks: five vision-language-action (VLA) models and three frontier models. Using these results, we investigate the following research questions:

\begin{itemize}
    \item \textbf{RQ1: Open-world capabilities.} What capabilities and failure modes do general-purpose robot models exhibit across diverse real-world tasks and environments?
    \item \textbf{RQ2: Task adaptation.} How effectively do pretrained robot policies adapt to unseen manipulation tasks given a fixed budget of task-specific demonstrations?
    \item \textbf{RQ3: Performance consistency.} How consistent are model rankings across evaluation tracks and robot embodiments?
    \item \textbf{RQ4: Evaluation reliability and efficiency.} How reliably can distributed pairwise evaluations estimate model rankings, and how much can active matchmaking reduce the number of physical trials required?
    \item \textbf{RQ5: Effect of evaluation scale.} How does the number of physical evaluations affect ranking uncertainty and the stability of the Open Track leaderboard?
\end{itemize}
\subsection{Experimental Setup}
\label{sec:exp_setup}

\paragraph{Models and protocols.}
We evaluate five VLA models---GR00T N1.7~\citep{gr00tn1_2025}, $\pi_{0.5}$~\citep{black25p05}, MolmoAct2~\citep{fang2026molmoact2}, G0.5~\citep{liu2026g05}, and LingBot-VLA v2~\citep{lingbotvla2}---and three API-accessible frontier models: GPT-6 Astra, Claude Opus 5.5, and Grok 4.7. The \texttt{Fine-tune Track} evaluates VLA models initialized from base pretrained checkpoints on 20 tasks across four embodiments, with five tasks per embodiment, 100 demonstrations per task, and 25 evaluation trials per model--task pair. The \texttt{Open Track} comprises 5,699 pairwise comparisons across 5,699 tasks, 124 sites, and four embodiments, following \Cref{sec:robocolosseum_benchmark}. VLA models use embodiment-specific checkpoints in this track; frontier-model agents participate exclusively in this track through the interface in \Cref{sec:setup}.

\paragraph{Metrics and statistical analysis.}
For the \texttt{Fine-tune Track}, we report success rate (SR) and normalized progress score (PS), and assess fine-tuning differences through paired comparisons with multiple-comparison correction. For the \texttt{Open Track}, DCR serves as the primary leaderboard metric, with Bradley--Terry--Davidson (BTD), sequential Elo, and average pairwise score as ranking baselines. For category-level analysis, we fit a separate progress-based BTD model within each manipulation category: higher normalized task progress determines the winner, and equal progress yields a tie. We report each system's \emph{expected comparison score}, defined as its probability of winning plus half its probability of tying against a hypothetical system of average ability; 50\% thus indicates average relative performance within that category. We report uncertainty for success rates and model ratings, quantify ranking consistency using Spearman's correlation and pairwise ordering agreement, and assess ranking stability through repeated subsampling at increasing evaluation budgets.

\subsection{Open-World Capabilities (RQ1)}
\label{sec:open_world_capabilities}







\begin{figure}[t]
    \centering
    \includegraphics[width=\textwidth]{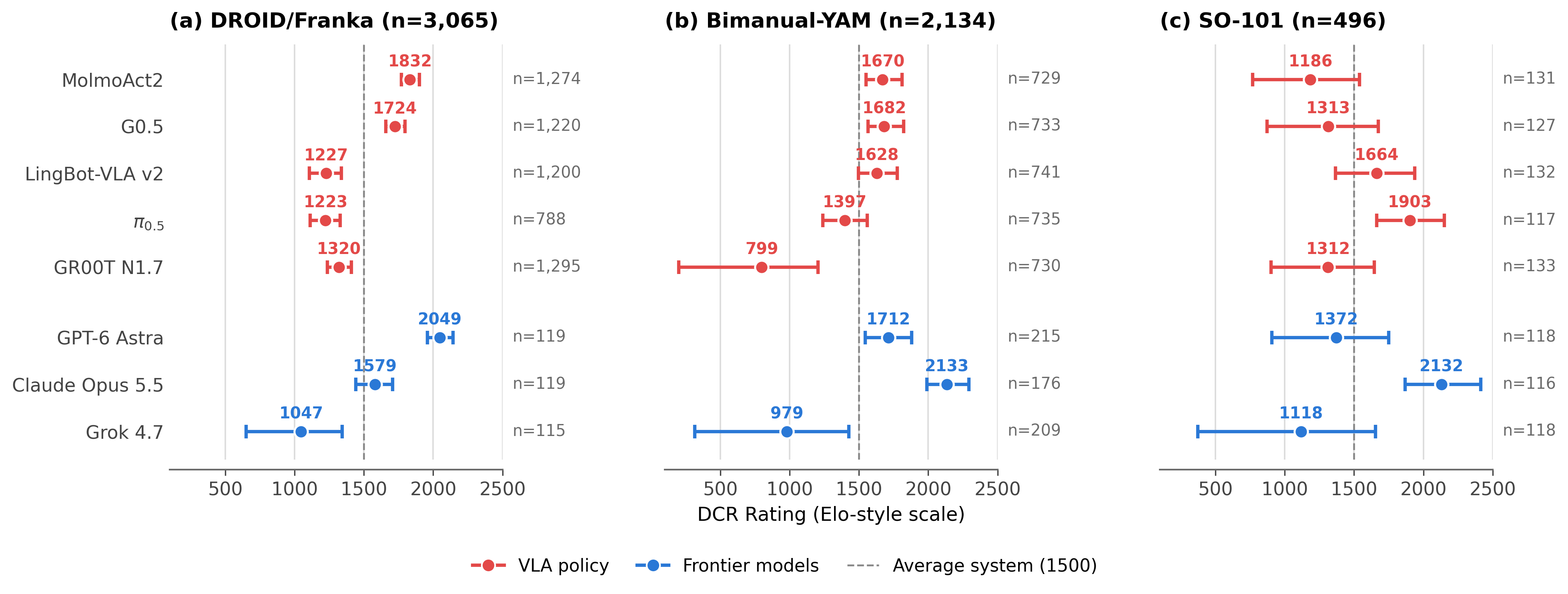}
    \caption{\textbf{Open Track DCR ratings across embodiments.}
    Points show posterior mean DCR ratings on an Elo-style scale, with 95\% credible intervals; $n$ is the number of evaluations involving each system. 
    }
    \label{fig:open_track_dcr_by_embodiment}
\end{figure}
\begin{figure}[t]
    \centering
    \includegraphics[width=\textwidth]{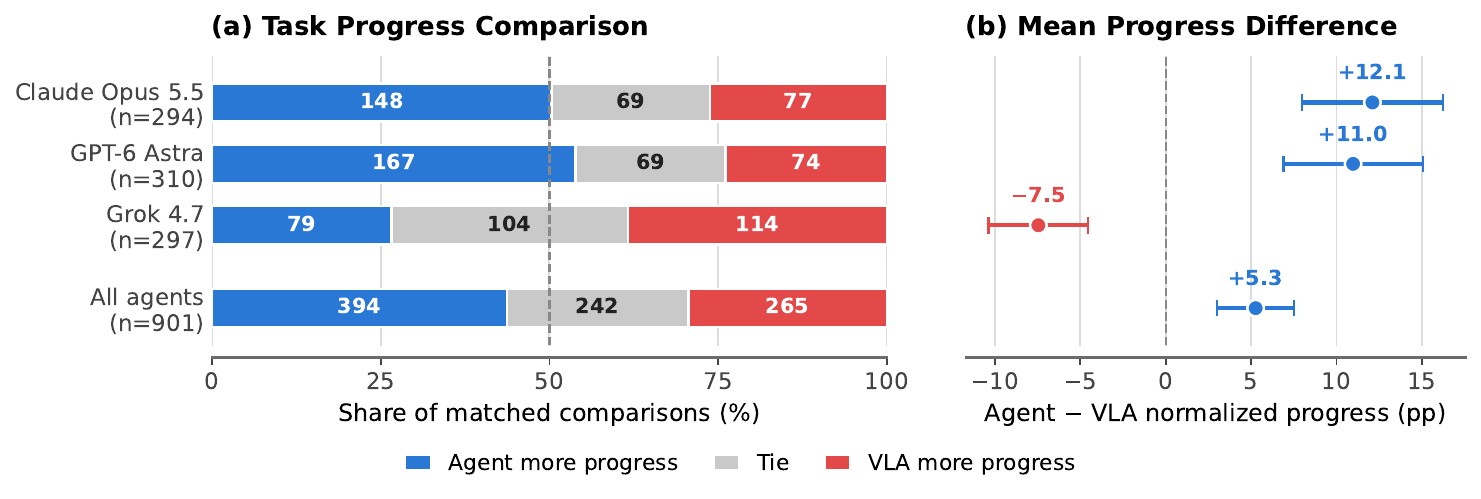}
    
\caption{\textbf{Task progress of frontier AI models versus VLA policies.}
Comparison of normalized task progress across 901 Open Track evaluations, each involving a frontier AI model and a VLA policy on the same task.
(a) Percentage of comparisons in which the agent makes more progress, the VLA makes more progress, or both make equal progress. Numbers inside the bars show comparison counts.
(b) Mean difference in task progress (agent minus VLA, in percentage points), with 95\% bootstrap confidence intervals. Positive values favor the agent.}

    \label{fig:agent_vla_paired_progress}
\end{figure}

\paragraph{Frontier agents achieve competitive manipulation performance, but performance varies across agents and embodiments.}
Success-based DCR rankings show that frontier AI models can achieve competitive manipulation performance compared with specialized VLA policies, although their relative strengths differ across robot embodiments. In \Cref{fig:open_track_dcr_by_embodiment}, GPT-6 Astra achieves the highest DCR point estimate on DROID/Franka, whereas Claude Opus 5.5 leads on SO-101 and Bimanual-YAM. This competitiveness is also reflected in normalized task progress. Across 901 paired evaluations, agents achieve greater progress in 394 comparisons, VLAs in 265, and the remaining 242 are tied, yielding an average agent advantage of 5.3 percentage points (\Cref{fig:agent_vla_paired_progress}). However, this advantage is not consistent across agents: Claude Opus 5.5 and GPT-6 Astra achieve mean advantages of 12.1 and 11.0 percentage points, respectively, whereas Grok 4.7 trails its VLA counterparts by 7.5 points. Together, these findings show that frontier agents can match or outperform specialized VLA policies in physical manipulation, but their relative performance depends on the agent and robot embodiment.


\paragraph{Strong relative performance does not necessarily translate into reliable task completion.}
To examine manipulation capabilities beyond aggregate model rankings, we group \texttt{Open Track} tasks into six categories: Pick/Place, Stack/Arrange, Open/Close, Reorient, Deformable, and Tool/Interaction, using a fixed language-model classifier following RoboArena~\citep{atreya2025roboarena} (details in Appendix~\ref{app:llm-task-classification}). \Cref{fig:open_track_capability_heatmap}(a) shows the estimated BTD comparison scores for each manipulation category, while (b)--(c) report mean normalized task progress and full-task success rates. MolmoAct2 achieves the highest BTD point estimate among the five VLA policies in all six categories, with expected comparison scores ranging from 58\% to 77\%. Despite this relative advantage, its observed task success rates range from only 6\% to 13\%, and no evaluated system exceeds 13\% success in any category. Although absolute success rates also depend on the tasks and embodiments evaluated for each system, the consistently low success rates across all six categories suggest that current policies' limitations extend beyond individual manipulation skills, with even the strongest evaluated systems struggling to reliably complete diverse open-world tasks.

\begin{figure}[t]
    \centering
    \includegraphics[width=\textwidth]{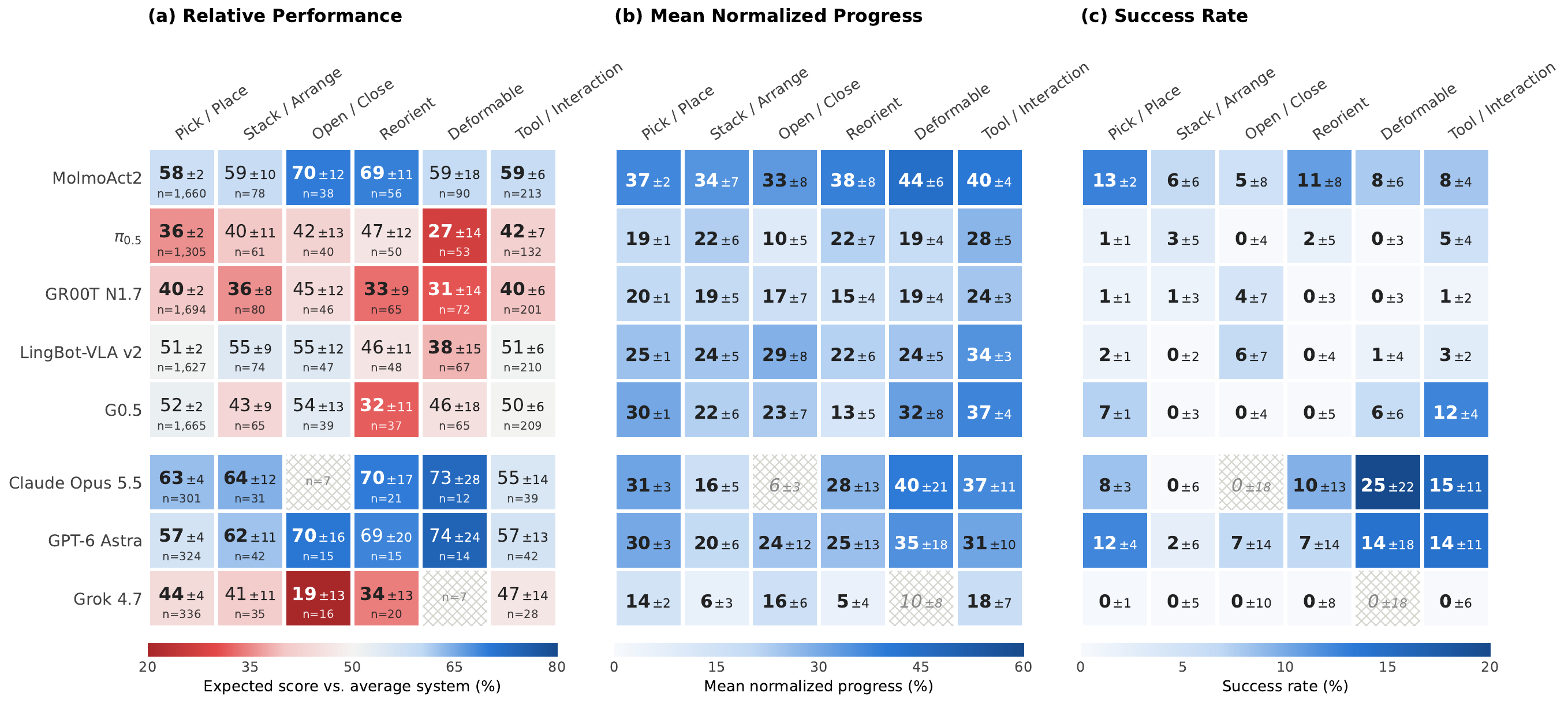}
    \caption{\textbf{Open Track performance by manipulation category.}
    (a) Relative performance: category-specific, progress-based BTD expected comparison scores, i.e., the probability of winning plus half the probability of tying against a system of average ability, where a comparison is won by the system with higher normalized progress and 50\% indicates average relative performance; bold values have 95\% intervals that exclude 50\%. (b) Mean normalized task progress and (c) success rate of each system's own executions, both in percent. Values are followed by $\pm$ half the width of the 95\% interval (bootstrap for (a) and (b), Wilson for (c)). $n$, shown in (a), is the number of evaluations involving each system and is the same in all three panels. Hatched cells have insufficient support ($n<10$ or no evaluations). Relative scores in (a) compare systems under matched conditions, whereas the absolute values in (b) and (c) also reflect differences in the embodiments and tasks on which each system was evaluated.}
    \label{fig:open_track_capability_heatmap}
\end{figure}

\begin{figure}[t]
    \centering
    \includegraphics[width=0.9\textwidth]{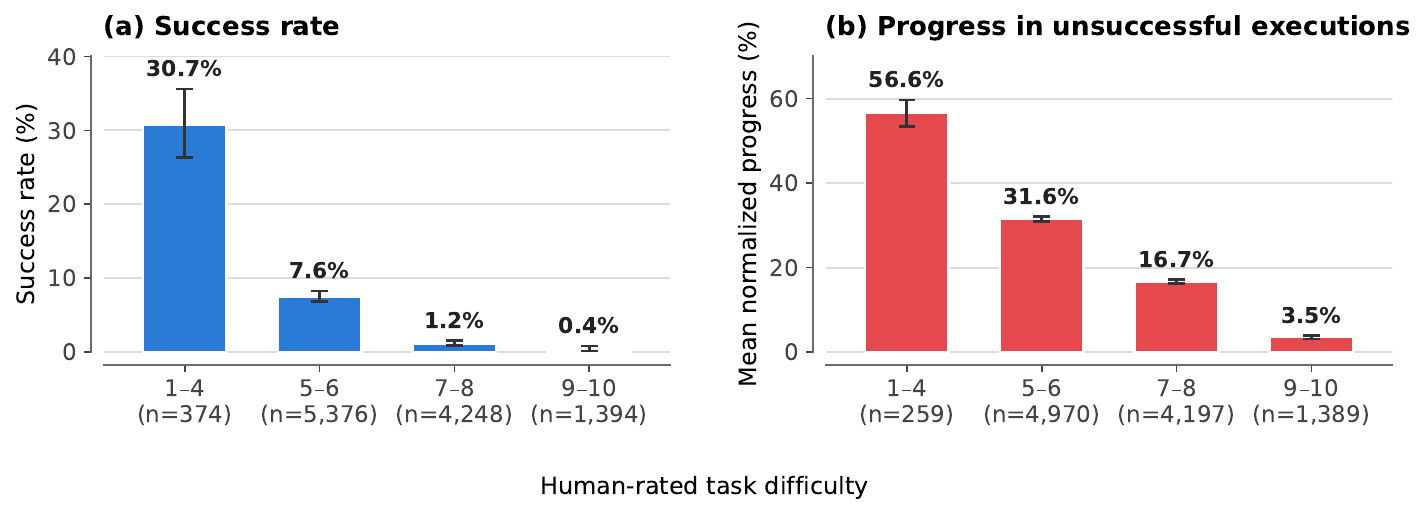}
    \caption{\textbf{Physical task performance across human-rated task difficulty.}
    Open Track policy executions grouped by the human-rated difficulty of the task, with each pairwise evaluation contributing one execution per compared system. (a) Success rate, with 95\% Wilson confidence intervals. (b) Mean normalized progress among unsuccessful executions, with 95\% bootstrap confidence intervals. $n$ is the number of executions in each difficulty range.}
    \label{fig:difficulty_success_progress}
\end{figure}

\paragraph{Policies struggle to make meaningful progress on the most difficult manipulation tasks.}
We examine how physical execution performance varies across human-rated task difficulty levels (\Cref{fig:difficulty_success_progress}). Full-task success decreases sharply with difficulty, from 30.7\% on tasks rated 1--4 to just 0.4\% on tasks rated 9--10. This decline extends beyond task completion: mean normalized progress also falls from 56.6\% to 3.5\% across the same difficulty ranges, indicating that policies satisfy substantially fewer intermediate progress criteria on more challenging tasks. Although these trends are observational and may partly reflect differences in task and embodiment composition, they reveal that current policies struggle not only to complete difficult tasks, but also to make meaningful progress toward their goals. On the most challenging tasks, failures appear to arise well before the final success condition, suggesting that current policies lack the capabilities needed to execute even substantial portions of complex manipulation tasks.

\begin{takeaway}
Frontier AI agents achieve manipulation performance competitive with specialized VLA policies, although their relative strengths vary across agents and robot embodiments. However, reliable open-world manipulation remains a challenge for both types of systems: even the strongest policies achieve low task completion rates across manipulation categories, and their ability to make meaningful progress declines sharply on more difficult tasks. These findings highlight a fundamental limitation of current robot policies: while they demonstrate promising manipulation capabilities, they struggle to sustain successful execution as task complexity increases.
\end{takeaway}

\subsection{Task Adaptation Finetune(RQ2)}

\paragraph{Post-fine-tuning performance differentiates policies which are broadly adaptable, from those that are task-adaptable.}
\Cref{tab:finetuning_track_results} shows that performance after adaptation with the same 100 demonstrations varies substantially across tasks, even within a single embodiment. On DROID/Franka, LingBot-VLA v2 achieves 100\% success on \texttt{open\_drawer} but 0\% on \texttt{marker\_in\_holder}, whereas MolmoAct2 exhibits the opposite pattern, achieving 0\% and 88\%, respectively. In contrast, some policies achieve consistently high success across multiple tasks.  On Bimanual-YAM, $\pi_{0.5}$ achieves 96--100\% success on four of the five tasks and 72\% on \texttt{stack\_cups}, demonstrating consistently strong performance across the tested suite. By comparison, G0.5 achieves 100\% success on both \texttt{drawer} and \texttt{microwave} but only 20\% on \texttt{stack\_cups}. These results show that successful adaptation to individual tasks does not necessarily imply broad adaptability: under the same demonstration budget, some policies adapt effectively across diverse tasks, while others achieve strong performance only on selected tasks.

\definecolor{RTBlue}{HTML}{1c6ef3}
\definecolor{RTBlueLight}{HTML}{d2e3fc}

\providecommand{\rtbest}[1]{\textbf{#1}}
\providecommand{\rtsecond}[1]{\underline{#1}}
\providecommand{\rtpm}[2]{#1\,$\pm$\,#2}

\begin{table*}[t]
\centering
\caption{\textbf{Per-task results on the Fine-tune track.} Cells report normalized progress score (PS, \%) and full-success rate (SR, \%) as mean $\pm$ standard error (SE), using 25 trial slots per policy and task. Unexecuted trials following a safety stop are scored zero. Embodiment averages weight the five tasks equally. Best and second-best means for each metric are marked in \textbf{bold} and \underline{underline}; ties share ranks at the displayed precision. $^\dagger$Evaluation stopped for safety reasons: LingBot-VLA v2 on YAM drawer after 1 trial, and GR00T N1.7 on YAM microwave after 9 trials.}
\label{tab:finetuning_track_results}
\footnotesize
\setlength{\tabcolsep}{2.4pt}
\renewcommand{\arraystretch}{1.12}
\resizebox{\textwidth}{!}{%
\begin{tabular}{@{}l*{5}{cc}@{}}
\toprule
\multirow{2}{*}{\textcolor{RTBlue}{\textbf{Task}}} & \multicolumn{2}{c}{\textcolor{RTBlue}{\textbf{GR00T N1.7}}} & \multicolumn{2}{c}{\textcolor{RTBlue}{\textbf{{\boldmath$\pi_{0.5}$}}}} & \multicolumn{2}{c}{\textcolor{RTBlue}{\textbf{MolmoAct2}}} & \multicolumn{2}{c}{\textcolor{RTBlue}{\textbf{G0.5}}} & \multicolumn{2}{c}{\textcolor{RTBlue}{\textbf{LingBot-VLA v2}}} \\
\cmidrule(lr){2-3}\cmidrule(lr){4-5}\cmidrule(lr){6-7}\cmidrule(lr){8-9}\cmidrule(lr){10-11}
 & PS & SR & PS & SR & PS & SR & PS & SR & PS & SR \\
\midrule
\multicolumn{11}{@{}l}{\textcolor{RTBlue}{\textbf{SO-101}}} \\
\quad stack\_bowls & \rtpm{\rtbest{97.0}}{3.0} & \rtpm{\rtbest{96.0}}{3.9} & \rtpm{\rtsecond{94.0}}{4.2} & \rtpm{\rtsecond{92.0}}{5.4} & \rtpm{88.0}{5.6} & \rtpm{84.0}{7.3} & \rtpm{51.0}{8.0} & \rtpm{36.0}{9.6} & \rtpm{62.0}{7.4} & \rtpm{40.0}{9.8} \\
\quad remove\_screwdriver & \rtpm{\rtsecond{90.0}}{4.1} & \rtpm{\rtsecond{80.0}}{8.0} & \rtpm{\rtbest{93.0}}{3.4} & \rtpm{\rtbest{84.0}}{7.3} & \rtpm{69.0}{5.5} & \rtpm{40.0}{9.8} & \rtpm{72.0}{6.7} & \rtpm{52.0}{10.0} & \rtpm{67.0}{7.2} & \rtpm{52.0}{10.0} \\
\quad upright\_bottle & \rtpm{\rtbest{98.0}}{2.0} & \rtpm{\rtbest{96.0}}{3.9} & \rtpm{\rtsecond{86.0}}{6.1} & \rtpm{\rtsecond{80.0}}{8.0} & \rtpm{84.0}{6.8} & \rtpm{\rtsecond{80.0}}{8.0} & \rtpm{66.0}{6.6} & \rtpm{36.0}{9.6} & \rtpm{84.0}{5.4} & \rtpm{64.0}{9.6} \\
\quad stack\_cubes & \rtpm{\rtsecond{81.0}}{6.2} & \rtpm{\rtsecond{68.0}}{9.3} & \rtpm{\rtbest{90.0}}{5.6} & \rtpm{\rtbest{84.0}}{7.3} & \rtpm{69.0}{7.0} & \rtpm{52.0}{10.0} & \rtpm{42.0}{5.7} & \rtpm{16.0}{7.3} & \rtpm{20.0}{2.0} & \rtpm{0.0}{0.0} \\
\quad cube\_drawer & \rtpm{\rtbest{79.0}}{7.3} & \rtpm{\rtbest{72.0}}{9.0} & \rtpm{\rtsecond{64.0}}{8.7} & \rtpm{48.0}{10.0} & \rtpm{49.0}{9.4} & \rtpm{40.0}{9.8} & \rtpm{63.0}{9.7} & \rtpm{\rtsecond{60.0}}{9.8} & \rtpm{43.0}{9.0} & \rtpm{28.0}{9.0} \\
\cmidrule(lr){1-11}
\quad\textit{Average} & \rtpm{\rtbest{89.0}}{2.2} & \rtpm{\rtbest{82.4}}{3.2} & \rtpm{\rtsecond{85.4}}{2.6} & \rtpm{\rtsecond{77.6}}{3.5} & \rtpm{71.8}{3.1} & \rtpm{59.2}{4.0} & \rtpm{58.8}{3.3} & \rtpm{40.0}{4.2} & \rtpm{55.2}{3.0} & \rtpm{36.8}{3.8} \\
\midrule
\multicolumn{11}{@{}l}{\textcolor{RTBlue}{\textbf{DROID Franka}}} \\
\quad close\_laptop & \rtpm{13.0}{6.1} & \rtpm{4.0}{3.9} & \rtpm{\rtbest{92.0}}{2.4} & \rtpm{\rtbest{68.0}}{9.3} & \rtpm{\rtsecond{80.0}}{3.2} & \rtpm{\rtsecond{28.0}}{9.0} & \rtpm{26.0}{1.0} & \rtpm{0.0}{0.0} & \rtpm{51.0}{7.7} & \rtpm{20.0}{8.0} \\
\quad mug\_on\_coaster & \rtpm{35.0}{7.4} & \rtpm{12.0}{6.5} & \rtpm{\rtbest{100.0}}{0.0} & \rtpm{\rtbest{100.0}}{0.0} & \rtpm{\rtsecond{86.0}}{4.3} & \rtpm{\rtsecond{56.0}}{9.9} & \rtpm{35.0}{3.8} & \rtpm{0.0}{0.0} & \rtpm{66.0}{7.6} & \rtpm{48.0}{10.0} \\
\quad open\_drawer & \rtpm{71.0}{8.4} & \rtpm{64.0}{9.6} & \rtpm{\rtsecond{80.0}}{8.2} & \rtpm{\rtsecond{80.0}}{8.0} & \rtpm{30.0}{2.0} & \rtpm{0.0}{0.0} & \rtpm{24.0}{1.0} & \rtpm{0.0}{0.0} & \rtpm{\rtbest{100.0}}{0.0} & \rtpm{\rtbest{100.0}}{0.0} \\
\quad marker\_in\_holder & \rtpm{\rtsecond{87.0}}{5.2} & \rtpm{\rtsecond{76.0}}{8.5} & \rtpm{85.0}{6.1} & \rtpm{\rtsecond{76.0}}{8.5} & \rtpm{\rtbest{95.0}}{2.9} & \rtpm{\rtbest{88.0}}{6.5} & \rtpm{45.0}{2.9} & \rtpm{0.0}{0.0} & \rtpm{0.0}{0.0} & \rtpm{0.0}{0.0} \\
\quad towel\_in\_basket & \rtpm{76.0}{7.3} & \rtpm{56.0}{9.9} & \rtpm{\rtsecond{99.0}}{1.0} & \rtpm{\rtsecond{96.0}}{3.9} & \rtpm{\rtbest{100.0}}{0.0} & \rtpm{\rtbest{100.0}}{0.0} & \rtpm{94.0}{2.6} & \rtpm{80.0}{8.0} & \rtpm{80.0}{3.5} & \rtpm{36.0}{9.6} \\
\cmidrule(lr){1-11}
\quad\textit{Average} & \rtpm{56.4}{3.1} & \rtpm{42.4}{3.6} & \rtpm{\rtbest{91.2}}{2.1} & \rtpm{\rtbest{84.0}}{3.1} & \rtpm{\rtsecond{78.2}}{1.3} & \rtpm{\rtsecond{54.4}}{3.0} & \rtpm{44.8}{1.1} & \rtpm{16.0}{1.6} & \rtpm{59.4}{2.3} & \rtpm{40.8}{3.2} \\
\midrule
\multicolumn{11}{@{}l}{\textcolor{RTBlue}{\textbf{YAMs}}} \\
\quad dustpan & \rtpm{84.0}{5.8} & \rtpm{72.0}{9.0} & \rtpm{\rtbest{100.0}}{0.0} & \rtpm{\rtbest{100.0}}{0.0} & \rtpm{\rtsecond{94.7}}{3.7} & \rtpm{\rtsecond{92.0}}{5.4} & \rtpm{\rtsecond{94.7}}{2.5} & \rtpm{84.0}{7.3} & \rtpm{70.7}{5.9} & \rtpm{44.0}{9.9} \\
\quad drawer & \rtpm{68.0}{4.0} & \rtpm{4.0}{3.9} & \rtpm{\rtsecond{99.0}}{1.0} & \rtpm{\rtsecond{96.0}}{3.9} & \rtpm{90.0}{2.9} & \rtpm{64.0}{9.6} & \rtpm{\rtbest{100.0}}{0.0} & \rtpm{\rtbest{100.0}}{0.0} & \rtpm{0.0}{0.0}$^\dagger$ & \rtpm{0.0}{0.0}$^\dagger$ \\
\quad microwave & \rtpm{0.0}{0.0}$^\dagger$ & \rtpm{0.0}{0.0}$^\dagger$ & \rtpm{\rtbest{100.0}}{0.0} & \rtpm{\rtbest{100.0}}{0.0} & \rtpm{90.7}{5.3} & \rtpm{88.0}{6.5} & \rtpm{\rtbest{100.0}}{0.0} & \rtpm{\rtbest{100.0}}{0.0} & \rtpm{\rtsecond{96.0}}{4.0} & \rtpm{\rtsecond{96.0}}{3.9} \\
\quad stack\_cups & \rtpm{44.0}{5.3} & \rtpm{0.0}{0.0} & \rtpm{\rtbest{87.0}}{5.0} & \rtpm{\rtbest{72.0}}{9.0} & \rtpm{27.0}{3.2} & \rtpm{0.0}{0.0} & \rtpm{\rtsecond{58.0}}{5.9} & \rtpm{\rtsecond{20.0}}{8.0} & \rtpm{5.0}{2.0} & \rtpm{0.0}{0.0} \\
\quad clean\_table & \rtpm{67.0}{5.1} & \rtpm{24.0}{8.5} & \rtpm{\rtbest{100.0}}{0.0} & \rtpm{\rtbest{100.0}}{0.0} & \rtpm{58.0}{3.5} & \rtpm{4.0}{3.9} & \rtpm{\rtsecond{93.0}}{2.3} & \rtpm{\rtsecond{72.0}}{9.0} & \rtpm{70.0}{3.2} & \rtpm{12.0}{6.5} \\
\cmidrule(lr){1-11}
\quad\textit{Average} & \rtpm{52.6}{2.0} & \rtpm{20.0}{2.6} & \rtpm{\rtbest{97.2}}{1.0} & \rtpm{\rtbest{93.6}}{2.0} & \rtpm{72.1}{1.7} & \rtpm{49.6}{2.7} & \rtpm{\rtsecond{89.1}}{1.4} & \rtpm{\rtsecond{75.2}}{2.8} & \rtpm{48.3}{1.6} & \rtpm{30.4}{2.5} \\
\midrule
\multicolumn{11}{@{}l}{\textcolor{RTBlue}{\textbf{G1 Humanoid}}} \\
\quad plate\_into\_rack & \rtpm{\rtbest{33.6}}{3.2} & \rtpm{\rtbest{0.0}}{0.0} & \rtpm{1.6}{1.1} & \rtpm{\rtbest{0.0}}{0.0} & \rtpm{27.2}{3.0} & \rtpm{\rtbest{0.0}}{0.0} & \rtpm{21.6}{1.1} & \rtpm{\rtbest{0.0}}{0.0} & \rtpm{\rtsecond{28.0}}{2.8} & \rtpm{\rtbest{0.0}}{0.0} \\
\quad carry\_box & \rtpm{11.4}{2.3} & \rtpm{\rtbest{0.0}}{0.0} & \rtpm{8.6}{3.3} & \rtpm{\rtbest{0.0}}{0.0} & \rtpm{\rtbest{26.3}}{1.1} & \rtpm{\rtbest{0.0}}{0.0} & \rtpm{\rtsecond{21.7}}{2.4} & \rtpm{\rtbest{0.0}}{0.0} & \rtpm{16.0}{0.9} & \rtpm{\rtbest{0.0}}{0.0} \\
\quad fold\_towel & \rtpm{16.8}{1.5} & \rtpm{\rtbest{0.0}}{0.0} & \rtpm{\rtbest{24.8}}{2.9} & \rtpm{\rtbest{0.0}}{0.0} & \rtpm{2.4}{1.3} & \rtpm{\rtbest{0.0}}{0.0} & \rtpm{3.2}{1.5} & \rtpm{\rtbest{0.0}}{0.0} & \rtpm{\rtsecond{22.4}}{1.3} & \rtpm{\rtbest{0.0}}{0.0} \\
\quad load\_clothes\_washing\_machine & \rtpm{29.6}{5.0} & \rtpm{\rtbest{0.0}}{0.0} & \rtpm{\rtbest{44.0}}{6.0} & \rtpm{\rtbest{0.0}}{0.0} & \rtpm{\rtsecond{43.2}}{4.6} & \rtpm{\rtbest{0.0}}{0.0} & \rtpm{0.0}{0.0} & \rtpm{\rtbest{0.0}}{0.0} & \rtpm{36.8}{6.3} & \rtpm{\rtbest{0.0}}{0.0} \\
\quad hang\_keys & \rtpm{\rtbest{22.0}}{2.5} & \rtpm{\rtbest{0.0}}{0.0} & \rtpm{2.7}{1.2} & \rtpm{\rtbest{0.0}}{0.0} & \rtpm{0.0}{0.0} & \rtpm{\rtbest{0.0}}{0.0} & \rtpm{\rtsecond{20.0}}{2.9} & \rtpm{\rtbest{0.0}}{0.0} & \rtpm{12.7}{1.5} & \rtpm{\rtbest{0.0}}{0.0} \\
\cmidrule(lr){1-11}
\quad\textit{Average} & \rtpm{\rtsecond{22.7}}{1.4} & \rtpm{\rtbest{0.0}}{0.0} & \rtpm{16.3}{1.5} & \rtpm{\rtbest{0.0}}{0.0} & \rtpm{19.8}{1.1} & \rtpm{\rtbest{0.0}}{0.0} & \rtpm{13.3}{0.8} & \rtpm{\rtbest{0.0}}{0.0} & \rtpm{\rtbest{23.2}}{1.4} & \rtpm{\rtbest{0.0}}{0.0} \\
\bottomrule
\end{tabular}%
}
\par\vspace{3pt}

\end{table*}

\begin{figure}[t]
    \centering
    \includegraphics[width=\textwidth]{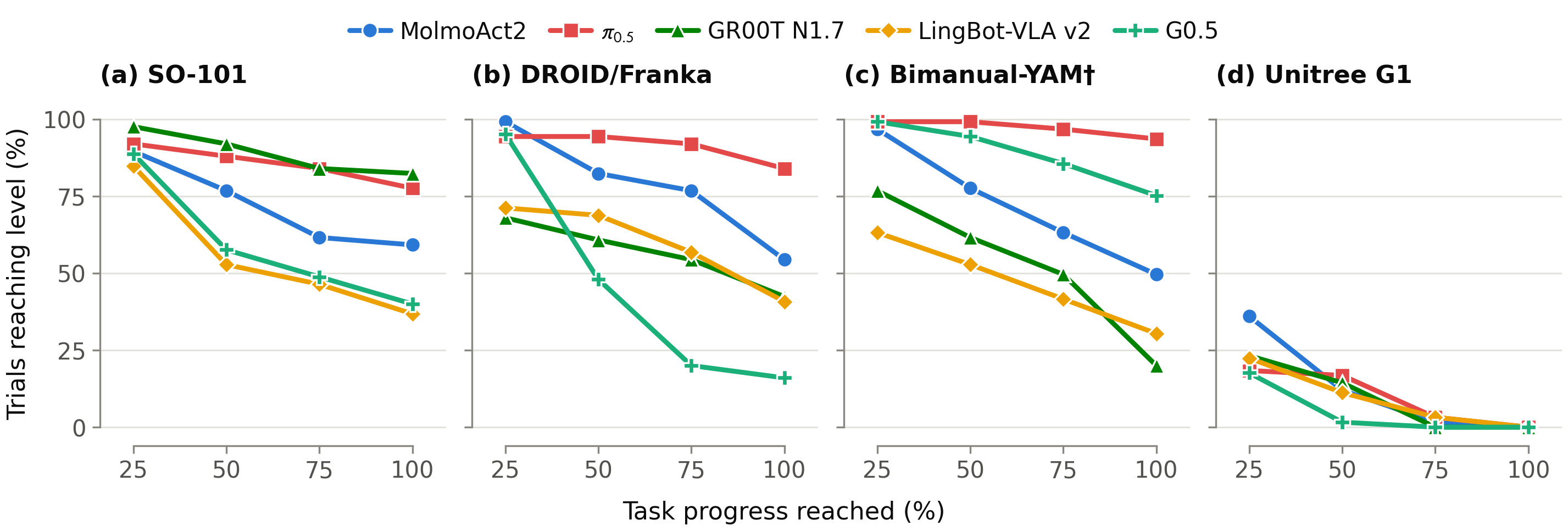}
    \caption{\textbf{Task progress reachability.} Percentage of trials satisfying at least a given fraction of task criteria, averaged over the five tasks of each embodiment with equal weight. The 100\% point is the full success rate. $^\dagger$LingBot-VLA v2 on FT-YAM-02 was stopped after $n{=}1$ trial, and GR00T N1.7 on FT-YAM-03 was stopped after $n{=}9$ trials, for safety.}
    \label{fig:task_progress_reachability}
\end{figure}

\paragraph{Policies differ in the extent of partial task completion.}
\Cref{fig:task_progress_reachability} reports the percentage of trials satisfying at least a given fraction of task criteria, averaged equally across the five tasks of each embodiment. On DROID/Franka, G0.5 satisfies at least one-quarter of the criteria in nearly all trials, but reaches three-quarters in only about 20\%, with 16\% full-task success. In contrast, $\pi_{0.5}$ maintains high criterion coverage across all thresholds, achieving 84\% full-task success. On Bimanual-YAM, MolmoAct2 satisfies at least one-quarter of the criteria in nearly all trials but completes only about half, whereas $\pi_{0.5}$ achieves 93.6\% success. The limitations are more pronounced on Unitree G1, where no policy successfully completes any task, and fewer than 40\% of trials satisfy even one-quarter of the criteria. These results show that incomplete executions can reflect different levels of task progress: some policies frequently satisfy intermediate requirements without completing the task, while others struggle to satisfy even a modest fraction of the evaluated criteria. Per-task curves for all 20 \texttt{Fine-tune Track} tasks are provided in \Cref{sec:appendix_progress_reachability}.

\begin{figure}[t]
    \centering
    \includegraphics[width=\textwidth]{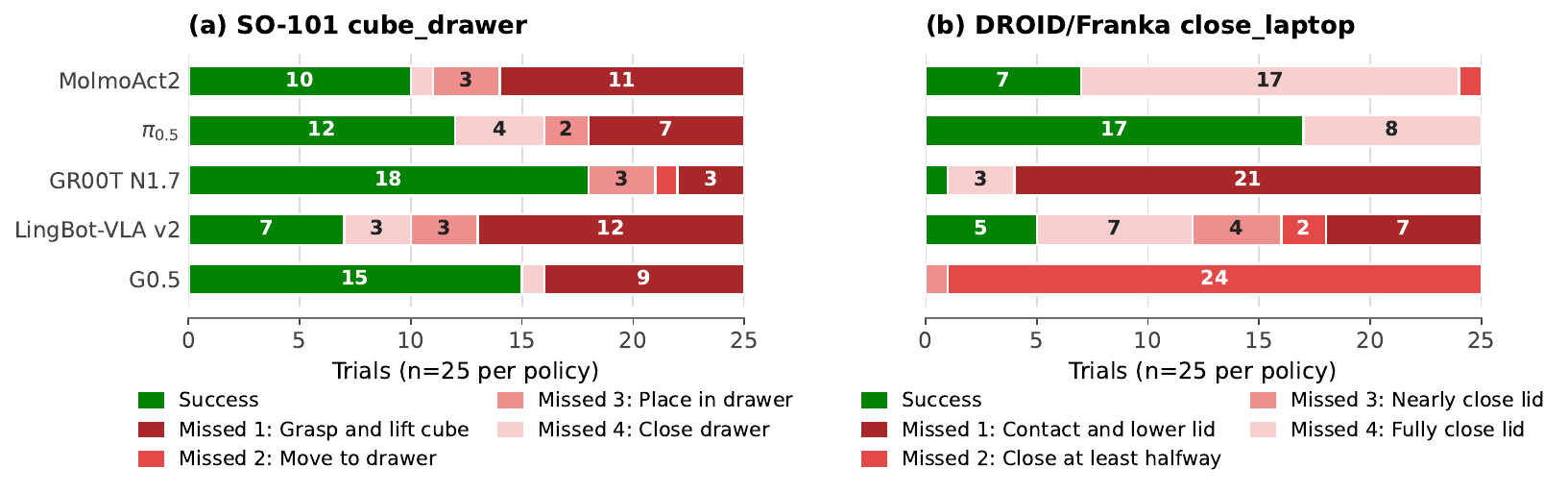}
    \caption{\textbf{Policy-specific failure stages after fine-tuning.}
We show two tasks whose progress criteria form an ordered sequence: SO-101 \texttt{cube\_drawer} and DROID/Franka \texttt{close\_laptop}. Each bar divides a policy's 25 evaluation trials into full-task successes (green) and failures, shaded by the first criterion the trial did not meet (darker red indicates an earlier failure). Numbers give trial counts; segments of a single trial are unlabeled. Policies trained on the same demonstrations exhibit markedly different failure locations, revealing task- and policy-specific execution bottlenecks that are hidden by success rate alone.}
    \label{fig:rq2_milestone_failures}
\end{figure}

\paragraph{Task failures reveal both shared and policy-specific limitations.}
To better understand where policies struggle during task execution, \Cref{fig:rq2_milestone_failures} groups unsuccessful trials by the earliest unmet criterion in two representative manipulation tasks. On SO-101 \texttt{cube\_drawer}, the initial grasp-and-lift criterion is unmet in 12--48\% of trials across policies, revealing a common difficulty whose frequency varies across models. On DROID/Franka \texttt{close\_laptop}, policies exhibit more distinct failure patterns. G0.5 satisfies the initial contact-and-lowering criterion in all 25 trials but fails to reach halfway closure in 24. In contrast, MolmoAct2 nearly closes the lid in 24 of 25 trials, yet completes the task in only seven. Similarly, $\pi_{0.5}$ reaches near closure in all 25 trials but successfully completes 17. These results show that policies can struggle with different requirements even on the same task, ranging from advancing beyond initial contact to satisfying the final completion condition. Even when fine-tuned on identical demonstrations, policies exhibit both shared difficulties and distinct execution limitations, suggesting that improving task completion may require addressing different weaknesses across models.

\begin{takeaway}
Pretrained robot policies exhibit varying degrees of adaptability across manipulation tasks. While some can be successfully fine-tuned across diverse tasks, others achieve strong performance only on selected tasks, despite receiving the same demonstration budget for each task. Beyond task success, policies also differ in how much partial progress they achieve and which task requirements they struggle to satisfy, revealing both shared and policy-specific execution limitations. Together, these findings suggest that current policies exhibit uneven adaptability, with differences not only in the breadth of tasks they can learn but also in the manipulation capabilities they acquire through fine-tuning.
\end{takeaway}




\subsection{Performance Consistency (RQ3)}
\label{sec:performance_consistency}

\begin{figure}[t]
    \centering
    \includegraphics[width=\linewidth]{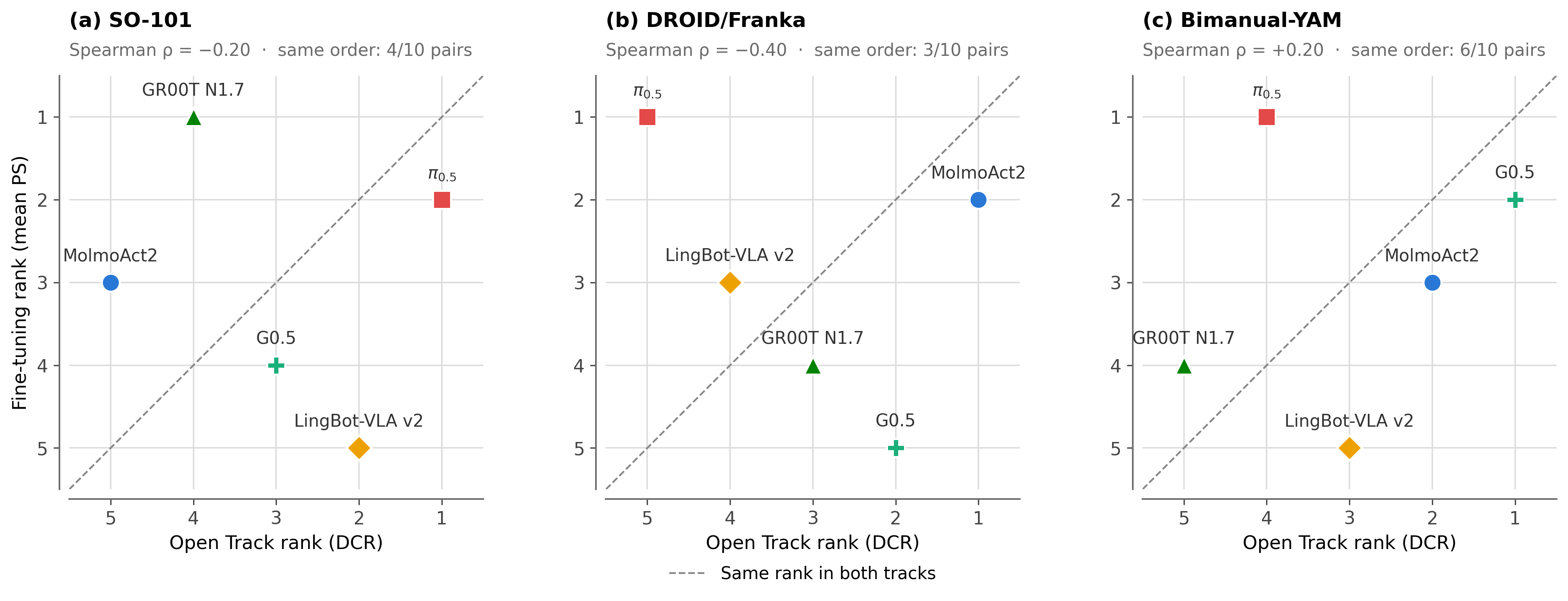}
    \caption{\textbf{Cross-track ranking consistency by embodiment.}
    Open Track and Fine-tune Track rankings of the five VLA policies on (a) SO-101, (b) DROID/Franka, and (c) Bimanual-YAM. For each embodiment, the Open Track rank is computed from the DCR rating, while the Fine-tuning rank is computed from the mean progress score (PS) across its five tasks. Rank 1 denotes the best-performing model, and the dashed diagonal indicates identical rankings across the two tracks. Each panel reports Spearman's rank correlation and the number of model pairs with the same relative ordering in both tracks.}
    \label{fig:cross_track_ranking}
\end{figure}

\paragraph{Strong open-world performance does not consistently predict post-fine-tuning performance.}
\Cref{fig:cross_track_ranking} compares the relative rankings of five VLA policies between the \texttt{Open Track} and \texttt{Fine-tune Track}. The rankings show limited agreement across all three shared embodiments, with Spearman's $\rho=-0.20$ on SO-101, $-0.40$ on DROID/Franka, and $0.20$ on Bimanual-YAM. Only 4, 3, and 6 of the 10 model pairs, respectively, preserve their relative ordering between tracks. For example, $\pi_{0.5}$ ranks fifth among VLA policies on DROID/Franka in the \texttt{Open Track} but first after fine-tuning, and similarly moves from fourth to first on Bimanual-YAM. Conversely, G0.5 ranks second on DROID/Franka in the \texttt{Open Track} but fifth after fine-tuning. These comparisons involve only five policies, and the tracks differ in task distributions, starting checkpoints, and ranking metrics, so the disagreement cannot be attributed solely to differences in generalization and adaptation capabilities. Nevertheless, the observed rank reversals show that a policy's relative strength in open-world execution does not necessarily carry over to its performance after task-specific adaptation.

\begin{figure*}[t]
    \centering

    \begin{minipage}[c]{0.49\textwidth}
        \centering
        \includegraphics[height=5.0cm,keepaspectratio]{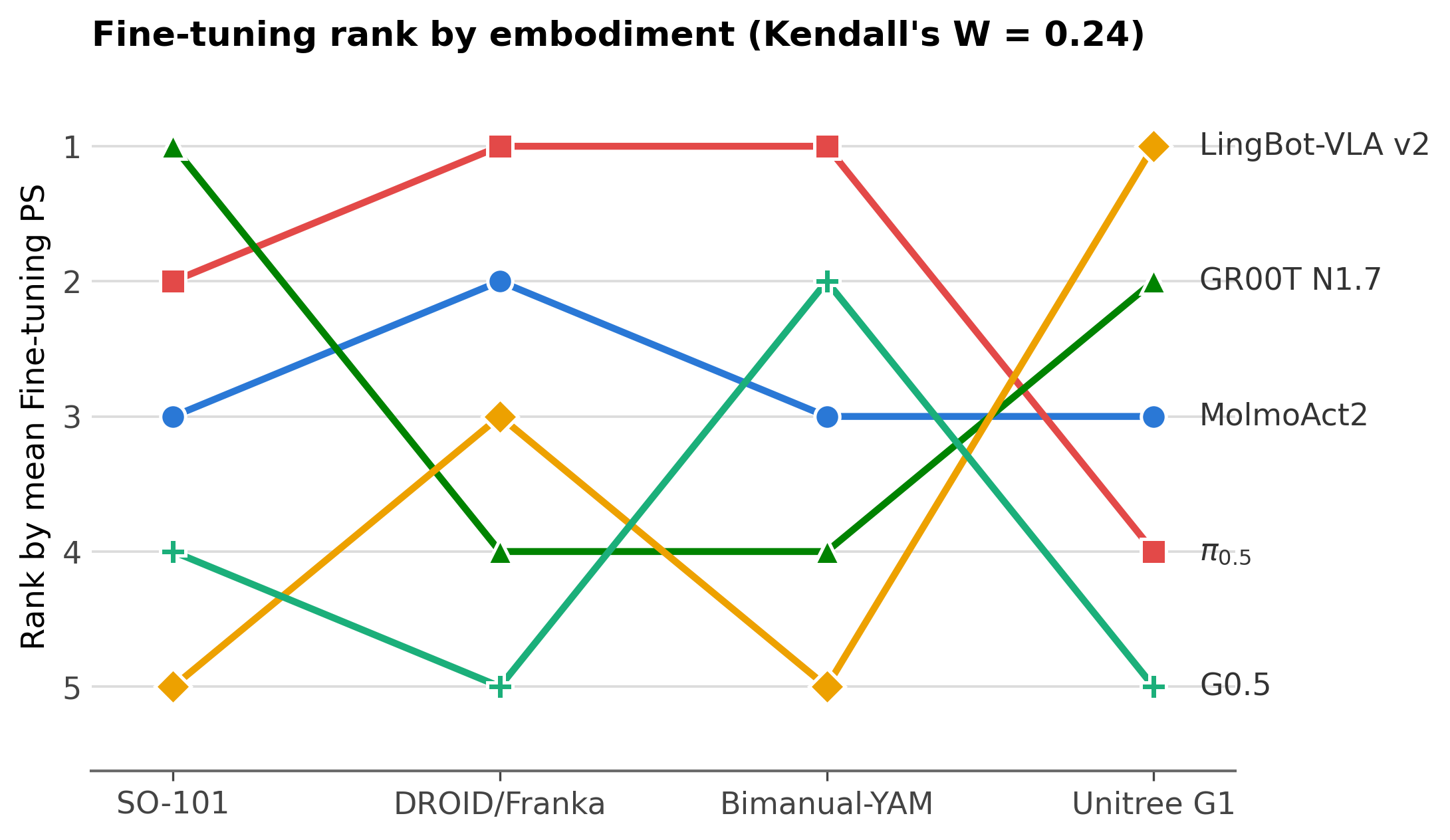}
        \subcaption{\textbf{Cross-embodiment ranking consistency}}
        \label{fig:cross_embodiment_ranking}
    \end{minipage}
    \hfill
    \begin{minipage}[c]{0.49\textwidth}
        \centering
        \includegraphics[height=5.0cm,keepaspectratio]{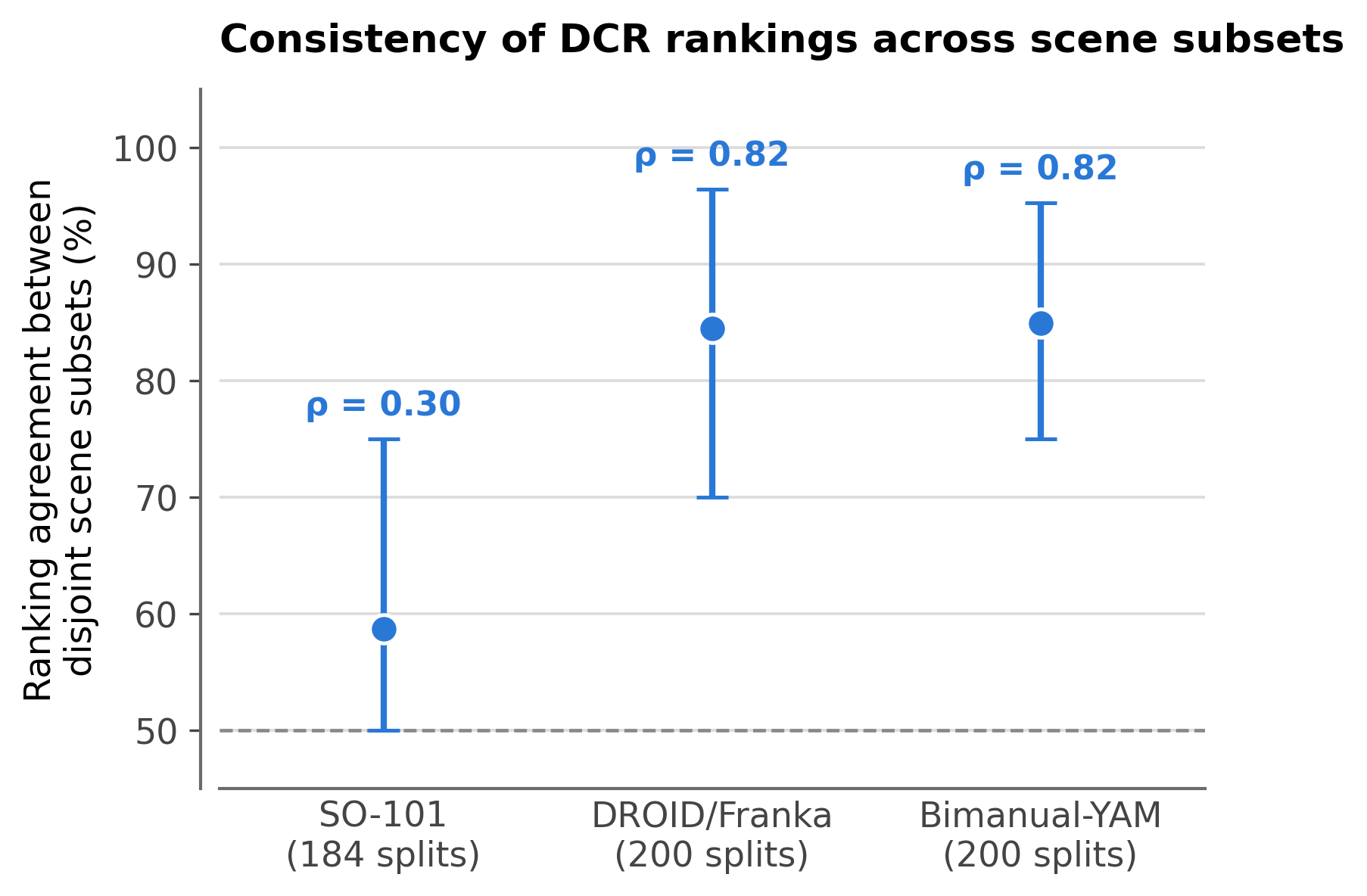}
        \subcaption{\textbf{Consistency of DCR rankings across subsets}}
        \label{fig:site_split_ranking_reliability}
    \end{minipage}

    \caption{\textbf{Ranking consistency.}
    \textbf{a.} Cross-embodiment ranking consistency based on Fine-tune Track progress score (PS) across SO-101, DROID, Bimanual-YAM, and Unitree G1. Rank 1 denotes the best-performing model on an embodiment. The substantial changes in model ordering across embodiments, together with Kendall's coefficient of concordance $W=0.24$, indicate weak agreement between embodiment-specific rankings.
    \textbf{b.} Consistency of DCR rankings across independent evaluation-site subsets. For each embodiment, evaluation sites are repeatedly divided into two disjoint subsets, and DCR is fit independently on each half. Points show the mean fraction of model pairs whose relative ordering agrees between the two rankings, and error bars show the 2.5th--97.5th percentile across valid random splits. The annotated $\rho$ is the mean Spearman rank correlation between the two rankings. Higher values indicate that the leaderboard is less sensitive to which subset of evaluation sites contributes the data.}
    \label{fig:ranking_consistency}
\end{figure*}

\paragraph{Relative policy strengths vary across robot embodiments and task settings.}
\Cref{fig:open_track_dcr_by_embodiment,fig:cross_embodiment_ranking} show that policy rankings vary across robot embodiments in both evaluation tracks. In the \texttt{Open Track}, the relative ordering of VLA policies changes across SO-101, DROID/Franka, and Bimanual-YAM, with different policies achieving the highest DCR ratings on each embodiment. Similar differences emerge in the \texttt{Fine-tune Track}, where Kendall's coefficient of concordance ($W=0.24$) indicates limited agreement in policy rankings across the four embodiment-specific task suites. For example, $\pi_{0.5}$ ranks first on both DROID/Franka and Bimanual-YAM but fourth on Unitree G1, whereas GR00T N1.7 leads on SO-101 but ranks fourth on DROID/Franka and Bimanual-YAM. Because each embodiment involves different tasks and environments, these differences cannot be attributed to robot morphology alone. Together, these findings show that relative policy strengths do not consistently carry across robot-task settings, whether evaluated without task-specific adaptation or after fine-tuning.


\begin{takeaway}
Relative policy strengths vary across both evaluation tracks and robot-task settings. Policies that perform well without task-specific adaptation do not necessarily achieve similarly strong results after fine-tuning, and those that lead on one embodiment may perform less competitively on others. Together, these findings show that current general-purpose robot policies exhibit limited consistency in their relative performance across different evaluation conditions, with no single policy maintaining a clear advantage across all settings.
\end{takeaway}

\subsection{Ranking Reliability and Efficiency (RQ4)}



\paragraph{Distributed pairwise evaluations produce broadly consistent rankings across independent site subsets.}
\Cref{fig:site_split_ranking_reliability} tests whether independent subsets of the distributed evaluation network produce the same DCR ranking. For each embodiment, we repeatedly partition the set of evaluation sites at random into two approximately equal, non-overlapping groups, assigning all evaluations from the same site to the same group, then fit DCR independently on each subset, and measure the fraction of model pairs whose relative ordering is preserved. DROID/Franka and Bimanual-YAM achieve mean pairwise ranking agreement of approximately 85\%, with mean Spearman correlations of $\rho=0.82$, indicating that substantially different subsets of sites recover similar model orderings. In contrast, SO-101 achieves only 59\% agreement and $\rho=0.30$, showing that its ranking is considerably more sensitive to which evaluation sites are included. These results indicate that distributed pairwise evaluation can produce reproducible rankings when sufficient evidence is available, but reliability is not uniform across embodiments.

\begin{figure*}[t]
    \centering

    \begin{minipage}[c]{0.49\textwidth}
        \centering
        \includegraphics[
            height=5.0cm,
            keepaspectratio
        ]{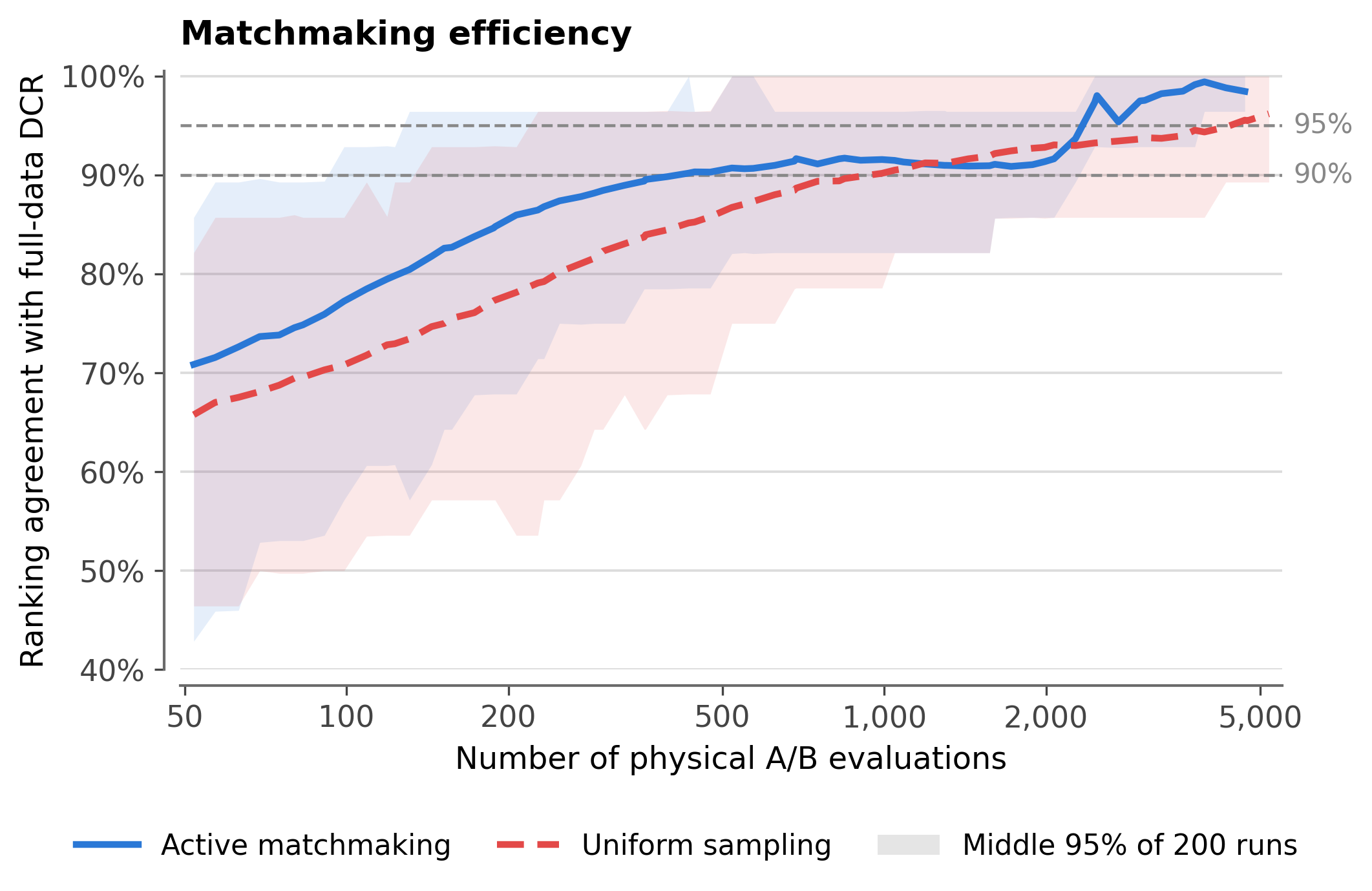}
        \subcaption{\textbf{Active matchmaking improves ranking efficiency}}
        \label{fig:matchmaking_efficiency}
    \end{minipage}
    \hfill
    \begin{minipage}[c]{0.49\textwidth}
        \centering
        \includegraphics[
            height=5.0cm,
            keepaspectratio
        ]{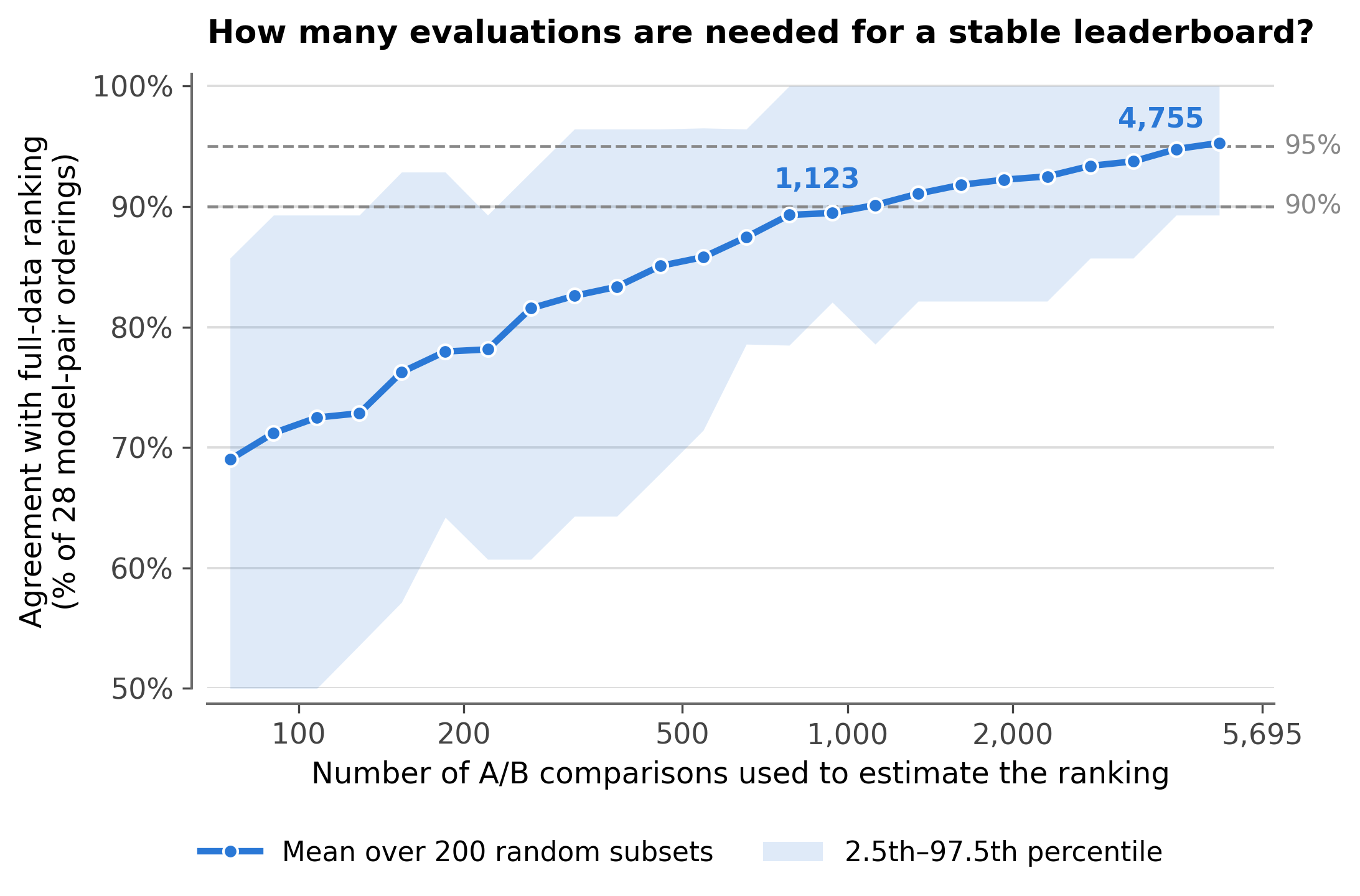}
        \subcaption{\textbf{Evaluation scale on Open Track ranking stability}}
        \label{fig:evaluation_scale_dcr}
    \end{minipage}

    \caption{\textbf{Ranking efficiency and stability under limited evaluation budgets.}
    \textbf{a.} Active matchmaking improves ranking efficiency. Ranking agreement with the full-data DCR leaderboard is shown as a function of the number of physical A/B evaluations. Active matchmaking prioritizes uncertain model pairs, while uniform sampling selects pairs without uncertainty-based allocation. Lines show the mean agreement over 200 independent runs, and shaded regions span the 2.5th--97.5th percentiles.
    \textbf{b.} Effect of evaluation scale on Open Track ranking stability. At each evaluation budget, 200 random subsets containing all eight systems are sampled, DCR is fit using only those evaluations, and the resulting ranking is compared with the DCR ranking obtained from the full 5,699-evaluation dataset. The line shows the mean agreement across subsets, the shaded region spans the 2.5th--97.5th percentiles, and dashed lines mark 90\% and 95\% mean ranking agreement.}
    \label{fig:ranking_efficiency_stability}
\end{figure*}

\paragraph{Active matchmaking recovers stable rankings with substantially fewer physical evaluations.}
\Cref{fig:matchmaking_efficiency} compares uncertainty-driven active matchmaking with uniform sampling using the same DCR ranking estimator. Active matchmaking reaches approximately 90\% agreement with the full-data ranking after only 400--500 evaluations, whereas uniform sampling requires roughly 1,000 evaluations to achieve comparable agreement. The advantage is largest at limited evaluation budgets and diminishes as both strategies approach the full-data ranking, showing that active matchmaking improves sample efficiency without changing the underlying ranking objective.

\begin{takeaway}
Distributed pairwise evaluation can recover reproducible model rankings across independent subsets of evaluation sites, although reliability depends on the amount and distribution of evidence available for each embodiment. Active matchmaking further improves efficiency by directing physical trials toward the comparisons that are most informative for resolving the leaderboard.
\end{takeaway}

\subsection{Effect of scaling evaluation (RQ5)}
\label{sec:effect_scaling}

\paragraph{Leaderboard stability improves steadily with evaluation scale, but high-confidence stability requires substantially more data.}
\Cref{fig:evaluation_scale_dcr} measures how closely DCR rankings estimated from random subsets of the Open Track data match the ranking obtained from the full dataset. As the number of physical A/B comparisons increases, the mean agreement in model-pair ordering rises steadily, reaching 90\% at approximately 1,123 evaluations and 95\% at approximately 4,755 evaluations. However, the 2.5th--97.5th percentile range remains substantially wider than the mean trend, showing that different subsets of the same size can still produce noticeably different rankings. Thus, relatively few evaluations are sufficient to recover most of the final ordering on average, while substantially more evaluations are required for the leaderboard to become consistently stable across different subsets of the data.

\begin{takeaway}
More physical evaluations improve both ranking stability and confidence in the Open Track leaderboard. The average model ordering emerges relatively early, but establishing a ranking that is robust to which evaluations are included requires a much larger evaluation budget.
\end{takeaway}

\section{Limitations \& Conclusion}
\label{sec:conclusion}

\paragraph{Conclusion.}
RoboColosseum provides a distributed framework for measuring physical intelligence across real robots, tasks, environments, and evaluation settings. Our results show that no single evaluation captures model capability completely: frontier AI agents can be competitive with specialized VLA policies without task-specific adaptation, fine-tuned policies exhibit strongly task- and embodiment-dependent strengths, and relative model rankings can change across both evaluation tracks and robot platforms. Progress-based evaluation further reveals meaningful behavioral differences that binary task success alone can hide.

At the same time, large-scale pairwise evaluation makes these comparisons increasingly reproducible, while active matchmaking can concentrate costly physical trials on the comparisons that are most informative for resolving the leaderboard. Together, these results suggest that evaluating physical intelligence requires not only more capable models, but also sufficiently diverse, large-scale, and statistically grounded real-world evaluation. \texttt{RoboColosseum} provides the infrastructure and protocols for carrying out such evaluation as the model, task, embodiment, and site coverage continue to grow.

\paragraph{Limitations.}
The current study represents an initial snapshot of \texttt{RoboColosseum} rather than an exhaustive evaluation of all robot foundation models and deployment settings. Coverage differs across model families and embodiments because participation depends on checkpoint and hardware compatibility, while the \texttt{Fine-tune Track} currently evaluates five tasks per embodiment. Future evaluations can naturally extend the benchmark with additional models, tasks, embodiments, and sites while retaining the same protocols.

In the \texttt{Fine-tune Track}, we fix the task-specific demonstrations and evaluation procedure but allow each policy to use its official adaptation recipe. The resulting comparison therefore measures practical end-to-end adaptation from the same data rather than isolating architectural differences under an identical optimizer and training schedule. In the \texttt{Open Track}, our ranking-stability analyses quantify reproducibility relative to the leaderboard estimated from the complete collected dataset; as the benchmark grows, the same analyses can be repeated to measure how rankings evolve with broader evaluation coverage.

\section{The Future}
\label{sec:future}
\texttt{RoboColosseum} is a long-term scientific megaproject designed to help the robotics community distinguish meaningful progress from noise through standardized evaluation protocols and large-scale, in-the-wild model deployment and testing. Its founding members; NUS MAGIC Lab, Hugging Face, the Singapore Embodied AI Centre, and BitRobot are committed to advancing this effort and enabling new scientific insights through real-world evaluation at scale. We pledge to expand the task datasets available for the fine-tuning track, provide access to more in-the-wild environments for open-track evaluation, and incorporate a broader range of robotic embodiments. To support broad academic participation, we are providing a total of 10,000 hours of free evaluation to be shared across all academic partners.

\section*{Acknowledgments}
\label{sec:ack}
We would like to thank the National University of Singapore’s School of Computing for its administrative support, and the Singapore Embodied AI Centre for its support. This project is funded by the National Robotics Programme 2.0 grant, “Real-World Robot Data Collection and Evaluation: A Singapore Dataset and Benchmark for Embodied AI.”

\bibliographystyle{abbrvnat}
\bibliography{reference}

\clearpage
\beginappendix


\section{Fine-tune Track Training Protocol}
\label{app:finetuning_protocol}

All \texttt{Fine-tune Track} models are trained with a common harness designed for a fair comparison under fixed data. Each VLA is trained with its own upstream repository, pinned to a fixed commit and left unmodified. The harness adds only the components around it: conversion of the demonstrations into each model's data format, per-model configurations registered from outside the repositories, and launch scripts. Settings shared by all models are listed in \Cref{tab:ft_shared}, and the per-model hyperparameters in \Cref{tab:ft_hparams}.

\begin{table}[h]
\centering
\caption{\textbf{Fine-tuning settings shared by all models.}}
\label{tab:ft_shared}
\small
\renewcommand{\arraystretch}{1.15}
\begin{tabular}{@{}ll@{}}
\toprule
\textbf{Setting} & \textbf{Value} \\
\midrule
Initialization & Base pretrained checkpoint (not the \texttt{Open Track} checkpoint) \\
Training data & 100 task demonstrations only \\
Training recipe & Closest example configuration in the official repository \\
Learning-rate scaling & Linear in effective batch size, relative to the recipe's batch size \\
Evaluated checkpoint & Final epoch \\
State and action space & Absolute joint positions \\
Model inputs & We follow each model's default training recipe for visual observations and task instruction \\
Precision & bf16 compute, fp32 master weights \\
\bottomrule
\end{tabular}
\end{table}

\begin{table}[!ht]
\centering
\caption{\textbf{Per-model fine-tuning hyperparameters.} Peak learning rates are after linear scaling from the recipe's batch size to the effective batch size. ``Upstream default'' denotes the value in the cited recipe, left unchanged.}
\label{tab:ft_hparams}
\footnotesize
\renewcommand{\arraystretch}{1.15}
\newcommand{\hptable}[2]{%
  \begin{tabular}{@{}>{\raggedright\arraybackslash}p{0.33\linewidth}>{\raggedright\arraybackslash}p{0.60\linewidth}@{}}
  \toprule
  \multicolumn{2}{@{}l}{\rule{0pt}{2.4ex}\textbf{#1}} \\
  \midrule
  #2
  \bottomrule
  \end{tabular}}
\begin{subtable}[t]{0.485\linewidth}
\hptable{GR00T N1.7}{
Base checkpoint & \texttt{nvidia/GR00T-N1.7-3B} \\
Recipe & \texttt{launch\_finetune} defaults \\
Recipe batch size & 64 \\
Effective batch size & 64 \\
Peak learning rate & $1\times10^{-4}$ \\
Schedule & Cosine, 5\% warmup \\
Trained modules & Full model (VLM and action head) \\
}
\end{subtable}\hfill
\begin{subtable}[t]{0.485\linewidth}
\hptable{\smash{\boldmath$\pi_{0.5}$}}{
Base checkpoint & \texttt{pi05\_base} \\
Recipe & \texttt{pi05\_aloha\_pen\_uncap} \\
Recipe batch size & 64 \\
Effective batch size & 64 \\
Peak learning rate & $2.5\times10^{-5}$ \\
Schedule & Cosine to $2.5\times10^{-6}$, 1k warmup \\
Other & EMA 0.99; Aloha adaptation off \\
}
\end{subtable}

\vspace{0.8em}
\begin{subtable}[t]{0.485\linewidth}
\hptable{MolmoAct2}{
Base checkpoint & \texttt{allenai/MolmoAct2} \\
Recipe & Full fine-tuning (official README) \\
Recipe batch size & 64 \\
Effective batch size & 128 \\
Peak learning rate & LLM $2\times10^{-5}$; ViT and connector $1\times10^{-5}$; action expert $1\times10^{-4}$ \\
Schedule & Upstream default \\
Trained modules & Full model \\
}
\end{subtable}\hfill
\begin{subtable}[t]{0.485\linewidth}
\hptable{G0.5}{
Base checkpoint & \texttt{OpenGalaxea/G05} (base) \\
Recipe & R1Lite post-training \\
Recipe batch size & 32 \\
Effective batch size & 32 \\
Peak learning rate & $4\times10^{-5}$ (vision encoder $\times0.1$) \\
Schedule & Cosine, 200 warmup steps \\
Other & Weight decay 0.03 \\
}
\end{subtable}

\vspace{0.8em}
\begin{subtable}[t]{0.485\linewidth}
\hptable{LingBot-VLA v2}{
Base checkpoint & \texttt{robbyant/lingbot-vla-v2-6b} \\
Recipe & \texttt{real\_robot.yaml} \\
Recipe batch size & 256 \\
Effective batch size & 64 \\
Peak learning rate & $1.25\times10^{-5}$ \\
Schedule & Constant \\
Optimizer & Muon \\
}
\end{subtable}\hfill
\begin{subtable}[t]{0.485\linewidth}
\mbox{}
\end{subtable}
\end{table}

\paragraph{Model-specific notes.}
For GR00T N1.7, the full vision-language backbone is fine-tuned together with the action head, and the effective batch size is kept at 64. An earlier configuration that trained only the action head with an effective batch size of 512 received too few optimizer updates within five epochs and converged to a constant prediction. SO-101 is an exception; we follow the default configuration, keeping the vision-language backbone frozen while training the projector and action-model components. For $\pi_{0.5}$, the Aloha-specific action-space adaptation in the recipe is disabled, since none of the evaluated embodiments is an Aloha platform.

\FloatBarrier

%

\providecommand{\realtaskfigdir}{images/fineturning_track_tasks}

\definecolor{RTBlue}{HTML}{1c6ef3}
\definecolor{RTBlueLight}{HTML}{d2e3fc}


\newcommand{\rtphoto}[2]{%
  \begin{tikzpicture}
    \useasboundingbox (0,0) rectangle (\linewidth,0.75\linewidth);
    \begin{scope}
      \clip[rounded corners=3pt] (0,0) rectangle (\linewidth,0.75\linewidth);
      \node[anchor=south west, inner sep=0pt, outer sep=0pt] at (0,0)
        {\includegraphics[width=\linewidth,height=0.75\linewidth]{#1}};
    \end{scope}
    \node[anchor=south west, inner sep=1.8pt, rounded corners=1.5pt,
          fill=black, fill opacity=0.6, text opacity=1, text=white,
          font=\tiny\bfseries] at (3pt,3pt) {#2};
  \end{tikzpicture}%
}

\newcommand{\rtchip}[1]{%
  \tikz[baseline=(c.base)]\node(c)[fill=black!6, rounded corners=1.5pt,
    inner xsep=3pt, inner ysep=1.5pt, font=\scriptsize\ttfamily, text=black!70]{#1};%
}

\newcommand{\rtfield}[2]{\noindent\textbf{#1}: #2\par\vspace{0.03em}}
\newcommand{\rtinline}[2]{\textbf{#1}: #2}
\newcommand{\rtsep}{\hspace{1.6em}}
\newcommand{\rtmeta}[1]{\noindent #1\par\vspace{0.03em}}

\newenvironment{rtprogress}{%
  \par\vspace{0.12em}\noindent\textbf{Partial Progress Criteria}:\par\vspace{0.06em}%
  \footnotesize\setlength{\baselineskip}{0.95\baselineskip}%
}{\par}
\newcommand{\rtscore}[2]{%
  \noindent\hangindent=1.6em\hangafter=1
  \makebox[1.6em][l]{\tikz[baseline=(s.base)]\node(s)[fill=RTBlueLight,
    rounded corners=1.5pt, inner sep=1.2pt, minimum width=1.15em,
    font=\scriptsize\bfseries, text=RTBlue]{#1};}#2\par\vspace{0.04em}%
}

\newcommand{\realtaskcard}[5]{%
\begin{tcolorbox}[colback=white, colframe=black!12, boxrule=0.45pt, arc=1.5pt,
    left=5pt, right=5pt, top=5pt, bottom=5pt, boxsep=0pt,
    before skip=0.4em, after skip=0.4em]
\noindent
\begin{minipage}[t]{0.26\linewidth}\vspace{0pt}\offinterlineskip
  \rtphoto{#3}{Workspace Setup}\par\vspace{4pt}
  \rtphoto{#4}{Object Setup}
\end{minipage}\hfill
\begin{minipage}[t]{0.718\linewidth}\vspace{0pt}
  {\normalsize\bfseries\textcolor{RTBlue}{#2}}\hfill\rtchip{#1}\par\vspace{0.3em}
  {\small\setlength{\parskip}{0pt}\setlength{\baselineskip}{0.95\baselineskip}#5}
\end{minipage}
\end{tcolorbox}%
}

\section{Fine-tune Track Task Details}
\label{app:finetuning_track_tasks}

We provide the full specification of the five fine-tuning tasks for each of the four embodiments (SO101, DROID, YAM, and G1). Each card shows the workspace and object setup photos together with the task instruction, objects used, object setup, success criteria, evaluation location, environment, timeout, and the partial progress criteria used for scoring.

\subsection{SO101}
\label{app:finetuning_track_so101}

\realtaskcard
{FT-SO101-01}
{stack\_bowls}
{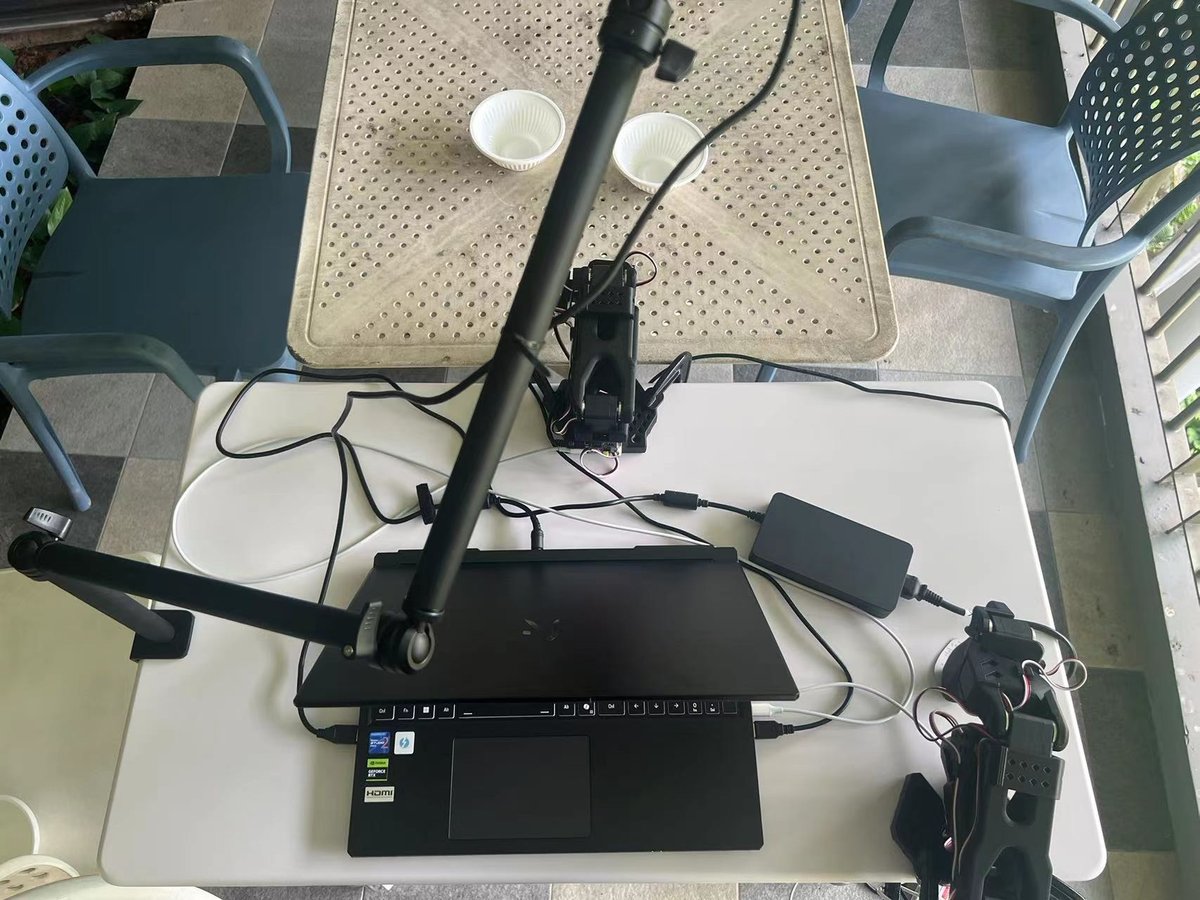}
{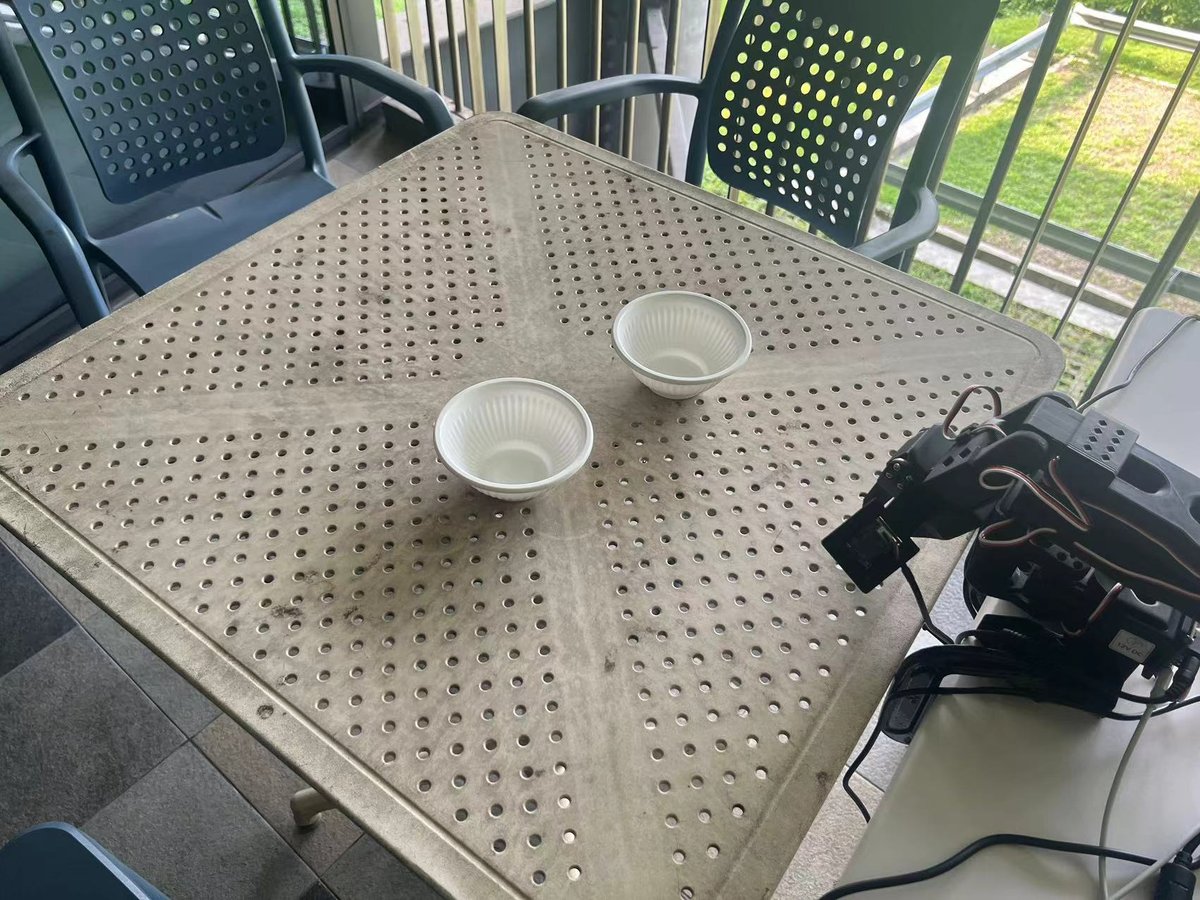}
{
\rtfield{Instruction}{Pick up the white plastic bowl on the right and stack it on top of the white plastic bowl on the left.}
\rtfield{Objects Used}{2 bowls}
\rtfield{Object Setup}{2 bowls on table, one left one right.}
\rtfield{Success Criteria}{The bowl on the right was successfully stacked onto the left one.}
\rtfield{Location}{National University of Singapore, place nearby COM3-B1-16}
\rtfield{Address}{National University of Singapore, 11 Research Link, COM3, \#B1-16, Singapore 119391}
\rtmeta{\rtinline{Environment}{Semi-open terrace}\rtsep\rtinline{Time}{3 PM}\rtsep\rtinline{Timeout}{25 s}}
\begin{rtprogress}
  \rtscore{0}{No meaningful progress.}
  \rtscore{1}{Right bowl is successfully grasped and lifted.}
  \rtscore{2}{Right bowl is moved above or close to the left bowl.}
  \rtscore{3}{Right bowl is placed on the left bowl but is unstable, misaligned, or not fully released.}
  \rtscore{4}{Right bowl is successfully stacked on the left bowl, stable, and released by the robot.}
\end{rtprogress}
}

\realtaskcard
{FT-SO101-02}
{remove\_screwdriver}
{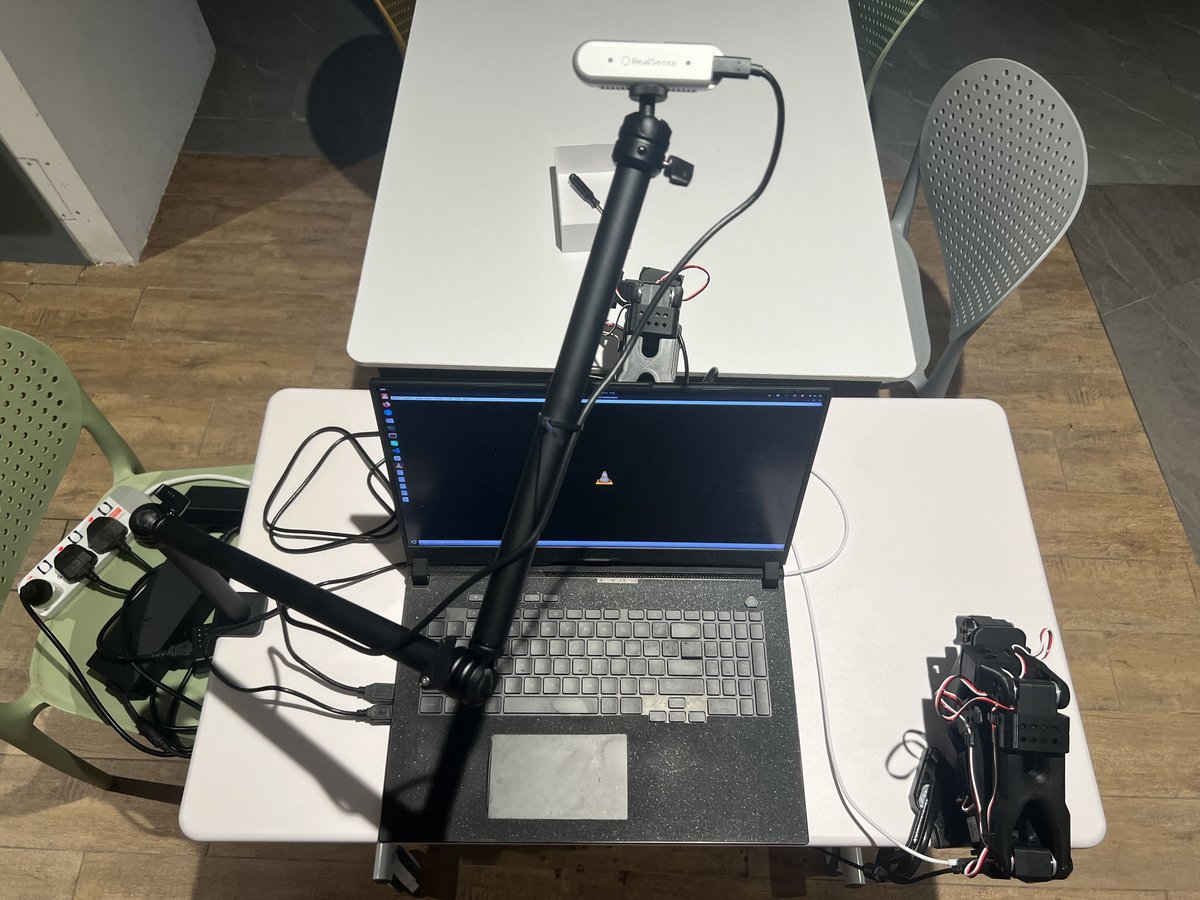}
{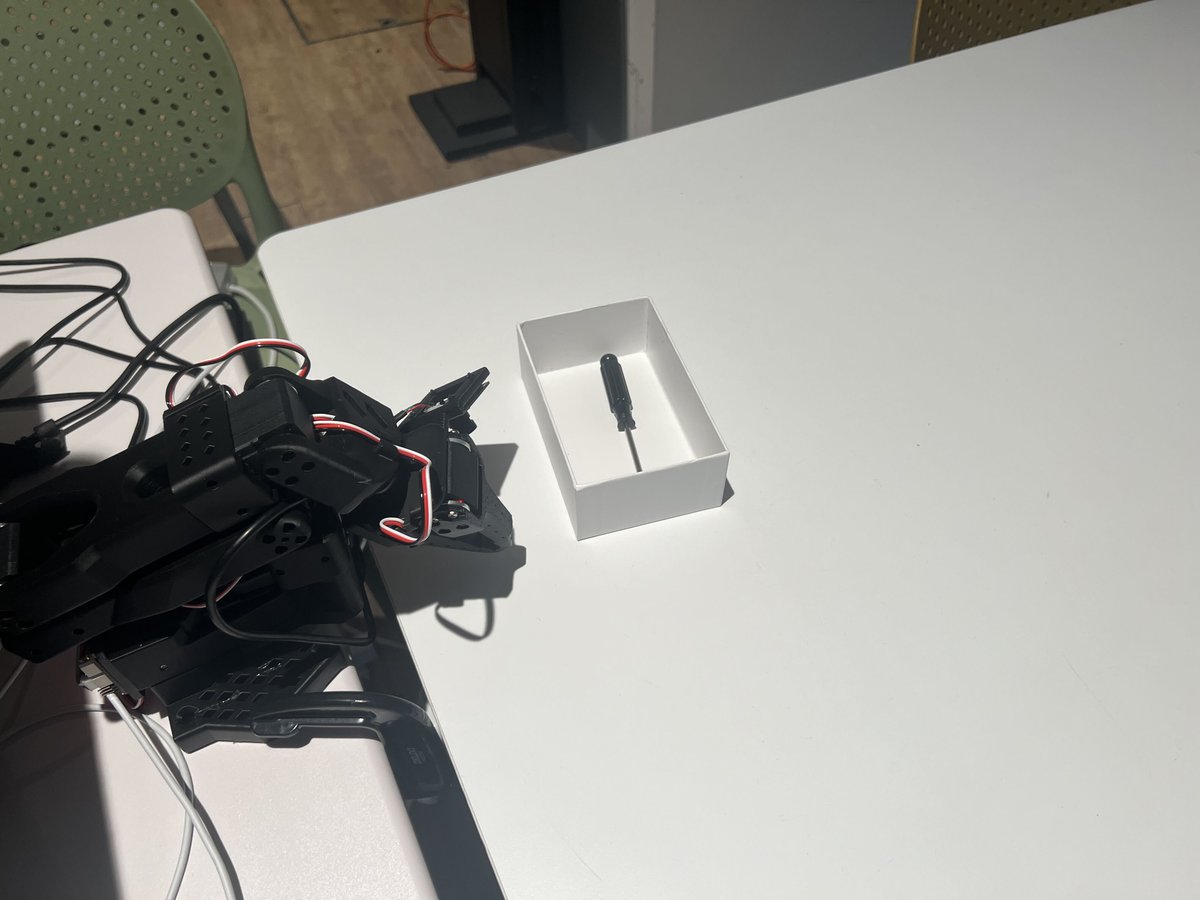}
{
\rtfield{Instruction}{Take the black screwdriver out of the box and place it on the table.}
\rtfield{Objects Used}{1 white box; 1 screwdriver}
\rtfield{Object Setup}{1 white box on table, with 1 screwdriver inside it. The box is placed horizontally, the screwdriver is also horizontal.}
\rtfield{Success Criteria}{Remove the screwdriver from the box and place it on the desktop on the right side of the box.}
\rtfield{Location}{National University of Singapore, a table at The Terrace}
\rtfield{Address}{The Terrace, 100 Edgedale Plains, Singapore 828690, Singapore}
\rtmeta{\rtinline{Environment}{Dining hall}\rtsep\rtinline{Time}{3 PM}\rtsep\rtinline{Timeout}{25 s}}
\begin{rtprogress}
  \rtscore{0}{No meaningful progress.}
  \rtscore{1}{Successfully approached the screwdriver.}
  \rtscore{2}{Successfully attempted grasping screwdriver (grasped/with contact but failed).}
  \rtscore{3}{Successfully grasped the screwdriver.}
  \rtscore{4}{Screwdriver is completely removed from the box and placed on the table on the right side of the box, stable and released.}
\end{rtprogress}
}

\realtaskcard
{FT-SO101-03}
{upright\_bottle}
{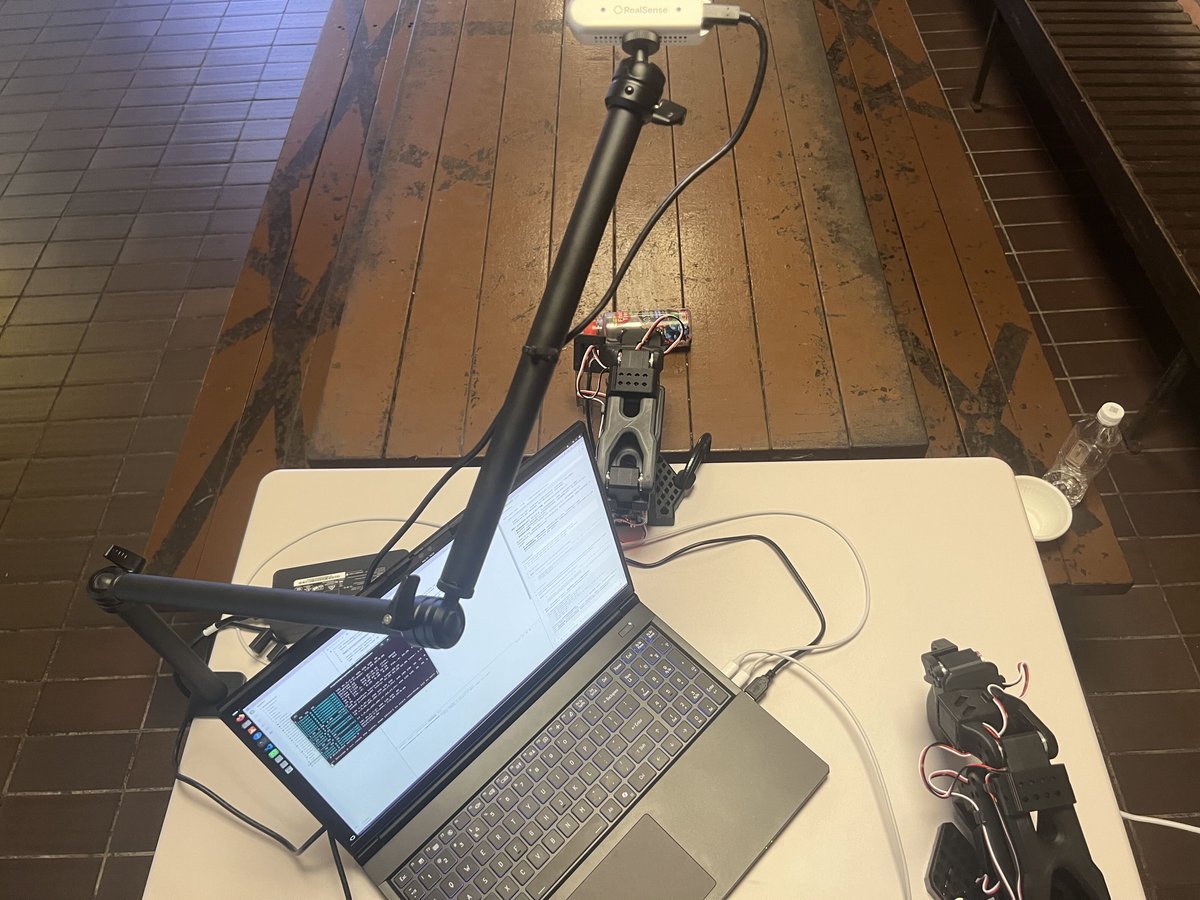}
{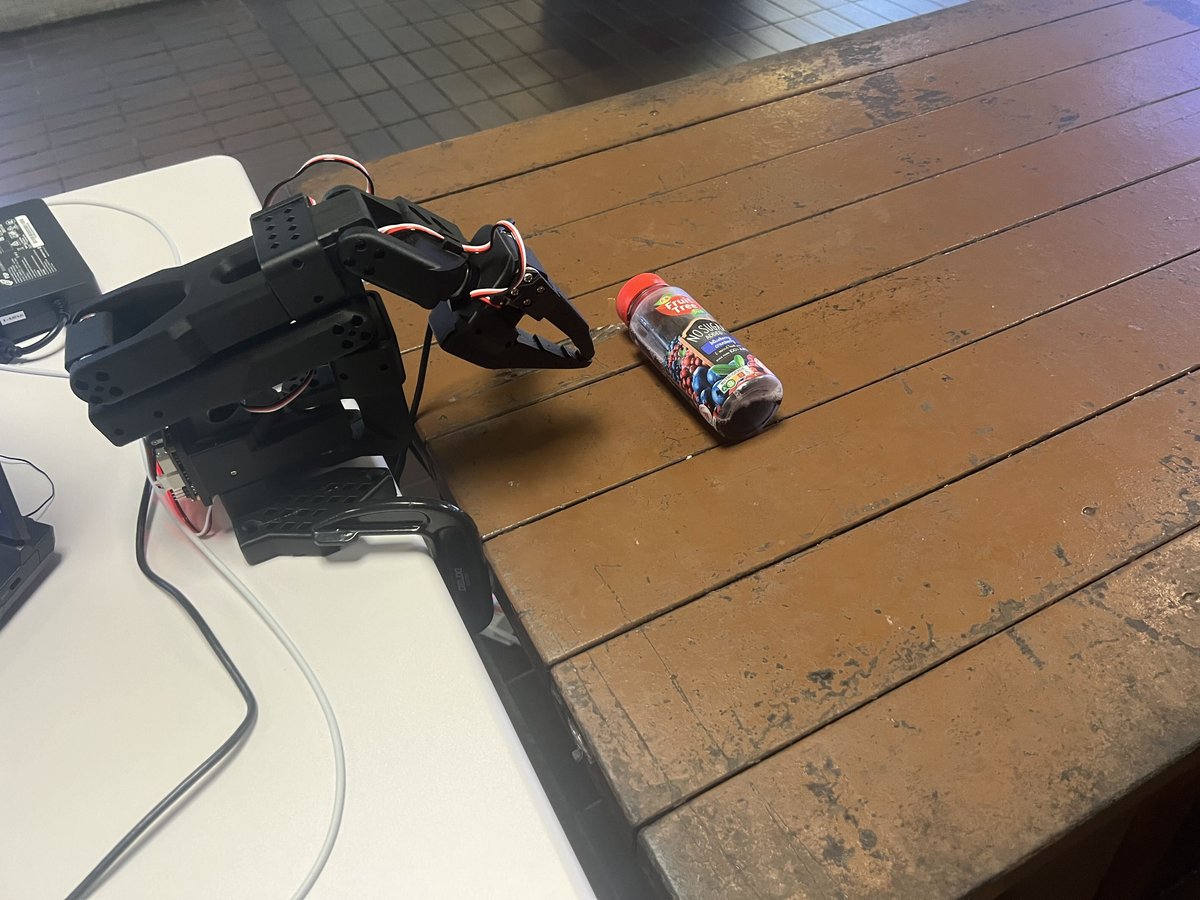}
{
\rtfield{Instruction}{Pick up the fallen juice bottle and stand it upright on the table.}
\rtfield{Objects Used}{1 juice bottle}
\rtfield{Object Setup}{The juice bottle is lying on its side on the tabletop within the robot's reachable workspace.}
\rtfield{Success Criteria}{The juice bottle is upright, stable on its base, and released by the robot.}
\rtfield{Location}{National University of Singapore, a table nearby The Terrace}
\rtfield{Address}{The Terrace, 100 Edgedale Plains, Singapore 828690, Singapore}
\rtmeta{\rtinline{Environment}{Semi-open terrace}\rtsep\rtinline{Time}{4 PM}\rtsep\rtinline{Timeout}{25 s}}
\begin{rtprogress}
  \rtscore{0}{No meaningful progress.}
  \rtscore{1}{Bottle is successfully grasped.}
  \rtscore{2}{Bottle is lifted from its initial lying position.}
  \rtscore{3}{Bottle is rotated toward an upright orientation or placed on its base but is unstable/not fully released.}
  \rtscore{4}{Bottle is upright, stable on its base, and released by the robot.}
\end{rtprogress}
}

\realtaskcard
{FT-SO101-04}
{stack\_cubes}
{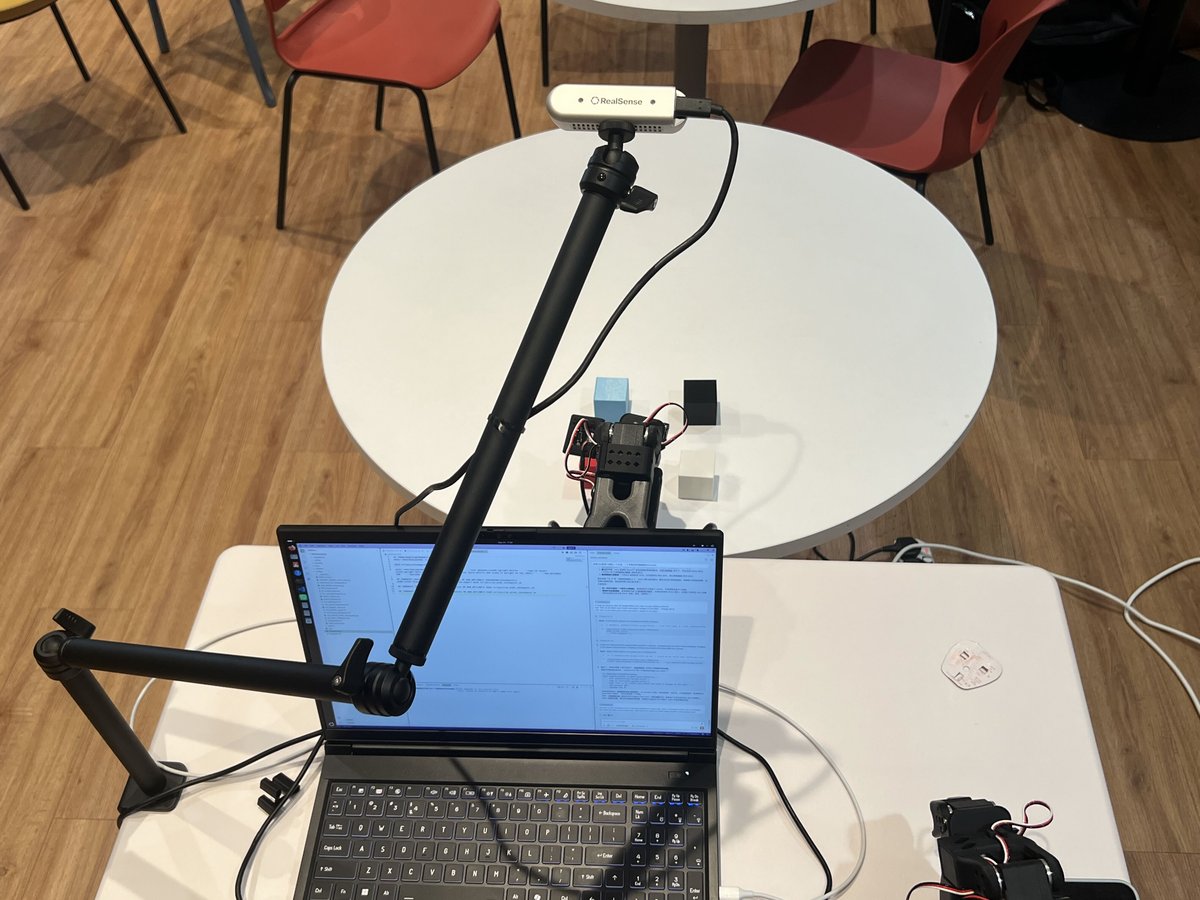}
{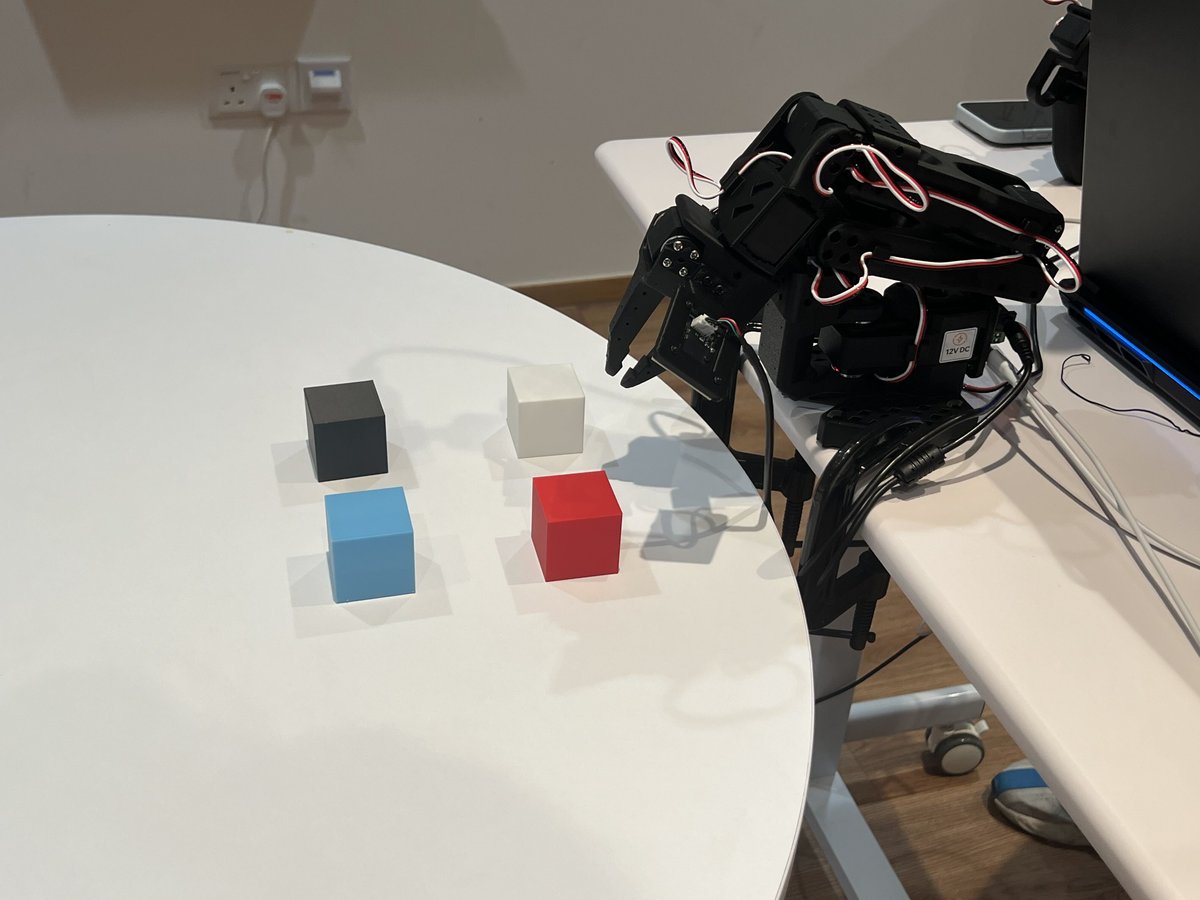}
{
\rtfield{Instruction}{Stack the blue cube on the red cube, then stack the white cube on the blue cube.}
\rtfield{Objects Used}{4 cubes}
\rtfield{Object Setup}{The red, blue, black, and white cubes are placed separately on the tabletop within the robot's reachable workspace.}
\rtfield{Success Criteria}{The cubes form a stable three-cube stack with the red cube at the bottom, the blue cube in the middle, and the white cube at the top.}
\rtfield{Location}{National University of Singapore, COM3-2, lounge}
\rtfield{Address}{National University of Singapore, 11 Research Link, COM3, \#B1-16, Singapore 119391}
\rtmeta{\rtinline{Environment}{Lounge}\rtsep\rtinline{Time}{5 PM}\rtsep\rtinline{Timeout}{30 s}}
\begin{rtprogress}
  \rtscore{0}{No meaningful progress.}
  \rtscore{1}{Blue cube is approached.}
  \rtscore{2}{Blue cube is successfully grasped and stacked on the red cube.}
  \rtscore{3}{White cube is moved onto or above the blue cube but the final three-cube stack is incomplete or unstable.}
  \rtscore{4}{Red, blue, and white cubes form a stable three-cube stack in the correct order: red bottom, blue middle, white top.}
\end{rtprogress}
}

\realtaskcard
{FT-SO101-05}
{cube\_drawer}
{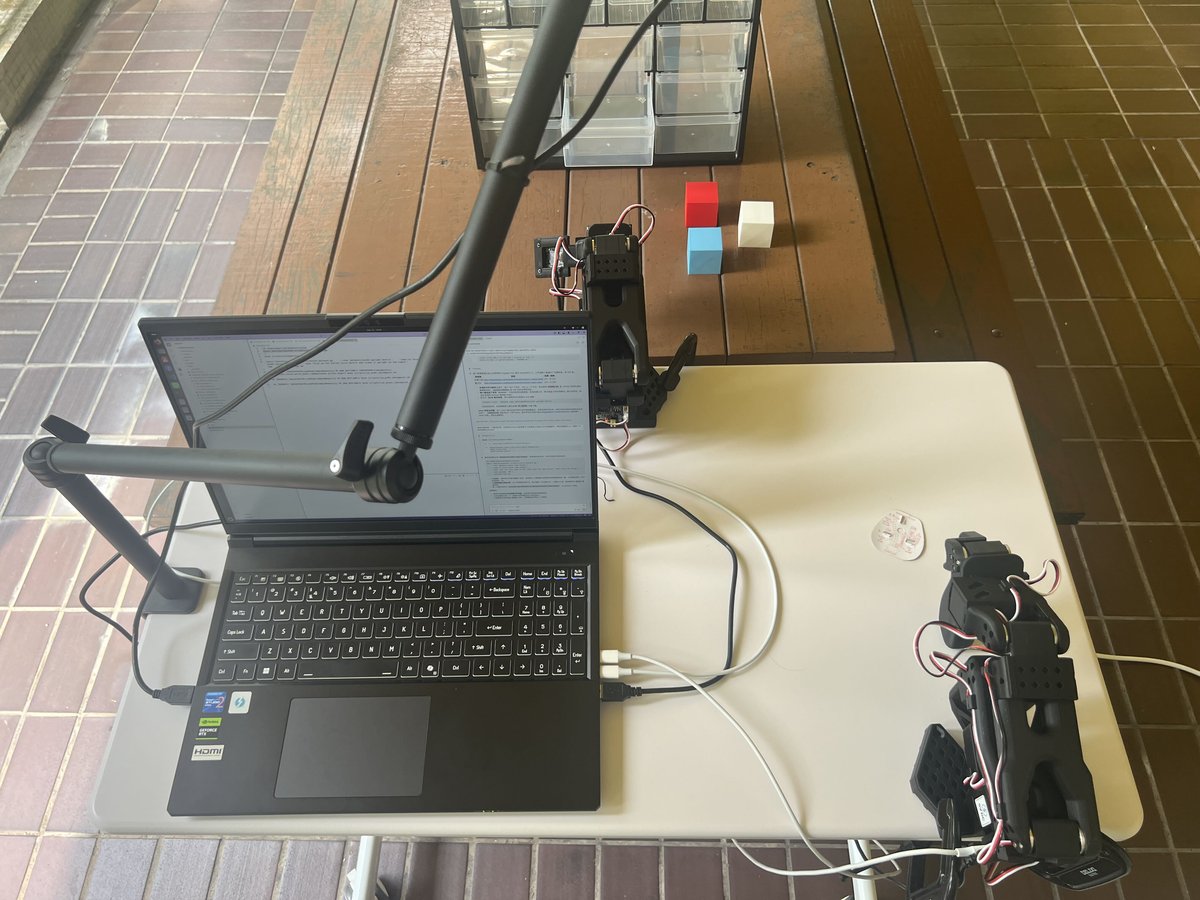}
{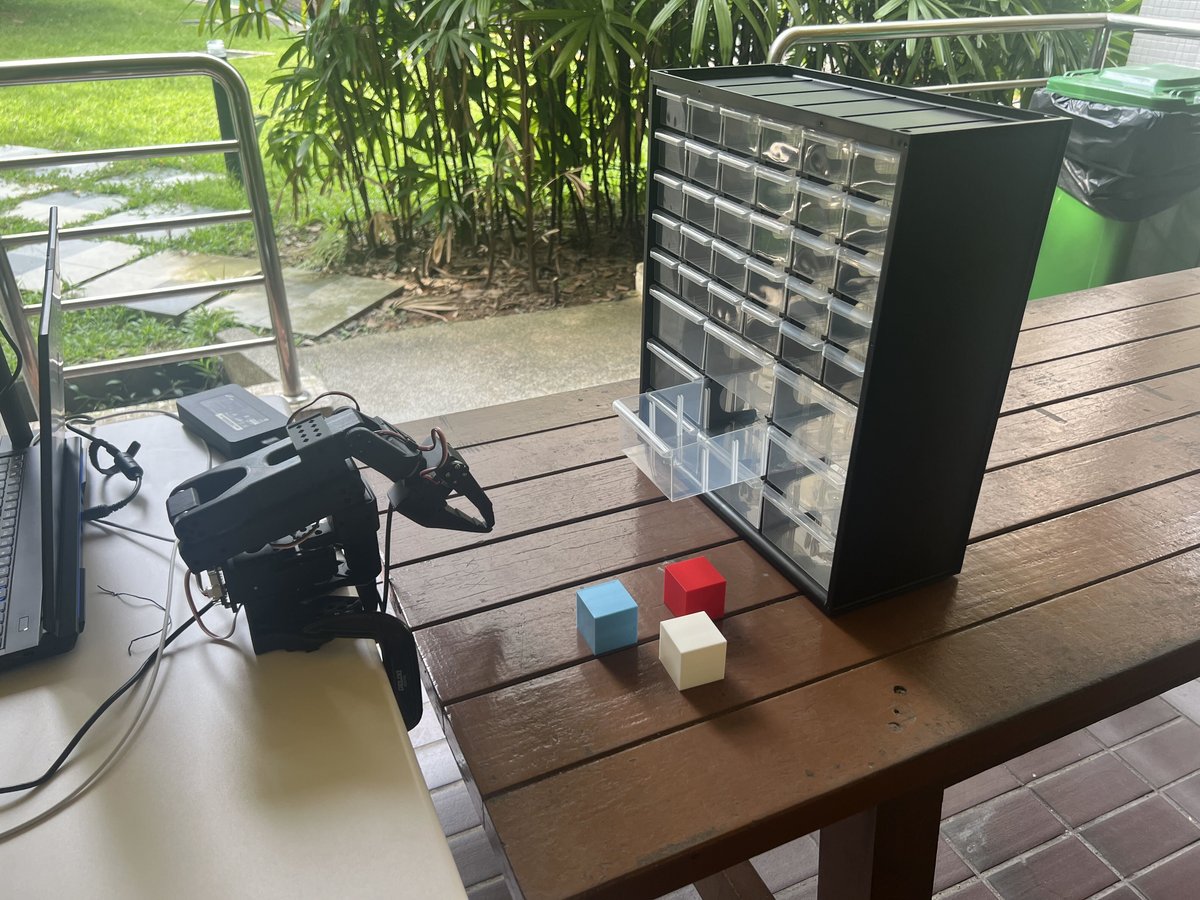}
{
\rtfield{Instruction}{Put the blue cube on the table into the open drawer, then close the drawer.}
\rtfield{Objects Used}{3 cubes; 1 drawer}
\rtfield{Object Setup}{The blue cube is placed on the tabletop, and the target drawer is initially open and accessible to the robot.}
\rtfield{Success Criteria}{The blue cube is fully contained inside the target drawer, and the drawer is fully closed.}
\rtfield{Location}{National University of Singapore, a table in front of LT19}
\rtfield{Address}{The Terrace, 100 Edgedale Plains, Singapore 828690, Singapore}
\rtmeta{\rtinline{Environment}{Semi-open terrace}\rtsep\rtinline{Time}{4 PM}\rtsep\rtinline{Timeout}{30 s}}
\begin{rtprogress}
  \rtscore{0}{No meaningful progress.}
  \rtscore{1}{Blue cube is successfully grasped and lifted.}
  \rtscore{2}{Blue cube is moved to or above the open drawer.}
  \rtscore{3}{Blue cube is fully placed inside the drawer, but the drawer is not fully closed.}
  \rtscore{4}{Blue cube is fully contained inside the target drawer and the drawer is fully closed.}
\end{rtprogress}
}

\subsection{DROID}
\label{app:finetuning_track_droid}

\realtaskcard
{FT-DROID-01}
{close\_laptop}
{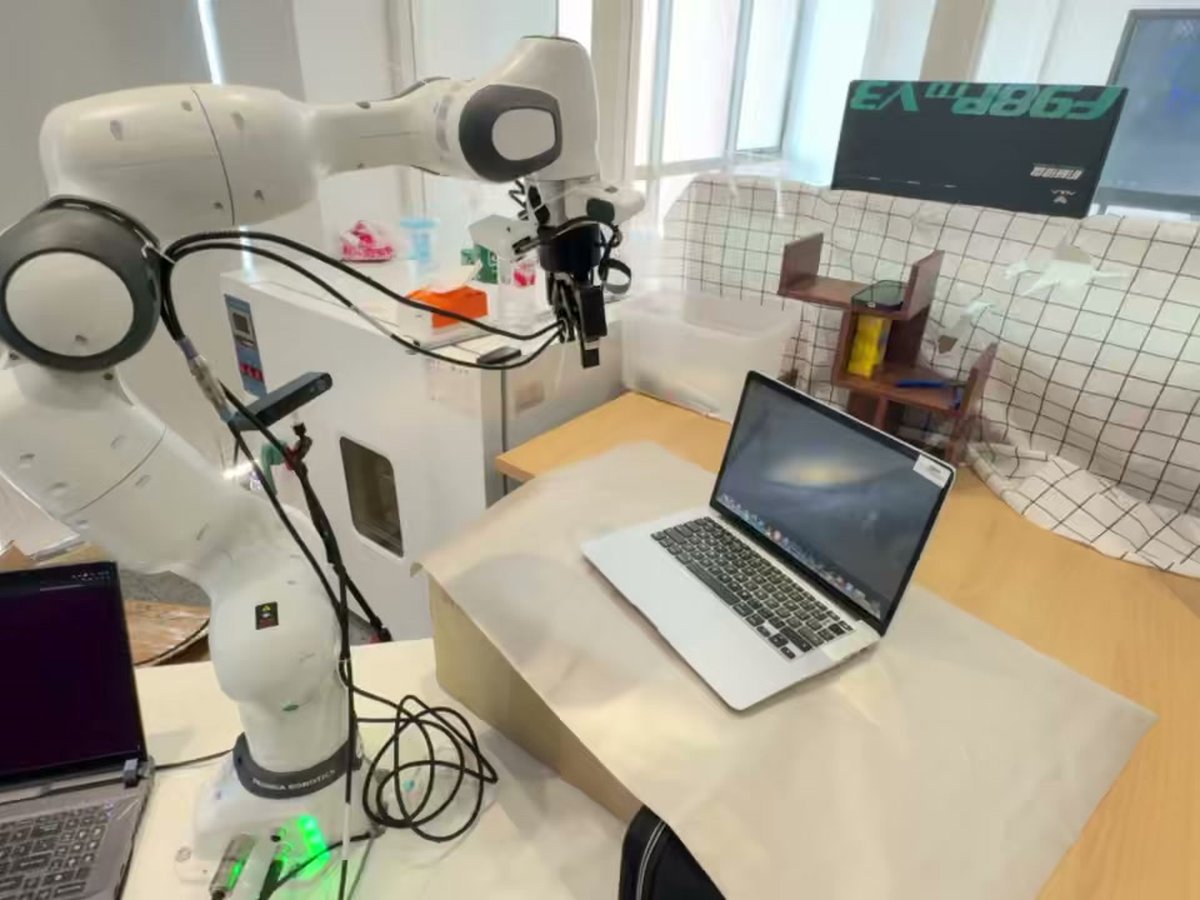}
{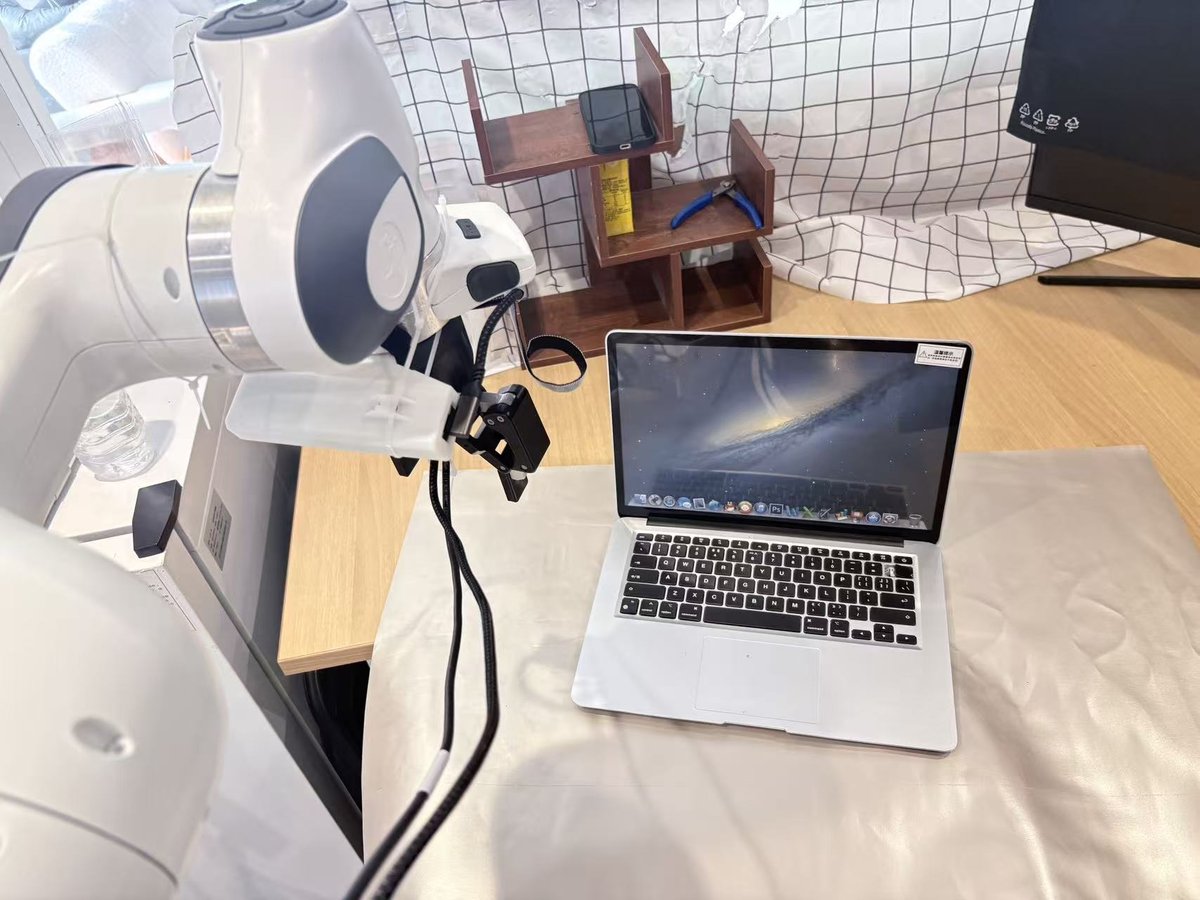}
{
\rtfield{Instruction}{Close laptop}
\rtfield{Objects Used}{1 Laptop}
\rtfield{Object Setup}{Laptop placed open on the table within the robot workspace, with the screen facing the robot and the laptop base resting flat on the table.}
\rtfield{Success Criteria}{Lid completely closed.}
\rtfield{Location}{Office 728, Haichuan Building}
\rtfield{Address}{Haichuan Building, Xixiang Subdistrict, Bao'an District, Shenzhen City, Guangdong Province (No.158 Wenchang Road), Postal Code: 518126}
\rtmeta{\rtinline{Environment}{Office}\rtsep\rtinline{Time}{7 PM}\rtsep\rtinline{Timeout}{20 s}}
\begin{rtprogress}
  \rtscore{0}{No meaningful progress.}
  \rtscore{1}{Robot makes controlled contact with the laptop lid and begins moving it downward.}
  \rtscore{2}{Lid is closed at least halfway from its initial open position.}
  \rtscore{3}{Lid is nearly closed but a visible gap remains between the lid and base.}
  \rtscore{4}{Lid is completely closed.}
\end{rtprogress}
}

\realtaskcard
{FT-DROID-02}
{mug\_on\_coaster}
{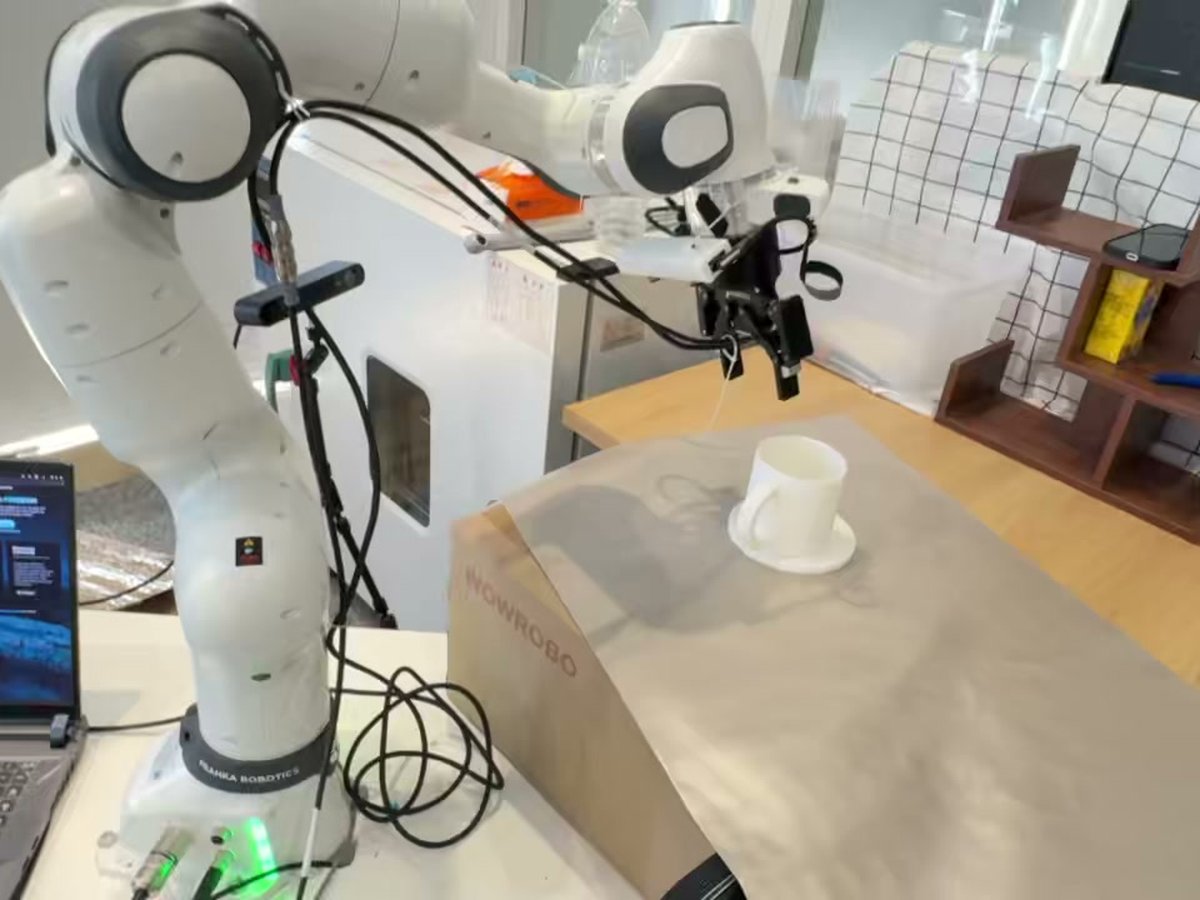}
{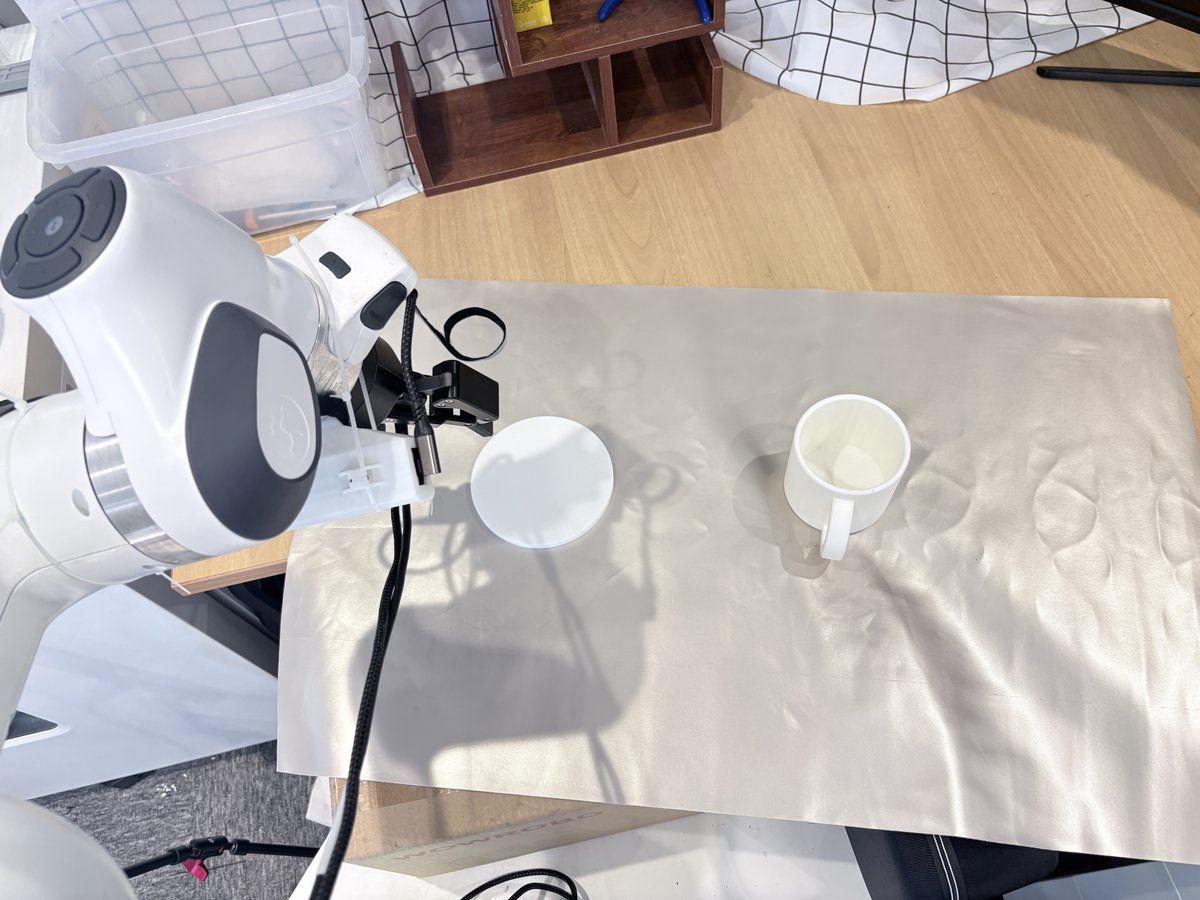}
{
\rtfield{Instruction}{Place the mug on the coaster}
\rtfield{Objects Used}{1 mug and 1 coaster}
\rtfield{Object Setup}{Mug and coaster placed separately on the table within the robot workspace. The mug starts upright and is not initially touching or overlapping the coaster.}
\rtfield{Success Criteria}{Mug fully supported by coaster.}
\rtfield{Location}{Office 728, Haichuan Building}
\rtfield{Address}{Haichuan Building, Xixiang Subdistrict, Bao'an District, Shenzhen City, Guangdong Province (No.158 Wenchang Road), Postal Code: 518126}
\rtmeta{\rtinline{Environment}{Office}\rtsep\rtinline{Time}{8 PM}\rtsep\rtinline{Timeout}{20 s}}
\begin{rtprogress}
  \rtscore{0}{No meaningful progress.}
  \rtscore{1}{Mug is successfully grasped and lifted.}
  \rtscore{2}{Mug is moved to the coaster area or positioned above the coaster.}
  \rtscore{3}{Mug is placed in contact with the coaster but is not fully supported by it.}
  \rtscore{4}{Mug is fully supported by the coaster and released by the robot.}
\end{rtprogress}
}

\realtaskcard
{FT-DROID-03}
{open\_drawer}
{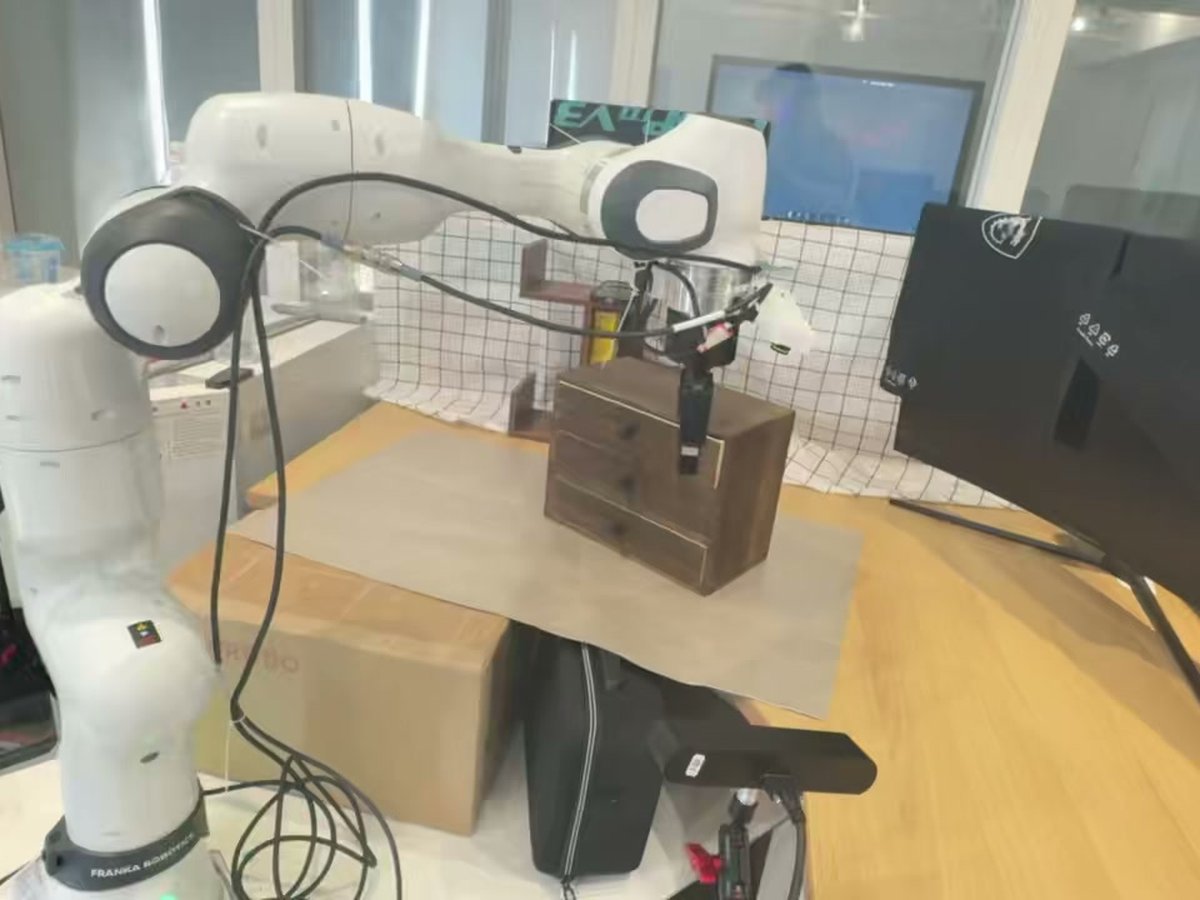}
{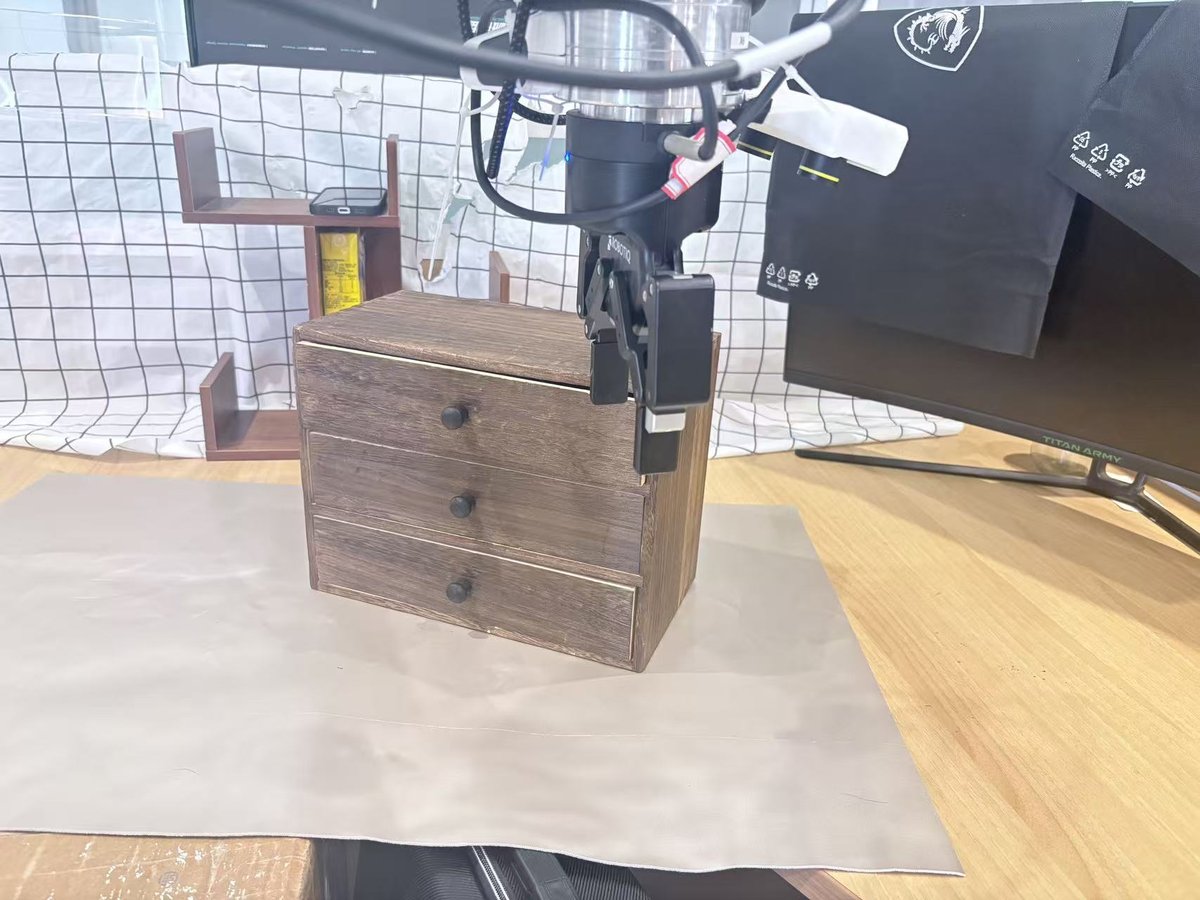}
{
\rtfield{Instruction}{Open the drawer}
\rtfield{Objects Used}{1 drawer}
\rtfield{Object Setup}{Drawer starts fully closed and within reach of the robot. The drawer handle is unobstructed and accessible to the robot gripper.}
\rtfield{Success Criteria}{Drawer displacement from the fully closed position is $\geq$70\% of its calibrated maximum travel and remains above this threshold for $\geq$1 s.}
\rtfield{Location}{Office 728, Haichuan Building}
\rtfield{Address}{Haichuan Building, Xixiang Subdistrict, Bao'an District, Shenzhen City, Guangdong Province (No.158 Wenchang Road), Postal Code: 518126}
\rtmeta{\rtinline{Environment}{Office}\rtsep\rtinline{Time}{6 PM}\rtsep\rtinline{Timeout}{25 s}}
\begin{rtprogress}
  \rtscore{0}{No meaningful progress.}
  \rtscore{1}{Robot successfully engages the drawer handle.}
  \rtscore{2}{Drawer is opened to at least 25\% but less than 50\% of its calibrated maximum travel.}
  \rtscore{3}{Drawer is opened to at least 50\% but less than 70\% of its calibrated maximum travel.}
  \rtscore{4}{Drawer displacement is $\geq$70\% of its calibrated maximum travel and remains above this threshold for $\geq$1 s.}
\end{rtprogress}
}

\realtaskcard
{FT-DROID-04}
{marker\_in\_holder}
{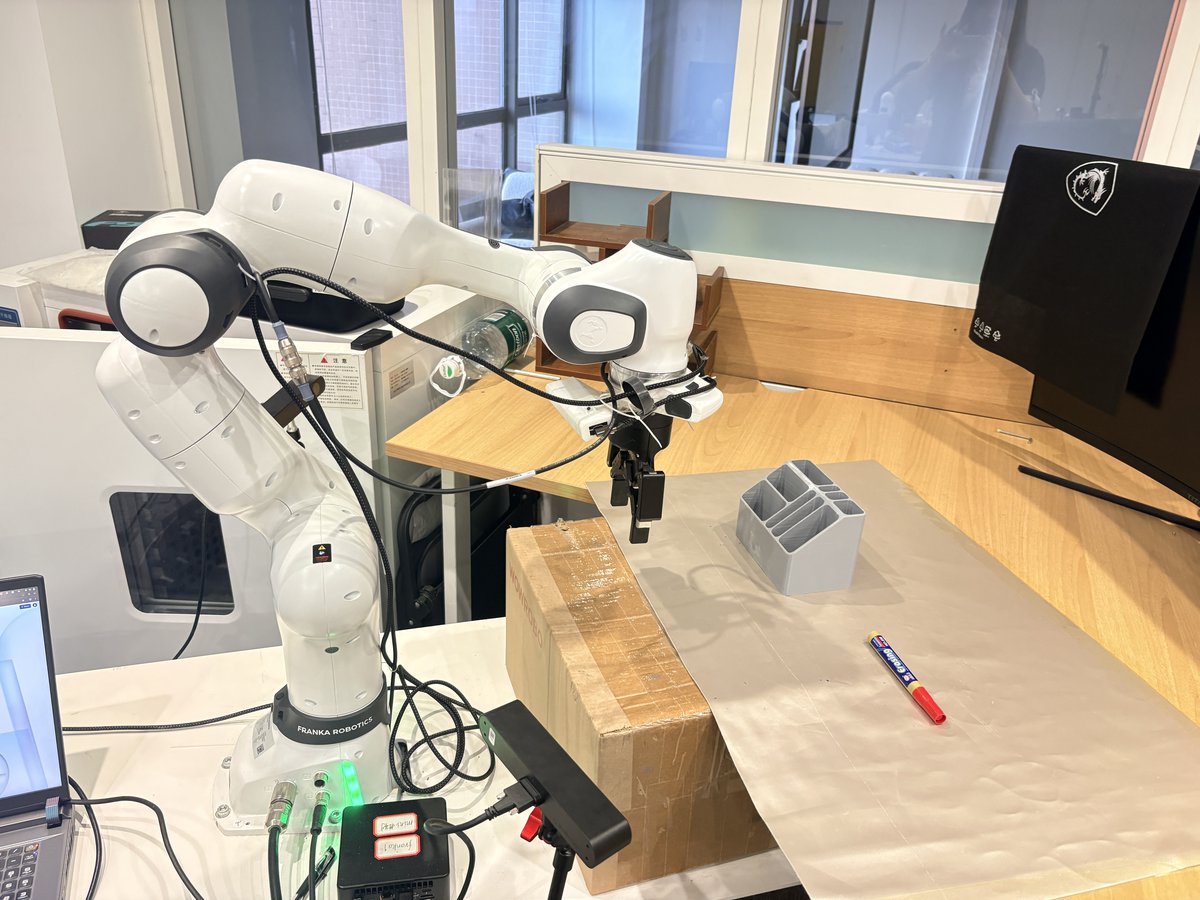}
{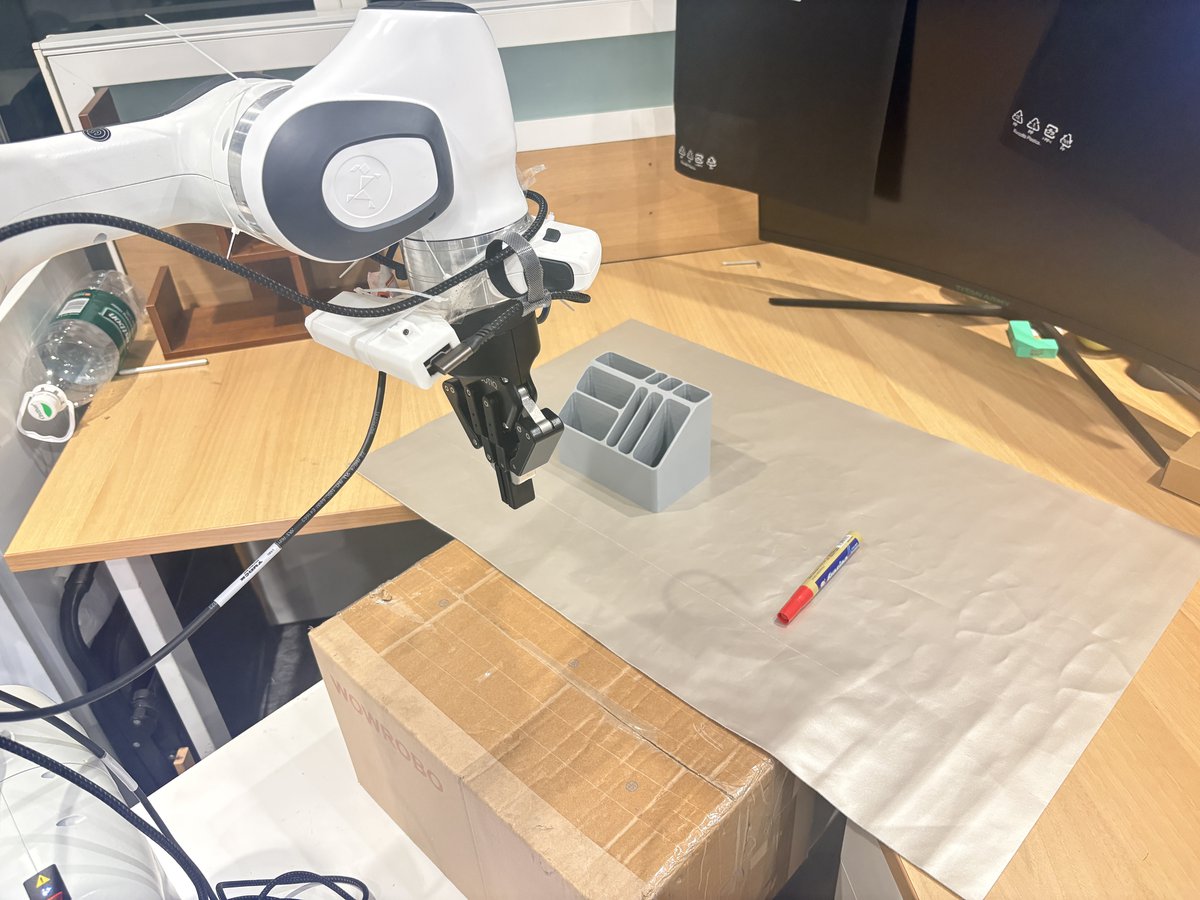}
{
\rtfield{Instruction}{Pick up the marker and place it into the pen holder.}
\rtfield{Objects Used}{1 marker, 1 pen holder / cup}
\rtfield{Object Setup}{A marker lies horizontally on the table within the robot workspace. An empty pen holder (or opaque cup) is placed upright 20 cm nearby and does not initially touch the marker.}
\rtfield{Success Criteria}{Marker is placed inside the pen holder and remains there after release for at least 1 second.}
\rtfield{Location}{Office 728, Haichuan Building}
\rtfield{Address}{Haichuan Building, Xixiang Subdistrict, Bao'an District, Shenzhen City, Guangdong Province (No.158 Wenchang Road), Postal Code: 518126}
\rtmeta{\rtinline{Environment}{Office}\rtsep\rtinline{Time}{9 PM}\rtsep\rtinline{Timeout}{30 s}}
\begin{rtprogress}
  \rtscore{0}{No meaningful progress.}
  \rtscore{1}{Robot successfully grasps and lifts the marker.}
  \rtscore{2}{Marker is moved above or next to the pen holder opening.}
  \rtscore{3}{Marker is partially inserted into the pen holder but not yet stably released.}
  \rtscore{4}{Marker is successfully placed inside the pen holder and remains there after release for at least 1 second.}
\end{rtprogress}
}

\realtaskcard
{FT-DROID-05}
{towel\_in\_basket}
{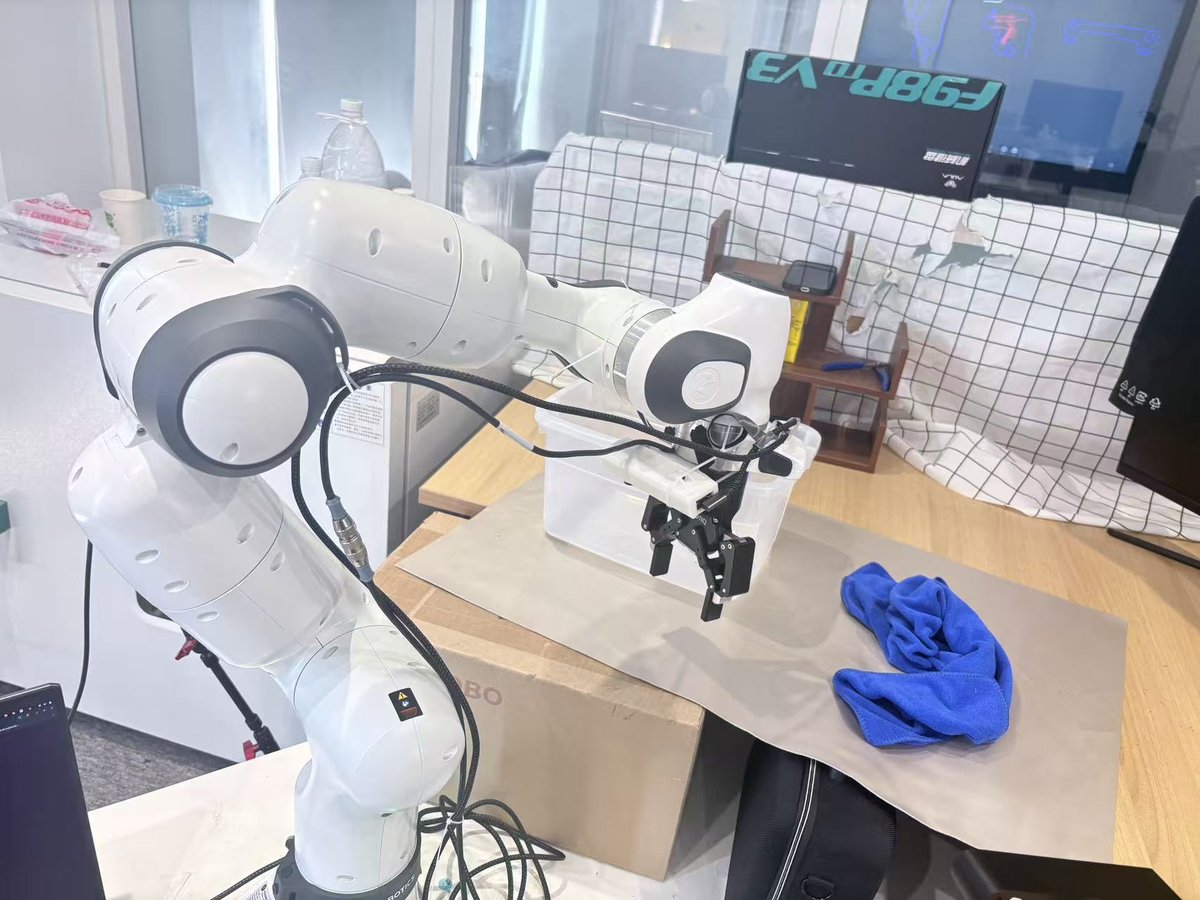}
{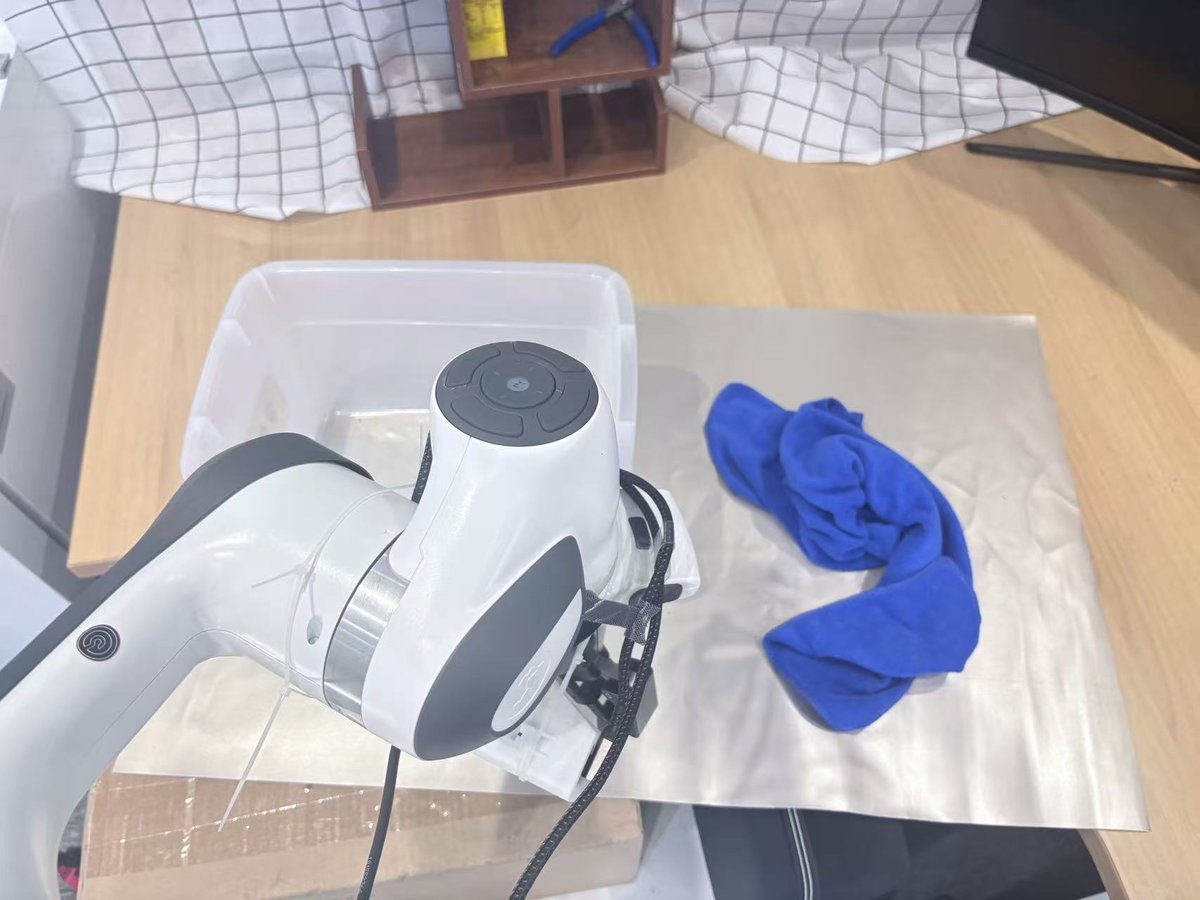}
{
\rtfield{Instruction}{Put the towel into the basket}
\rtfield{Objects Used}{1 towel and 1 basket}
\rtfield{Object Setup}{Towel is placed outside the basket on the table within the robot workspace. The basket is upright with its opening unobstructed and the towel is fully outside the basket at the start.}
\rtfield{Success Criteria}{Towel completely inside basket.}
\rtfield{Location}{Office 728, Haichuan Building}
\rtfield{Address}{Haichuan Building, Xixiang Subdistrict, Bao'an District, Shenzhen City, Guangdong Province (No.158 Wenchang Road)}
\rtmeta{\rtinline{Environment}{Office}\rtsep\rtinline{Time}{4 PM}\rtsep\rtinline{Timeout}{20 s}}
\begin{rtprogress}
  \rtscore{0}{No meaningful progress.}
  \rtscore{1}{Towel is successfully grasped and lifted.}
  \rtscore{2}{Towel is moved to or above the basket opening.}
  \rtscore{3}{Towel is partially inside the basket but some portion remains outside.}
  \rtscore{4}{Towel is completely inside the basket and released by the robot.}
\end{rtprogress}
}

\subsection{YAM}
\label{app:finetuning_track_yam}

\realtaskcard
{FT-YAM-01}
{dustpan}
{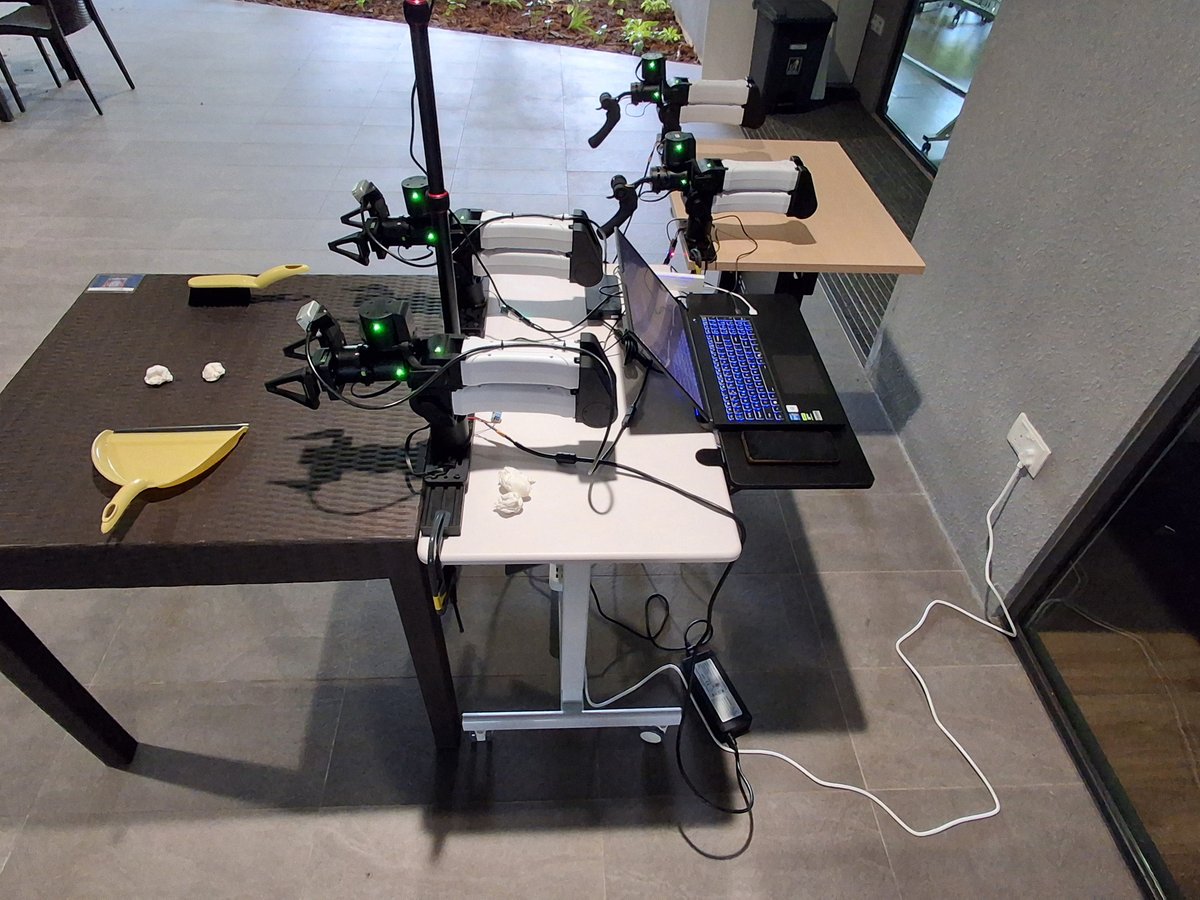}
{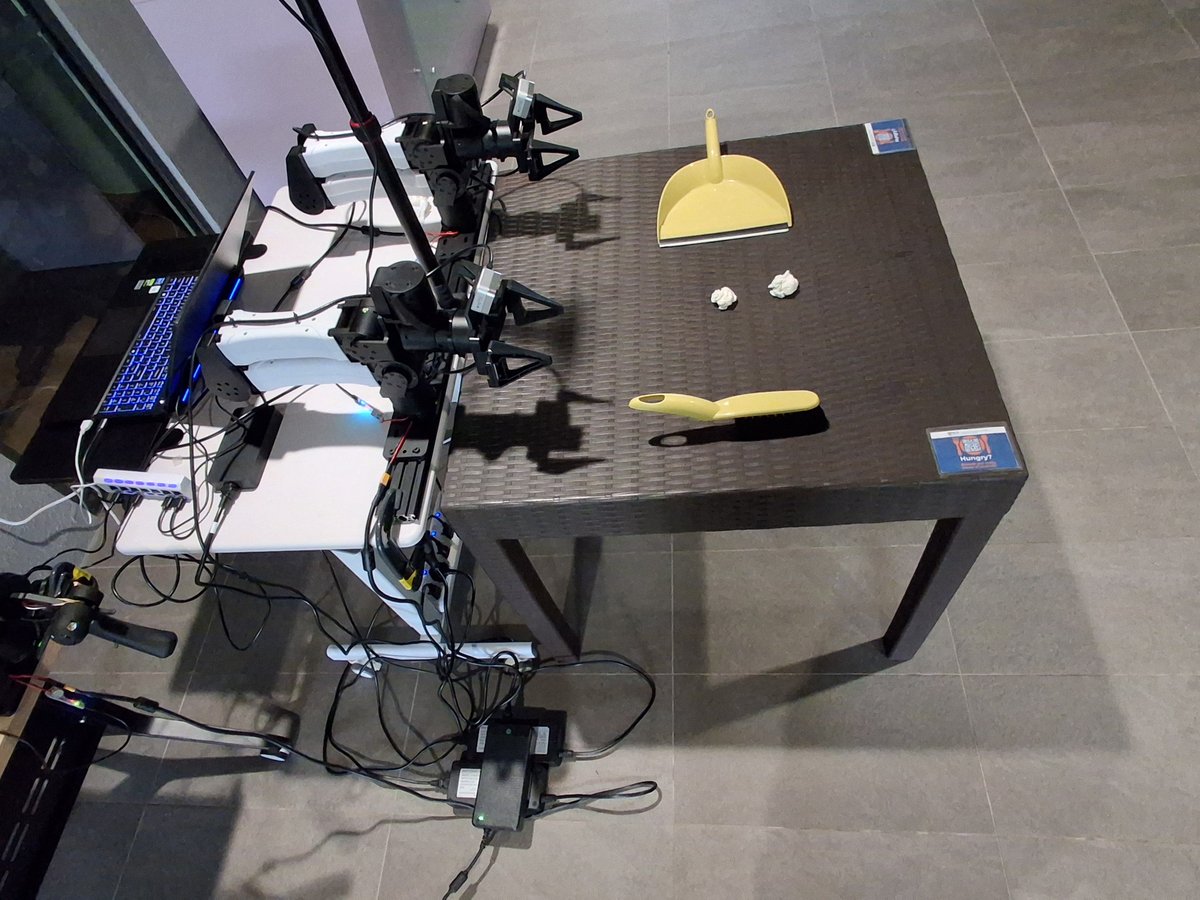}
{
\rtfield{Instruction}{Clean the table.}
\rtfield{Objects Used}{1 dustpan, 1 brush, and 2 tissues}
\rtfield{Object Setup}{1 dustpan left side, 1 brush right side and 2 tissues, in the middle.}
\rtfield{Success Criteria}{Tissues on the dustpan at the end.}
\rtfield{Location}{National University of Singapore, place nearby COM3-B1-16}
\rtfield{Address}{National University of Singapore, 11 Research Link, COM3, \#B1-16, Singapore 119391}
\rtmeta{\rtinline{Environment}{Semi-open terrace}\rtsep\rtinline{Time}{4 PM}\rtsep\rtinline{Timeout}{30 s}}
\begin{rtprogress}
  \rtscore{0}{No meaningful progress.}
  \rtscore{1}{The pan is picked up by the left arm.}
  \rtscore{2}{The sweep is picked up by the right arm.}
  \rtscore{3}{The tissue balls are swept into the pan.}
\end{rtprogress}
}

\realtaskcard
{FT-YAM-02}
{drawer}
{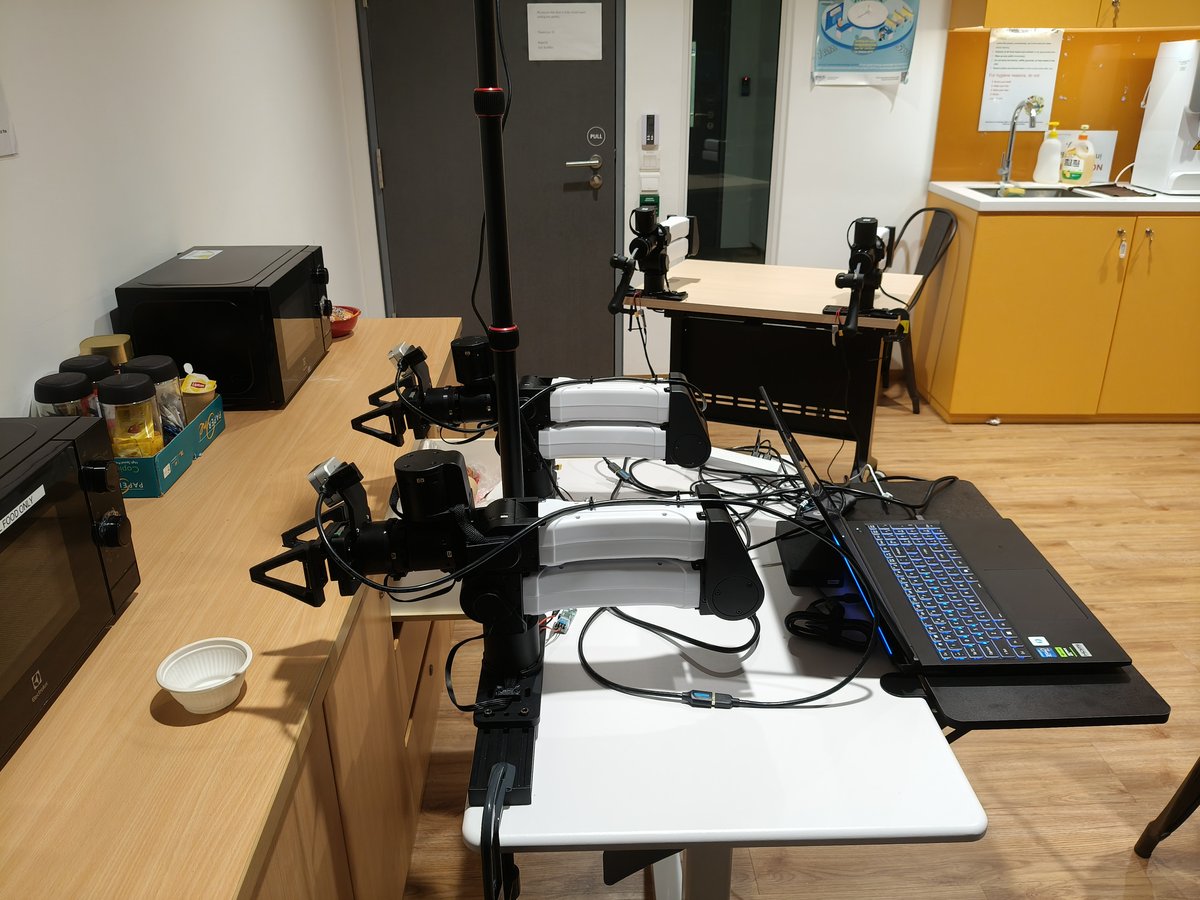}
{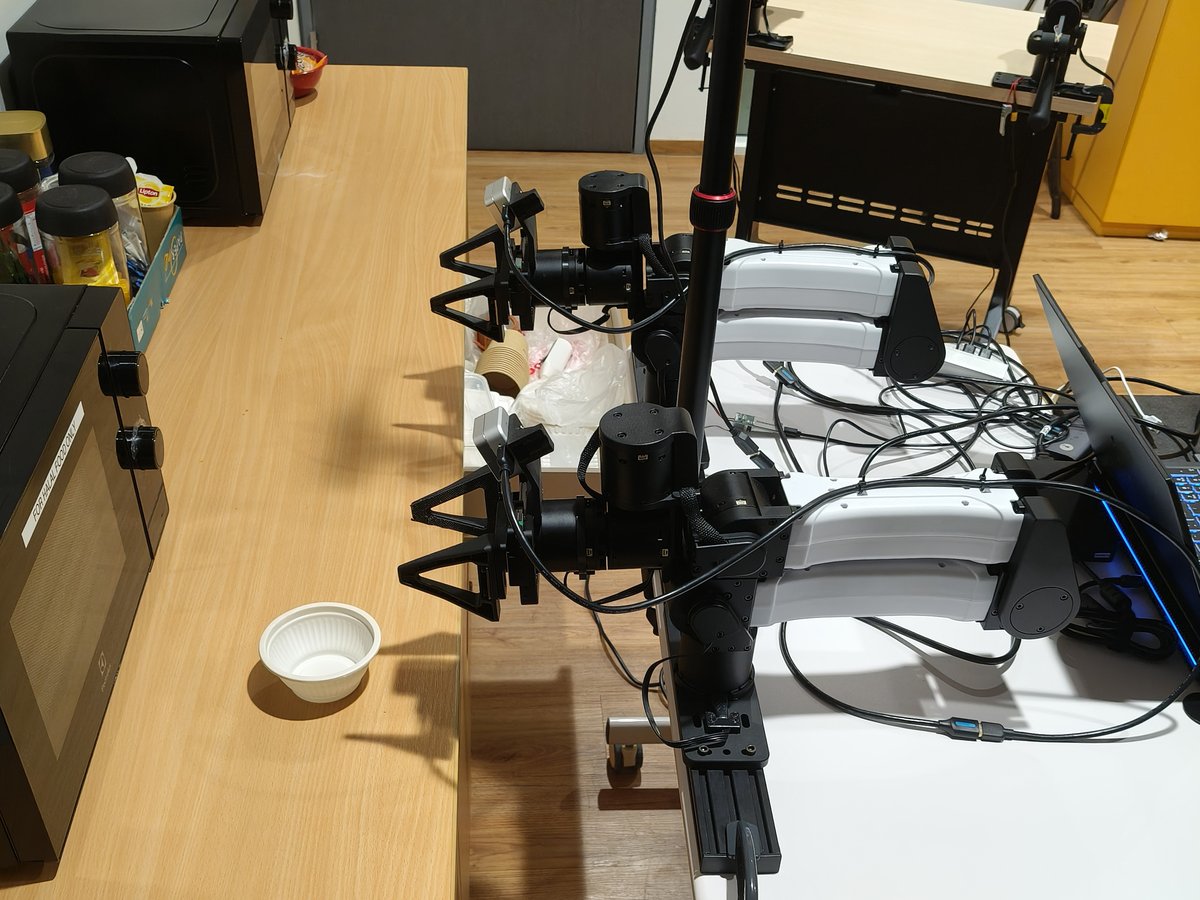}
{
\rtfield{Instruction}{Put the cup into the drawer.}
\rtfield{Objects Used}{Plastic cup, drawer}
\rtfield{Object Setup}{Plastic cup kept in front of the left arm, right side drawer open.}
\rtfield{Success Criteria}{Cup inside the drawer and drawer closed.}
\rtfield{Location}{NUS, COM3-B1- Pantry}
\rtfield{Address}{National University of Singapore, 11 Research Link, COM3, \#B1-16, Singapore 119391}
\rtmeta{\rtinline{Environment}{Pantry}\rtsep\rtinline{Time}{4 PM}\rtsep\rtinline{Timeout}{90 s}}
\begin{rtprogress}
  \rtscore{0}{No meaningful progress.}
  \rtscore{1}{The cup is picked up and placed to the right.}
  \rtscore{2}{The right arm picks up the cup.}
  \rtscore{3}{The right arm drops the cup into the drawer.}
  \rtscore{4}{The drawer is fully closed.}
\end{rtprogress}
}

\realtaskcard
{FT-YAM-03}
{microwave}
{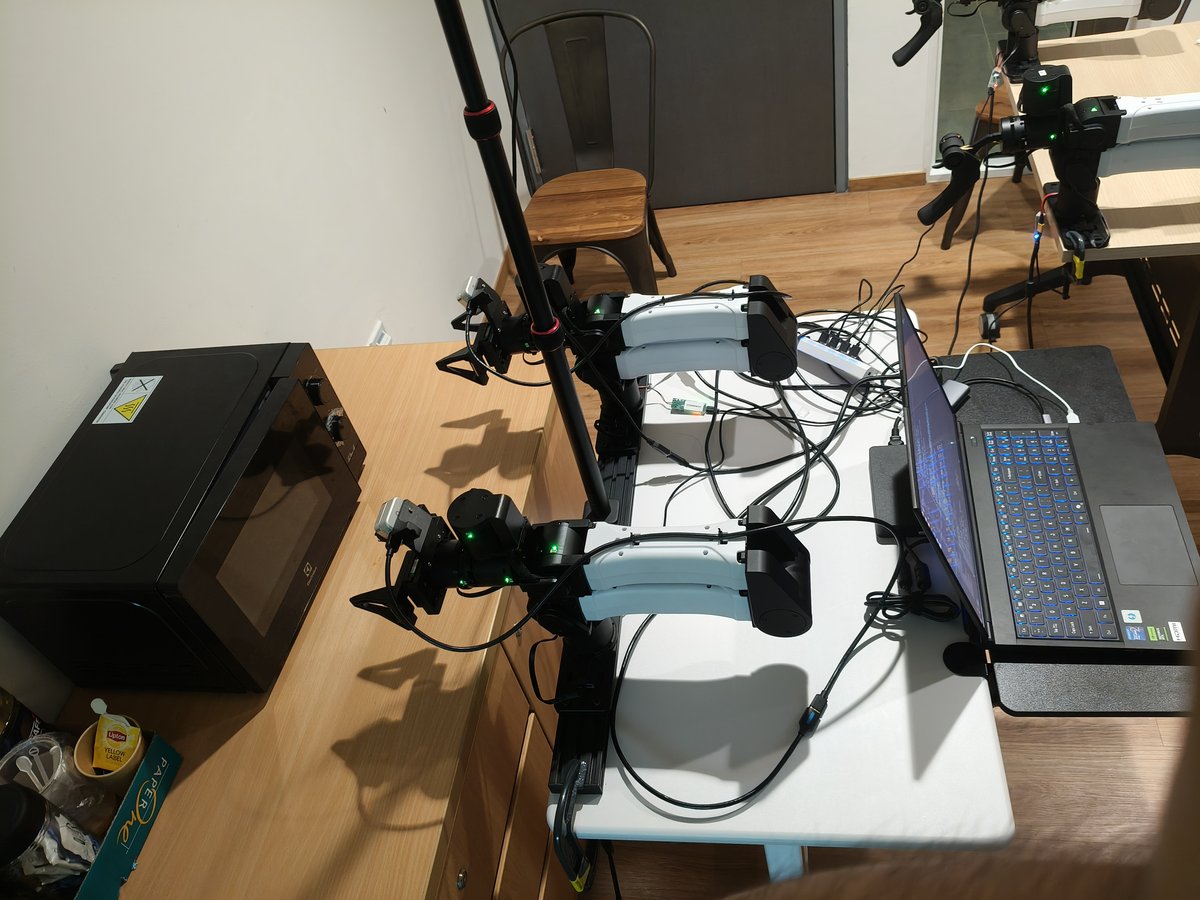}
{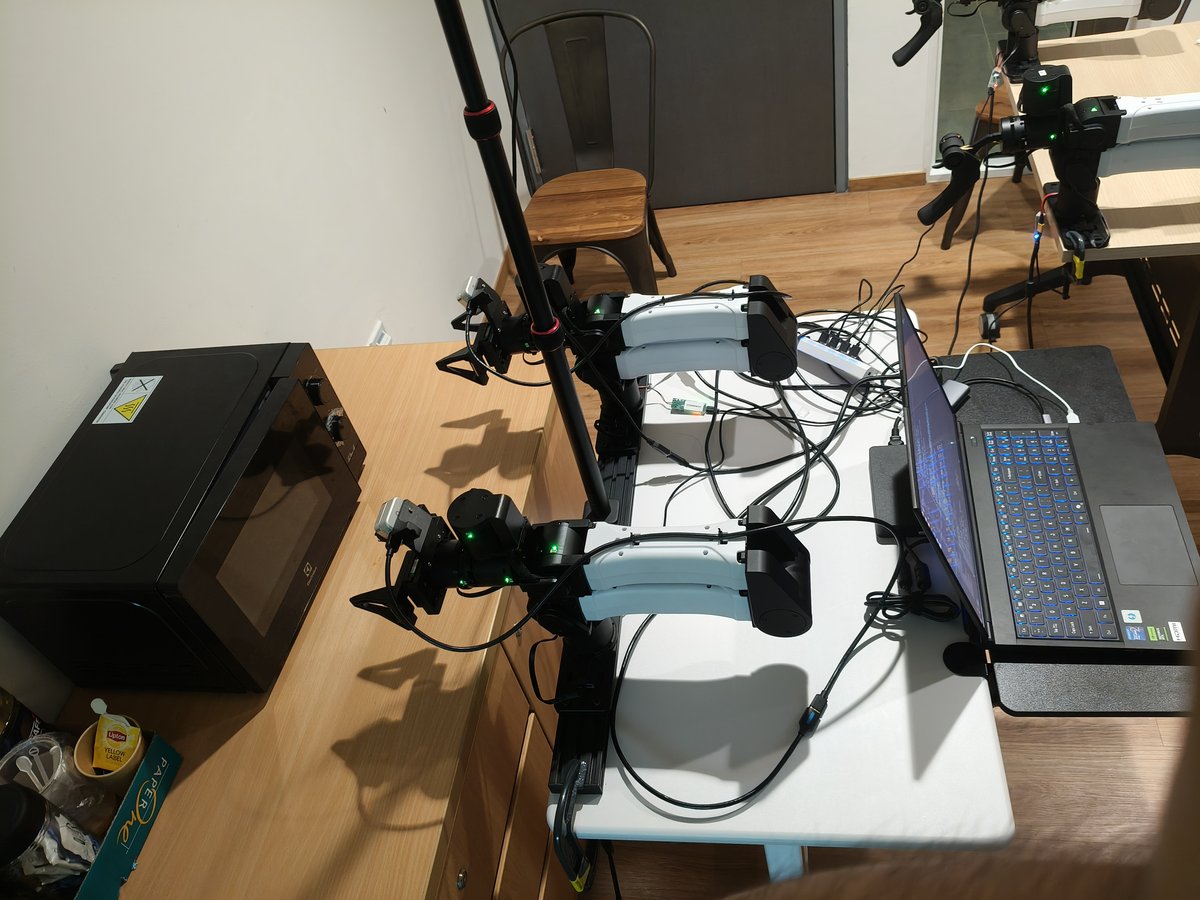}
{
\rtfield{Instruction}{Open the microwave and take out the bowl.}
\rtfield{Objects Used}{Microwave, red metal bowl, bread}
\rtfield{Object Setup}{Bread inside the bowl, bowl inside the microwave. Microwave half closed.}
\rtfield{Success Criteria}{Red bowl on the table outside of the microwave and microwave door half closed.}
\rtfield{Location}{NUS, COM3-B1- Pantry}
\rtfield{Address}{National University of Singapore, 11 Research Link, COM3, \#B1-16, Singapore 119391}
\rtmeta{\rtinline{Environment}{Pantry}\rtsep\rtinline{Time}{6 PM}\rtsep\rtinline{Timeout}{90 s}}
\begin{rtprogress}
  \rtscore{0}{No meaningful progress.}
  \rtscore{1}{The microwave door is opened by the left arm.}
  \rtscore{2}{The bowl is picked up and placed outside of the microwave by the right arm.}
  \rtscore{3}{The microwave door is pushed closed by the left arm.}
\end{rtprogress}
}

\realtaskcard
{FT-YAM-04}
{stack\_cups}
{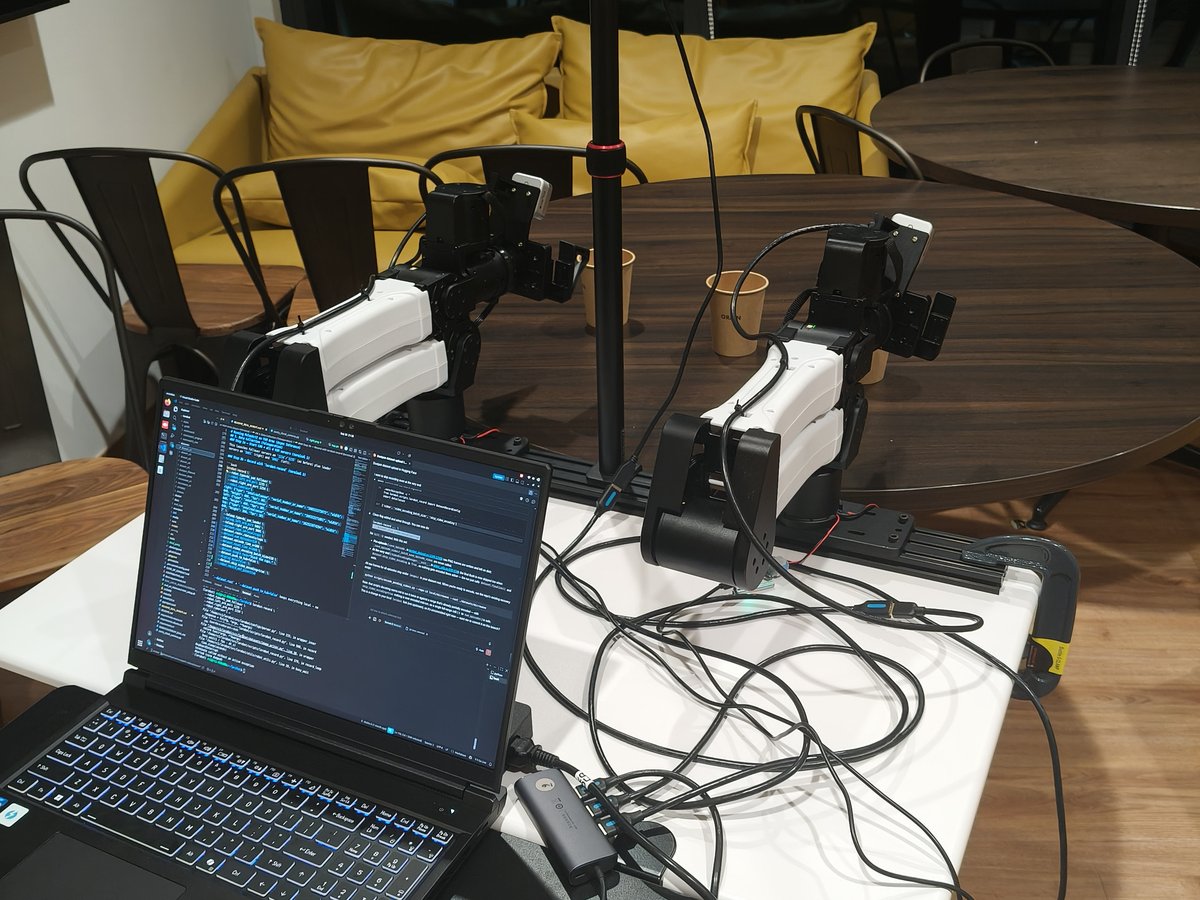}
{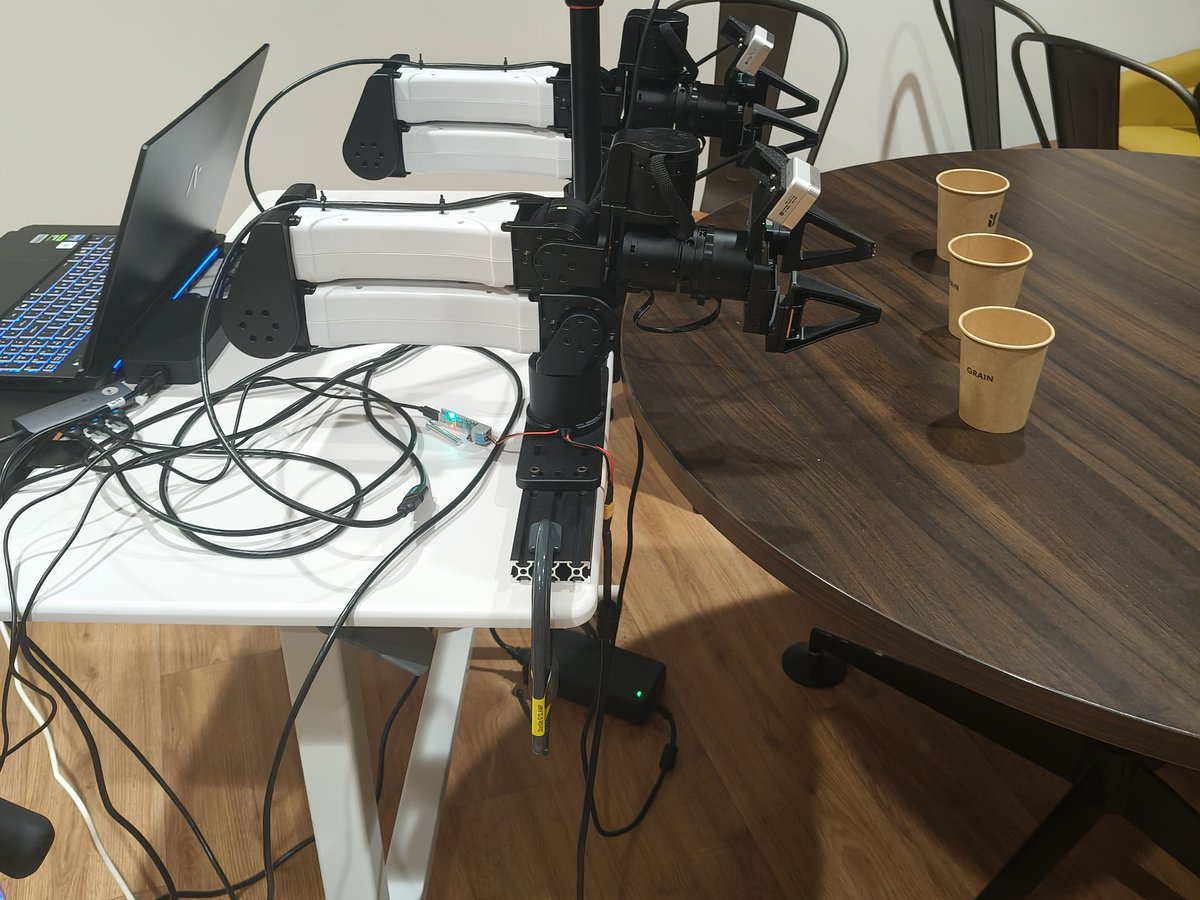}
{
\rtfield{Instruction}{Stack the cups.}
\rtfield{Objects Used}{3 paper cups}
\rtfield{Object Setup}{3 paper cups in a line on the table.}
\rtfield{Success Criteria}{The left and right paper cups stacked on top of the middle cup.}
\rtfield{Location}{NUS, COM3-B1- Pantry}
\rtfield{Address}{National University of Singapore, 11 Research Link, COM3, \#B1-16, Singapore 119391}
\rtmeta{\rtinline{Environment}{Pantry}\rtsep\rtinline{Time}{9:30 PM}\rtsep\rtinline{Timeout}{50 s}}
\begin{rtprogress}
  \rtscore{0}{No meaningful progress.}
  \rtscore{1}{The left cup is being picked up.}
  \rtscore{2}{The right cup is being picked up.}
  \rtscore{3}{The left cup is stacked on top of the central cup.}
  \rtscore{4}{The right cup is stacked on top of the central cup.}
\end{rtprogress}
}

\realtaskcard
{FT-YAM-05}
{clean\_table}
{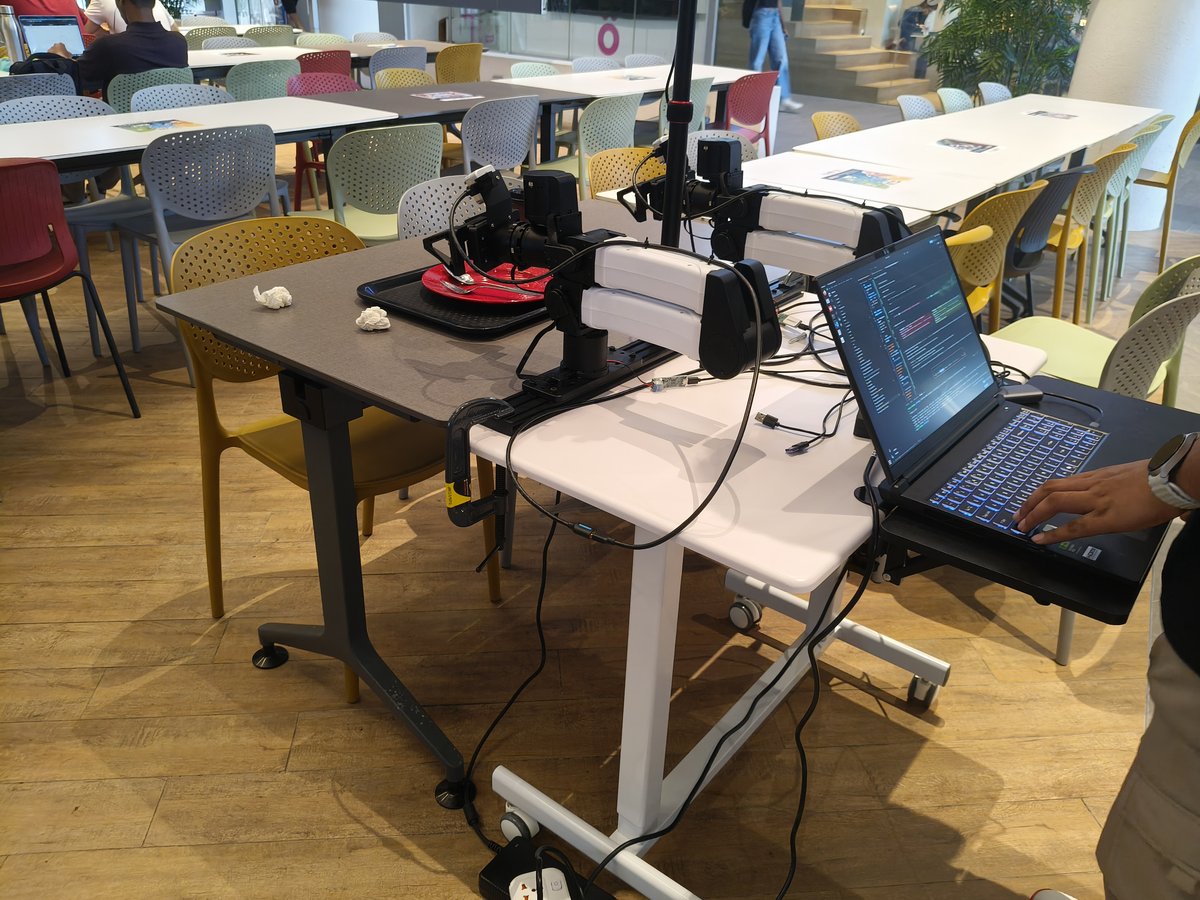}
{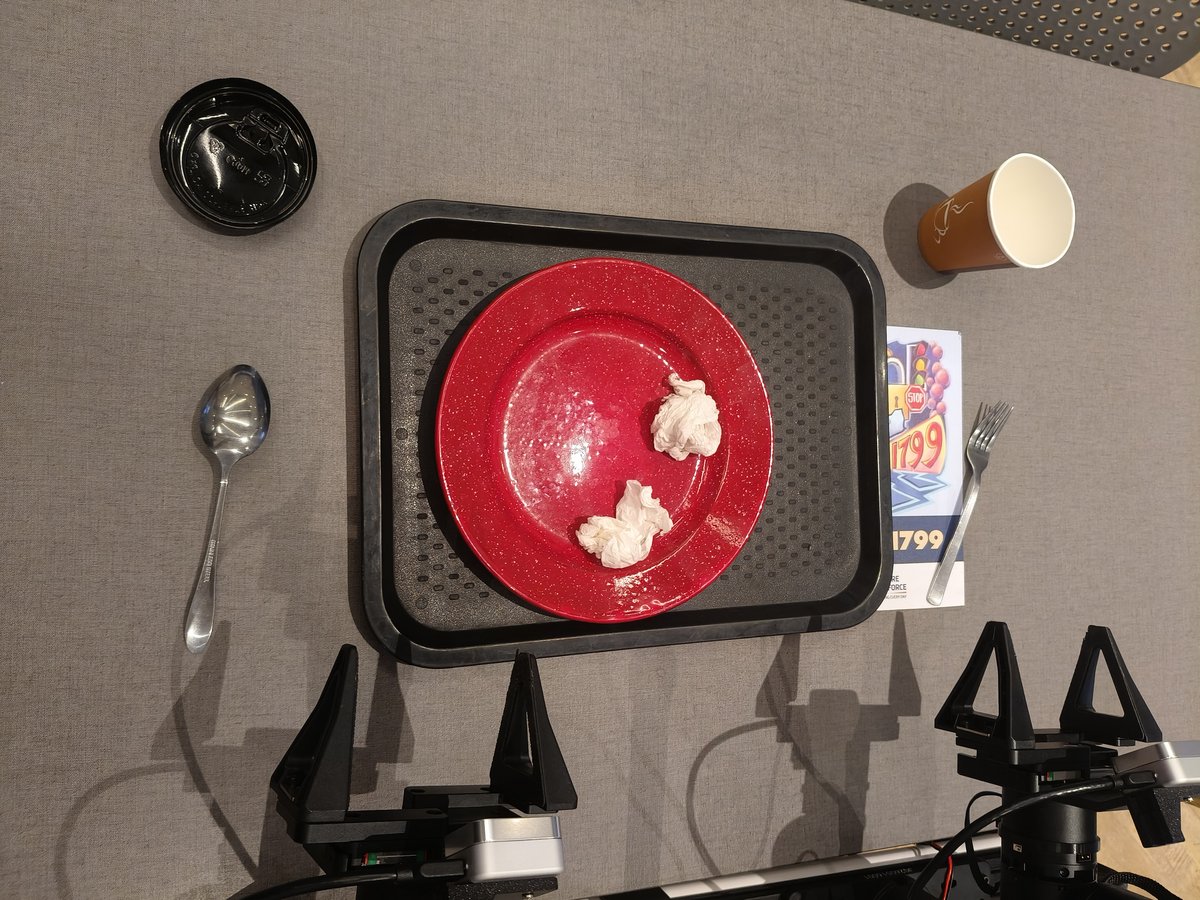}
{
\rtfield{Instruction}{Clean the table.}
\rtfield{Objects Used}{1 spoon, 1 fork and 1 black tray from terrace, 1 red plate, 2 tissue paper balls, 1 cool spot cup with cap}
\rtfield{Object Setup}{2 tissue balls on the red plate inside the tray. Fork and cup on the right side of the tray with cup on top, and spoon and cup cap on the left side of the tray with cap on top.}
\rtfield{Success Criteria}{Both utensils, the cup and the cap inside the tray.}
\rtfield{Location}{NUS, COM3-L1-Terrace}
\rtfield{Address}{The Terrace, 100 Edgedale Plains, Singapore 828690, Singapore}
\rtmeta{\rtinline{Environment}{Canteen}\rtsep\rtinline{Time}{5 PM}\rtsep\rtinline{Timeout}{40 s}}
\begin{rtprogress}
  \rtscore{0}{No meaningful progress.}
  \rtscore{1}{The spoon is being picked up and placed in the tray by the left arm.}
  \rtscore{2}{The fork is being picked up and placed in the tray by the right arm.}
  \rtscore{3}{The cap is being picked up and placed in the tray by the left arm.}
  \rtscore{4}{The cup is being picked up and placed in the tray by the right arm.}
\end{rtprogress}
}

\subsection{G1}
\label{app:finetuning_track_g1}

\realtaskcard
{FT-G1-01}
{plate\_into\_rack}
{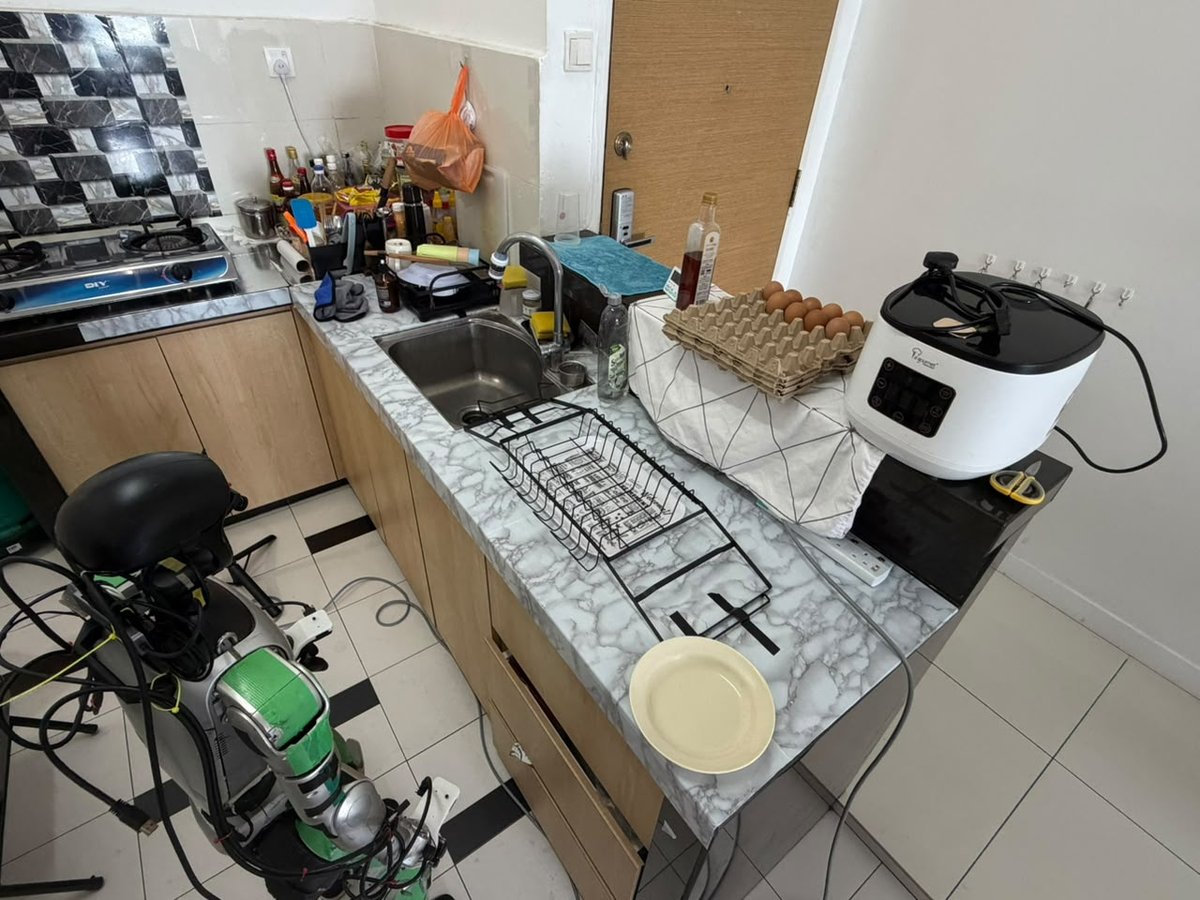}
{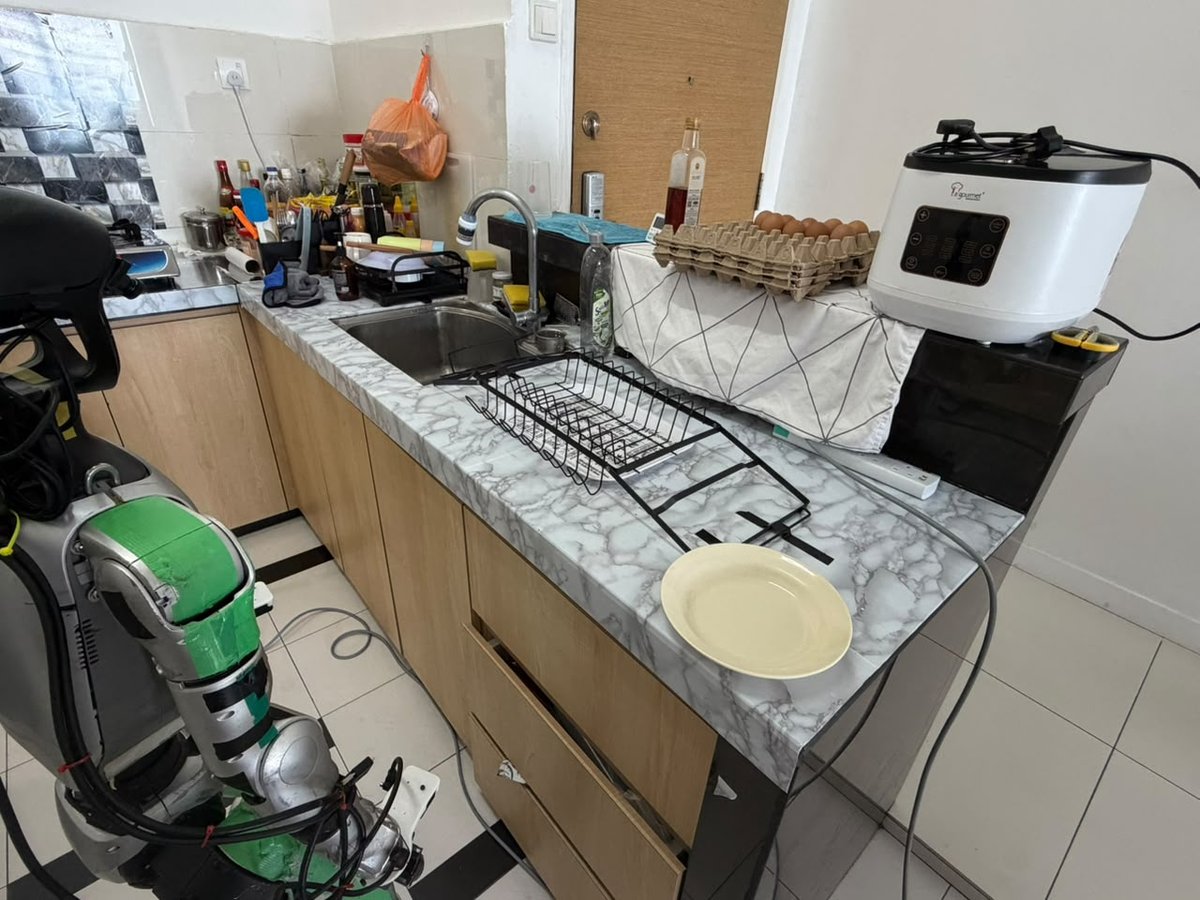}
{
\rtfield{Instruction}{Pick up the plate and place it upright into the dish rack}
\rtfield{Objects Used}{1 plate; 1 dish rack}
\rtfield{Object Setup}{The plate is placed flat on the countertop within the robot's reachable workspace. The empty dish rack is placed beside the plate with the slots facing the robot.}
\rtfield{Success Criteria}{The plate is placed upright in a slot of the dish rack, fully supported by the rack, stable, and released by the robot.}
\rtfield{Location}{Real Homes in Malaysia}
\rtmeta{\rtinline{Environment}{Kitchen}\rtsep\rtinline{Timeout}{150 s}}
\begin{rtprogress}
  \rtscore{0}{No meaningful progress.}
  \rtscore{1}{Walk forward.}
  \rtscore{2}{Pick up the plate.}
  \rtscore{3}{Switch hands.}
  \rtscore{4}{Turn around.}
  \rtscore{5}{Put on the dish rack.}
\end{rtprogress}
}

\realtaskcard
{FT-G1-02}
{carry\_box}
{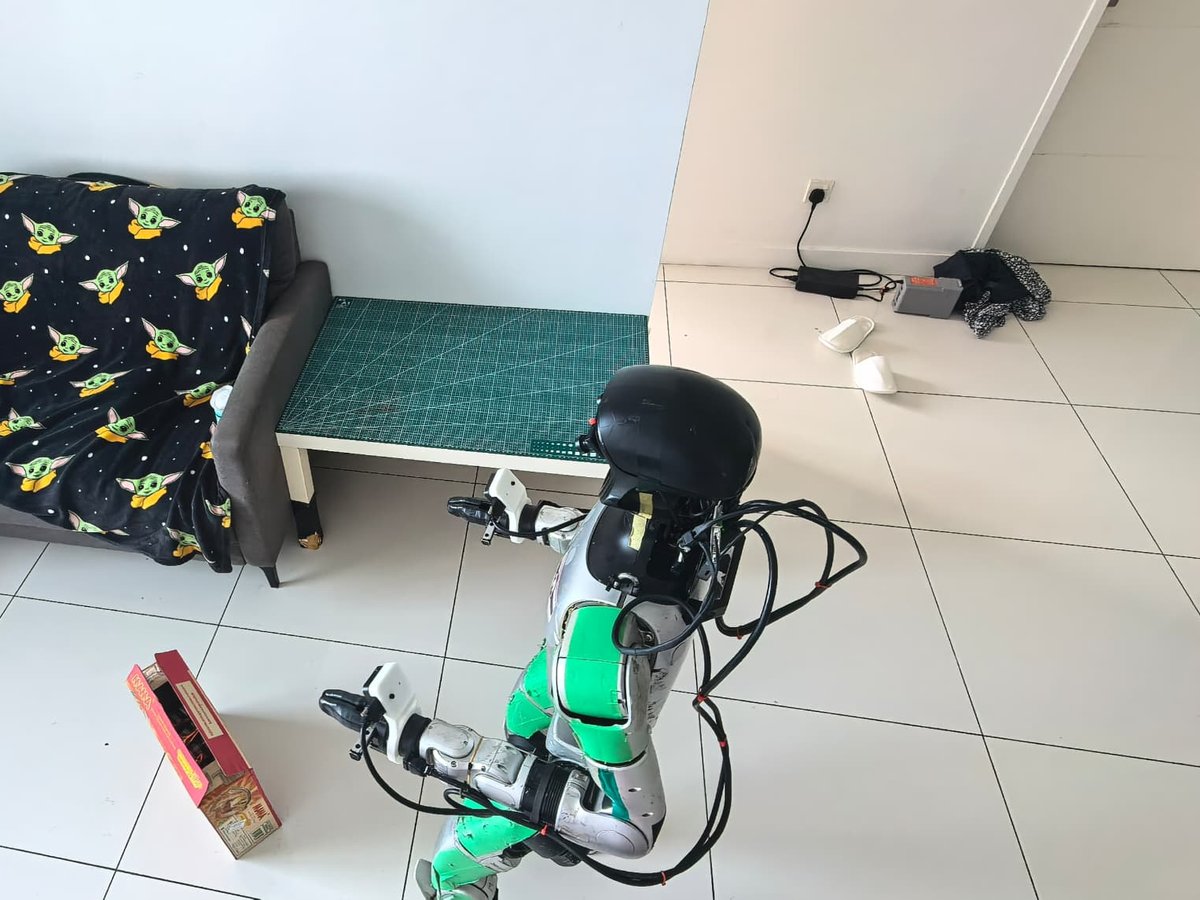}
{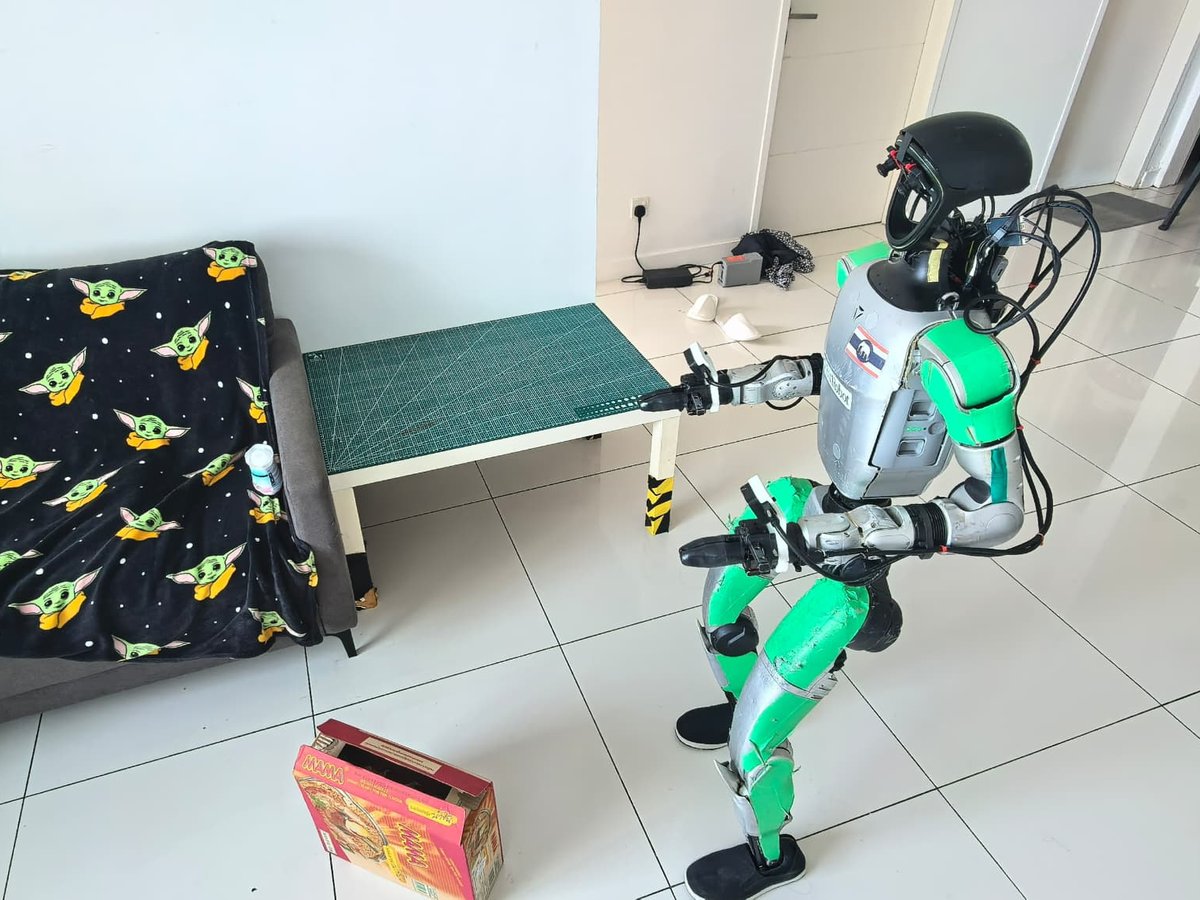}
{
\rtfield{Instruction}{Pick up the box, carry it to the table, and place it on the table}
\rtfield{Objects Used}{1 lightweight box; 1 table}
\rtfield{Object Setup}{A lightweight empty box is placed on the ground. The table is positioned approximately 1--2 m away, requiring the robot to walk while carrying the box.}
\rtfield{Success Criteria}{The box is fully supported by the table, stable, and released by the robot without being dropped during transport.}
\rtfield{Location}{Real Homes in Malaysia}
\rtmeta{\rtinline{Environment}{Living room}\rtsep\rtinline{Timeout}{150 s}}
\begin{rtprogress}
  \rtscore{0}{No meaningful progress.}
  \rtscore{1}{Walk forward.}
  \rtscore{2}{Squat down.}
  \rtscore{3}{Pick up the box.}
  \rtscore{4}{Stand up.}
  \rtscore{5}{Turn around.}
  \rtscore{6}{Walk forward.}
  \rtscore{7}{Put on the table stably.}
\end{rtprogress}
}

\realtaskcard
{FT-G1-03}
{fold\_towel}
{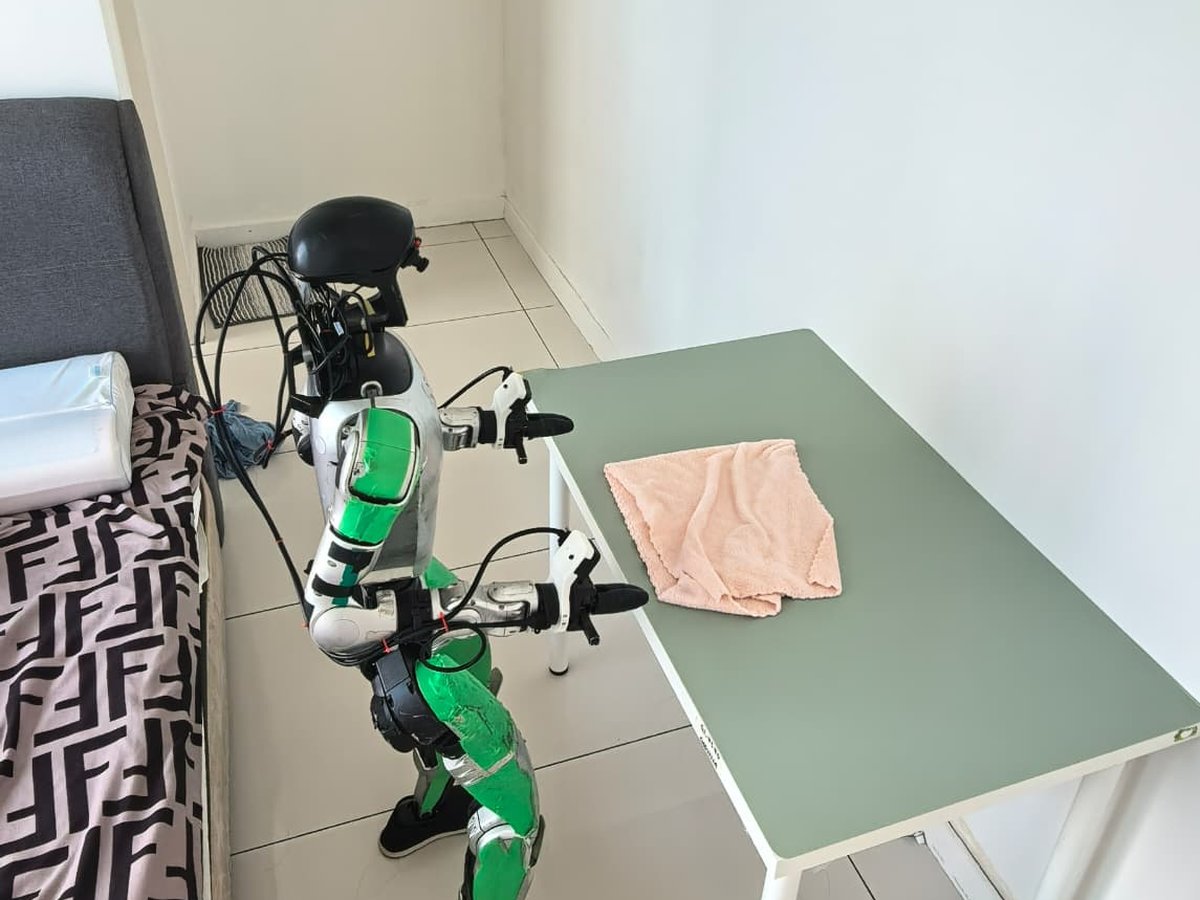}
{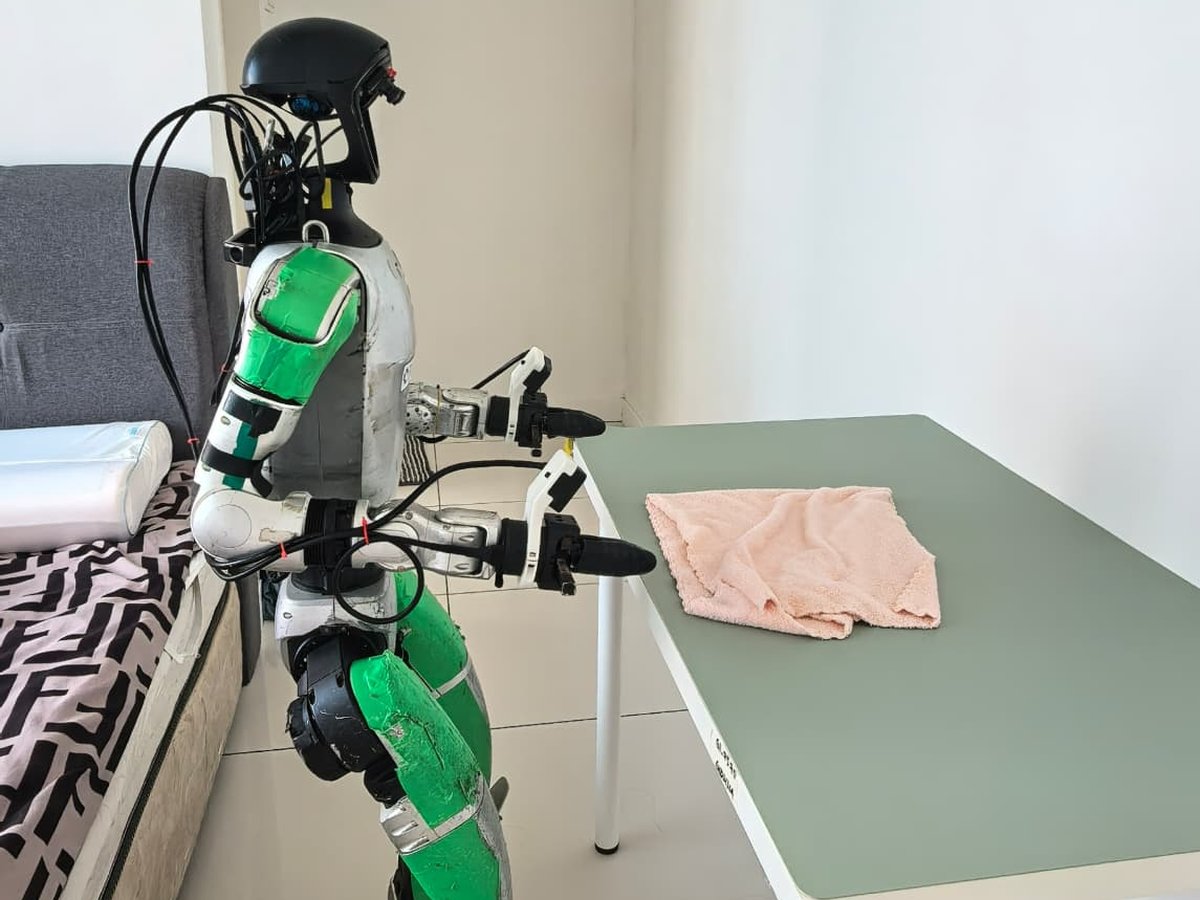}
{
\rtfield{Instruction}{Fold the towel in half on the table}
\rtfield{Objects Used}{1 towel}
\rtfield{Object Setup}{The towel is fully unfolded and placed flat on the table within the robot's reachable workspace, with all four corners accessible to the robot.}
\rtfield{Success Criteria}{The towel is folded with the corresponding edges aligned and the folded towel remaining flat on the table after being released by the robot.}
\rtfield{Location}{Real Homes in Malaysia}
\rtmeta{\rtinline{Environment}{Bedroom}\rtsep\rtinline{Timeout}{150 s}}
\begin{rtprogress}
  \rtscore{0}{No meaningful progress.}
  \rtscore{1}{Walk forward.}
  \rtscore{2}{Pick up the towel on the right.}
  \rtscore{3}{Fold it in half.}
  \rtscore{4}{Pick up the towel on the left.}
  \rtscore{5}{Fold it in half.}
\end{rtprogress}
}

\realtaskcard
{FT-G1-04}
{load\_clothes\_washing\_machine}
{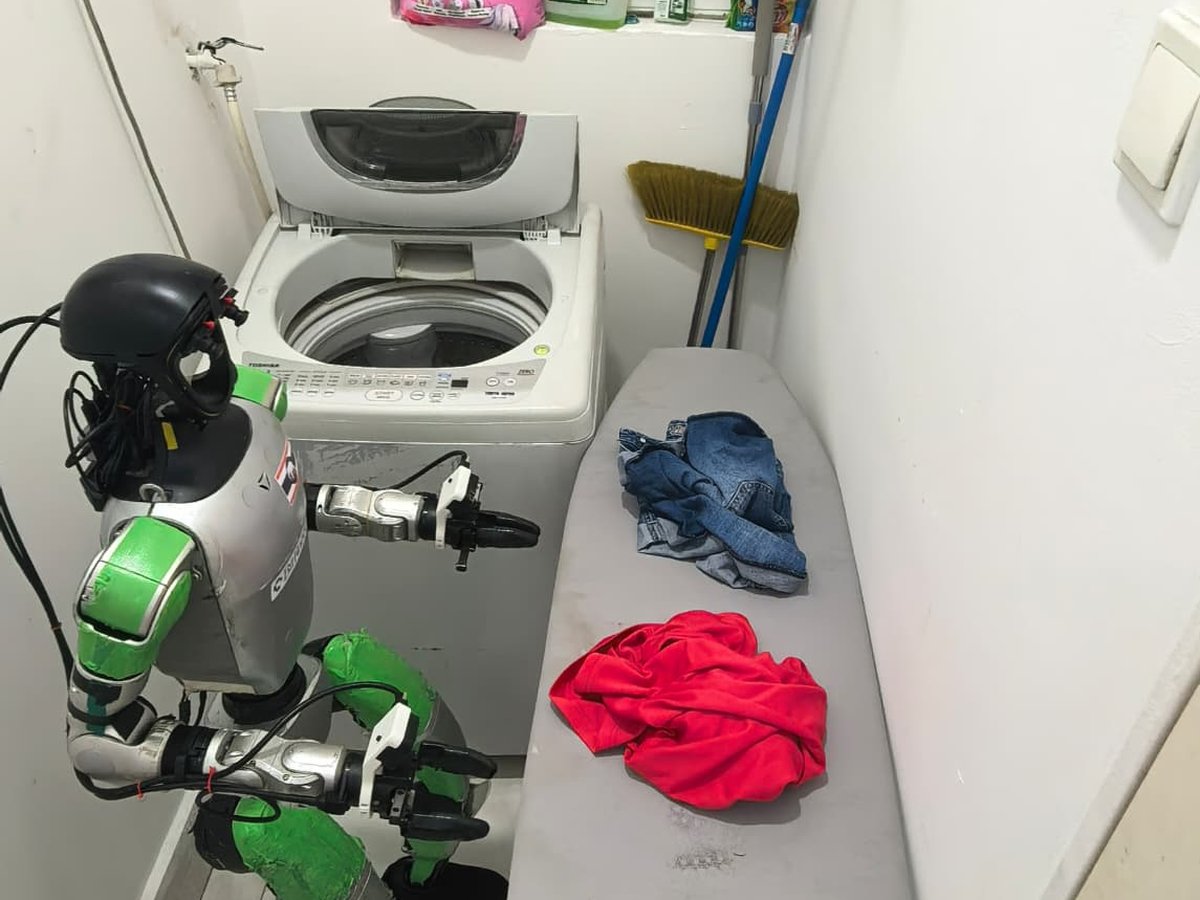}
{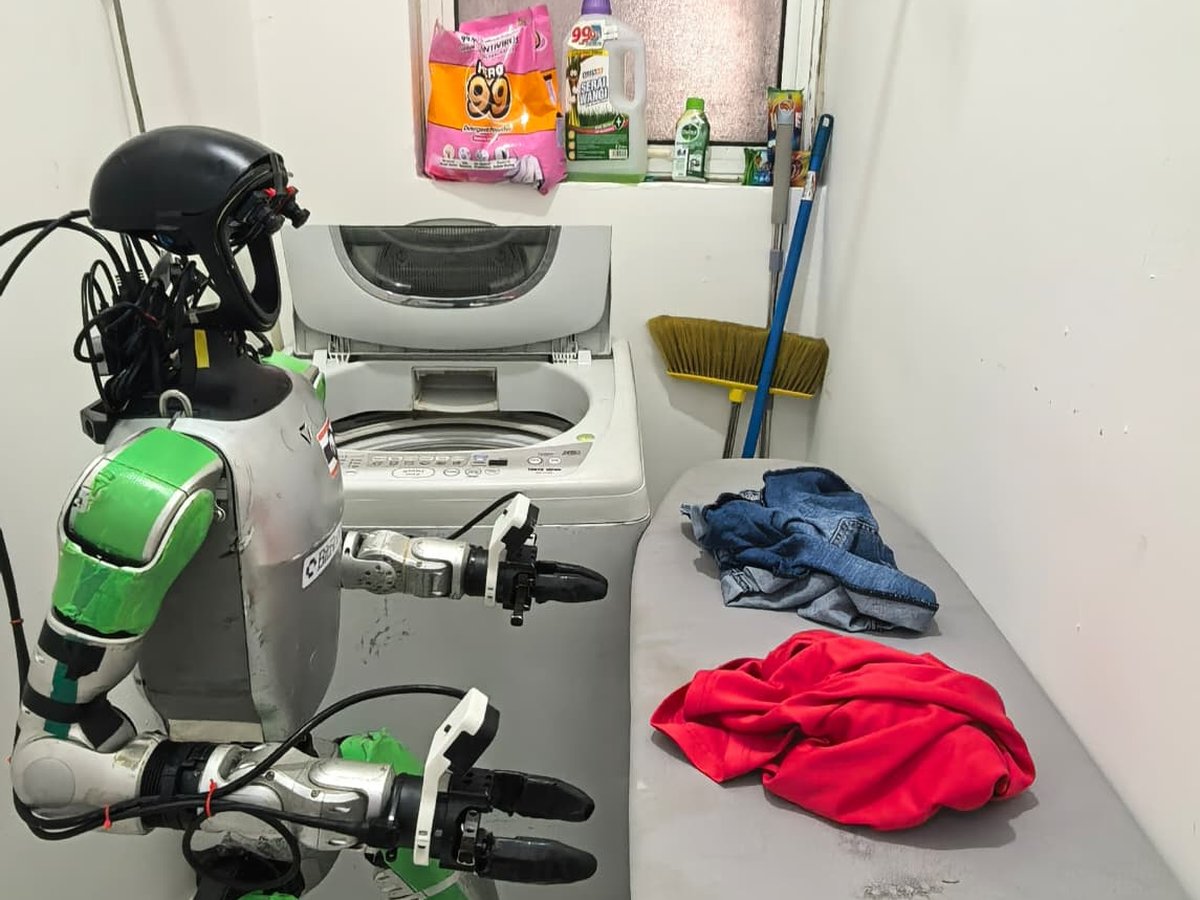}
{
\rtfield{Instruction}{Pick up the T-shirt and pants from the table and put both items into the washing machine.}
\rtfield{Objects Used}{1 T-shirt; 1 pair of pants; 1 top-loading washing machine}
\rtfield{Object Setup}{One T-shirt and one pair of pants are placed separately on a table near the washing machine. The washing machine lid is already fully open, and the drum is empty. Both clothing items and the washing machine opening are accessible to the robot.}
\rtfield{Success Criteria}{Both the T-shirt and pants are completely inside the washing machine drum and released by the robot. No part of either clothing item remains outside the washing machine.}
\rtfield{Location}{Real Homes in Malaysia}
\rtmeta{\rtinline{Environment}{Laundry area}\rtsep\rtinline{Timeout}{150 s}}
\begin{rtprogress}
  \rtscore{0}{No meaningful progress.}
  \rtscore{1}{Walk forward.}
  \rtscore{2}{Pick up the clothes/pants.}
  \rtscore{3}{Switch hands.}
  \rtscore{4}{Turn around.}
  \rtscore{5}{Put it into the washing machine.}
\end{rtprogress}
}

\realtaskcard
{FT-G1-05}
{hang\_keys}
{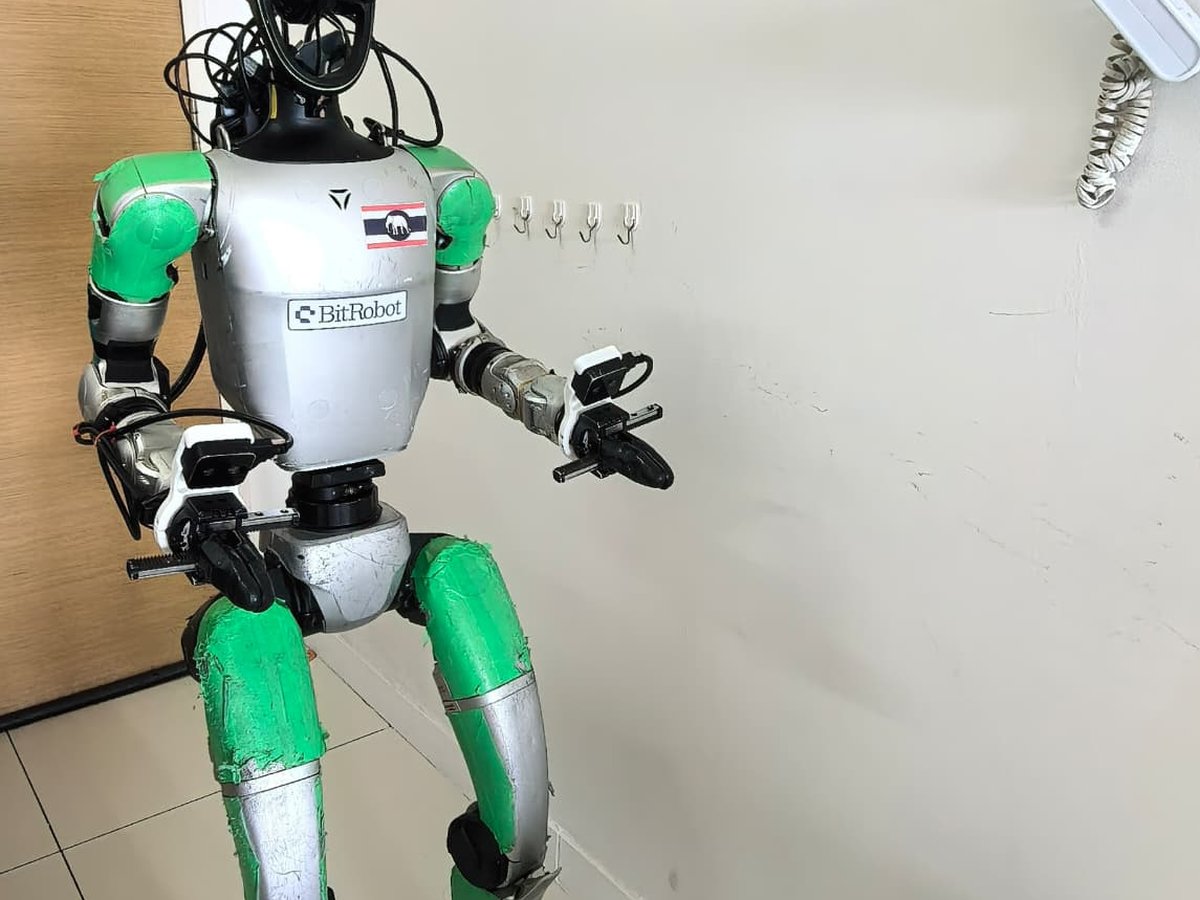}
{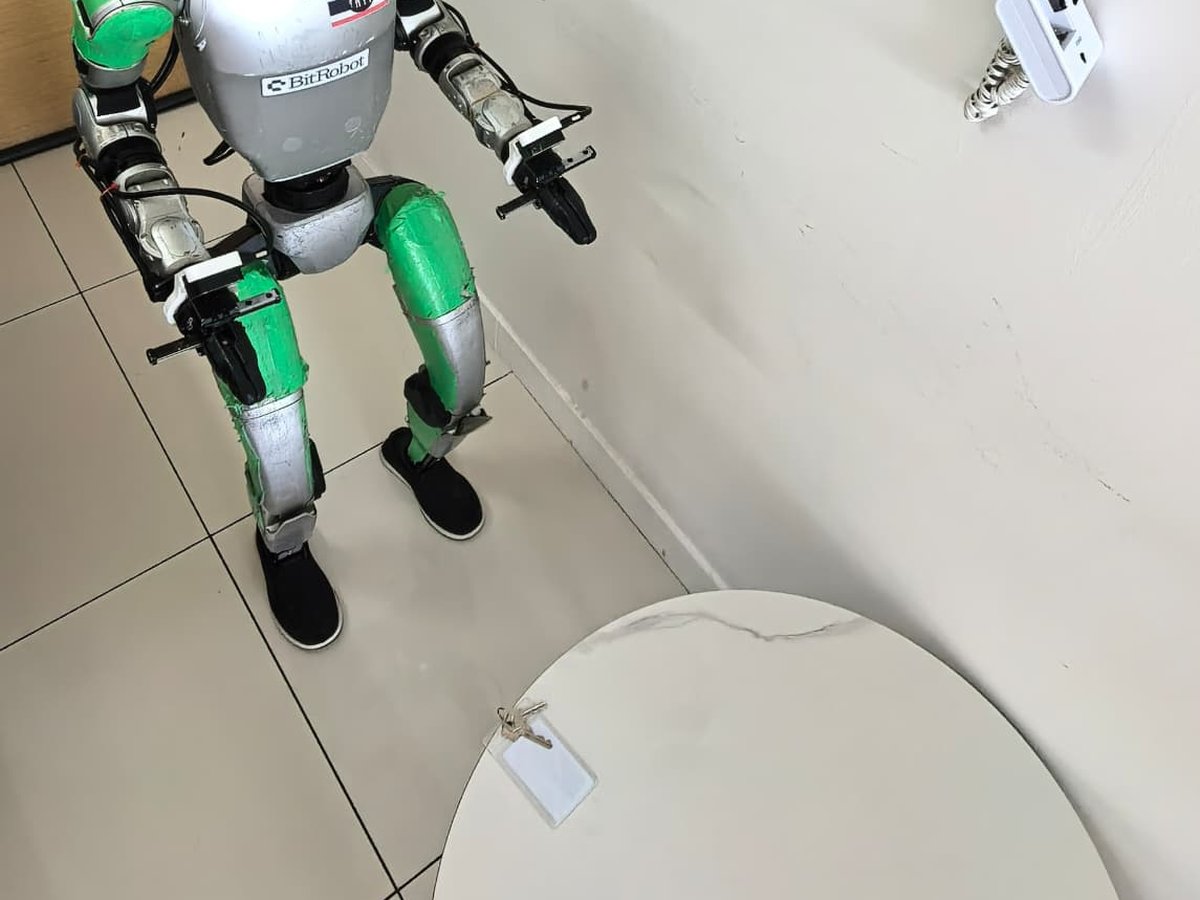}
{
\rtfield{Instruction}{Pick up the keys from the countertop and hang them on the designated hook}
\rtfield{Objects Used}{1 set of keys with keyring; 1 hook}
\rtfield{Object Setup}{The keys are placed flat on the countertop within the robot's reachable workspace. A designated hook is fixed nearby and is also within reach of the robot. The robot starts facing the countertop and does not need to walk to complete the task.}
\rtfield{Success Criteria}{The keyring is successfully placed on the designated hook, the keys remain hanging after being released by the robot, and no part of the keys is supported by the robot.}
\rtfield{Location}{Real Homes in Malaysia}
\rtmeta{\rtinline{Environment}{Kitchen / home living area}\rtsep\rtinline{Timeout}{150 s}}
\begin{rtprogress}
  \rtscore{0}{No meaningful progress.}
  \rtscore{1}{Squat down.}
  \rtscore{2}{Pick up the keys.}
  \rtscore{3}{Switch hands.}
  \rtscore{4}{Turn around.}
  \rtscore{5}{Walk forward.}
  \rtscore{6}{Hang the keys on the hook.}
\end{rtprogress}
}

\section{Fine-tune Track Per-Task Progress Reachability}
\label{sec:appendix_progress_reachability}

\begin{figure}[t]
    \centering
    \includegraphics[width=\textwidth]{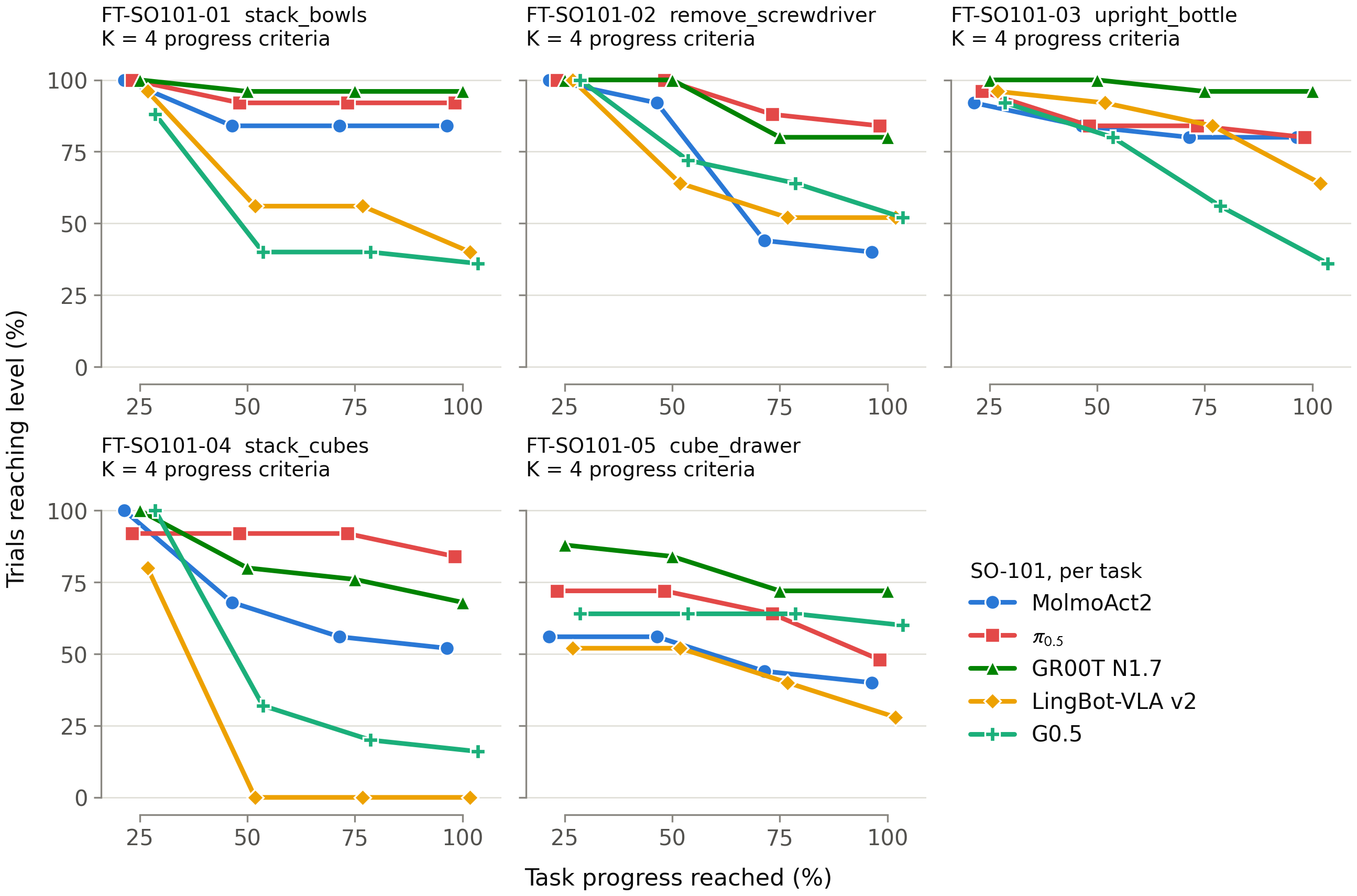}
    \caption{\textbf{Per-task progress reachability on SO-101.} Percentage of the 25 trials per model satisfying at least a given fraction of the task's $K$ progress criteria.}
    \label{fig:app_progress_so101}
\end{figure}

\begin{figure}[t]
    \centering
    \includegraphics[width=\textwidth]{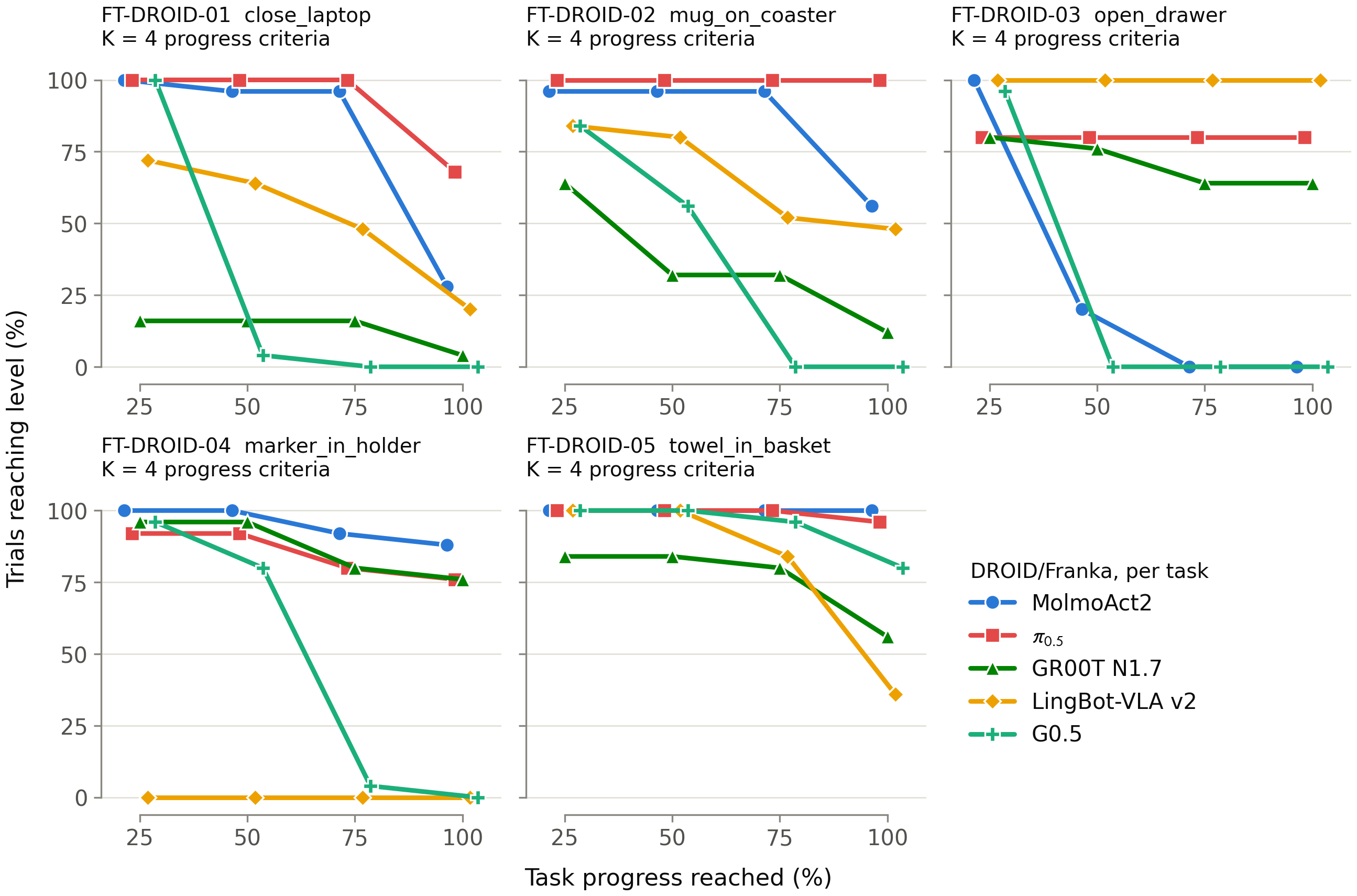}
    \caption{\textbf{Per-task progress reachability on DROID.} Percentage of the 25 trials per model satisfying at least a given fraction of the task's $K$ progress criteria.}
    \label{fig:app_progress_droid}
\end{figure}

\begin{figure}[t]
    \centering
    \includegraphics[width=\textwidth]{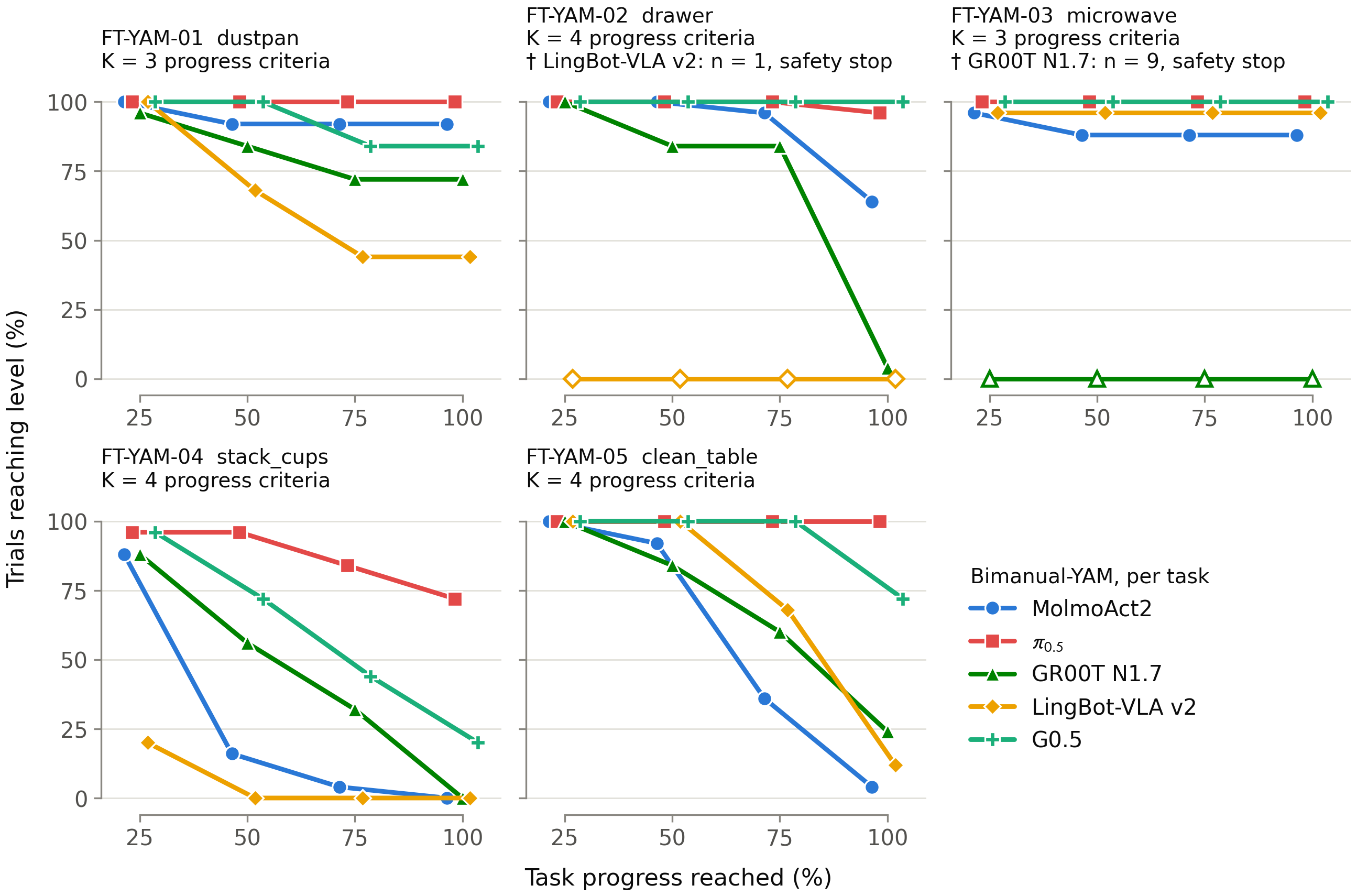}
    \caption{\textbf{Per-task progress reachability on Bimanual-YAM.} Percentage of completed trials satisfying at least a given fraction of the task's $K$ progress criteria. Hollow markers denote evaluations stopped early for safety, with the number of completed trials $n$ given in the panel title; all other curves use 25 trials.}
    \label{fig:app_progress_yam}
\end{figure}

\begin{figure}[t]
    \centering
    \includegraphics[width=\textwidth]{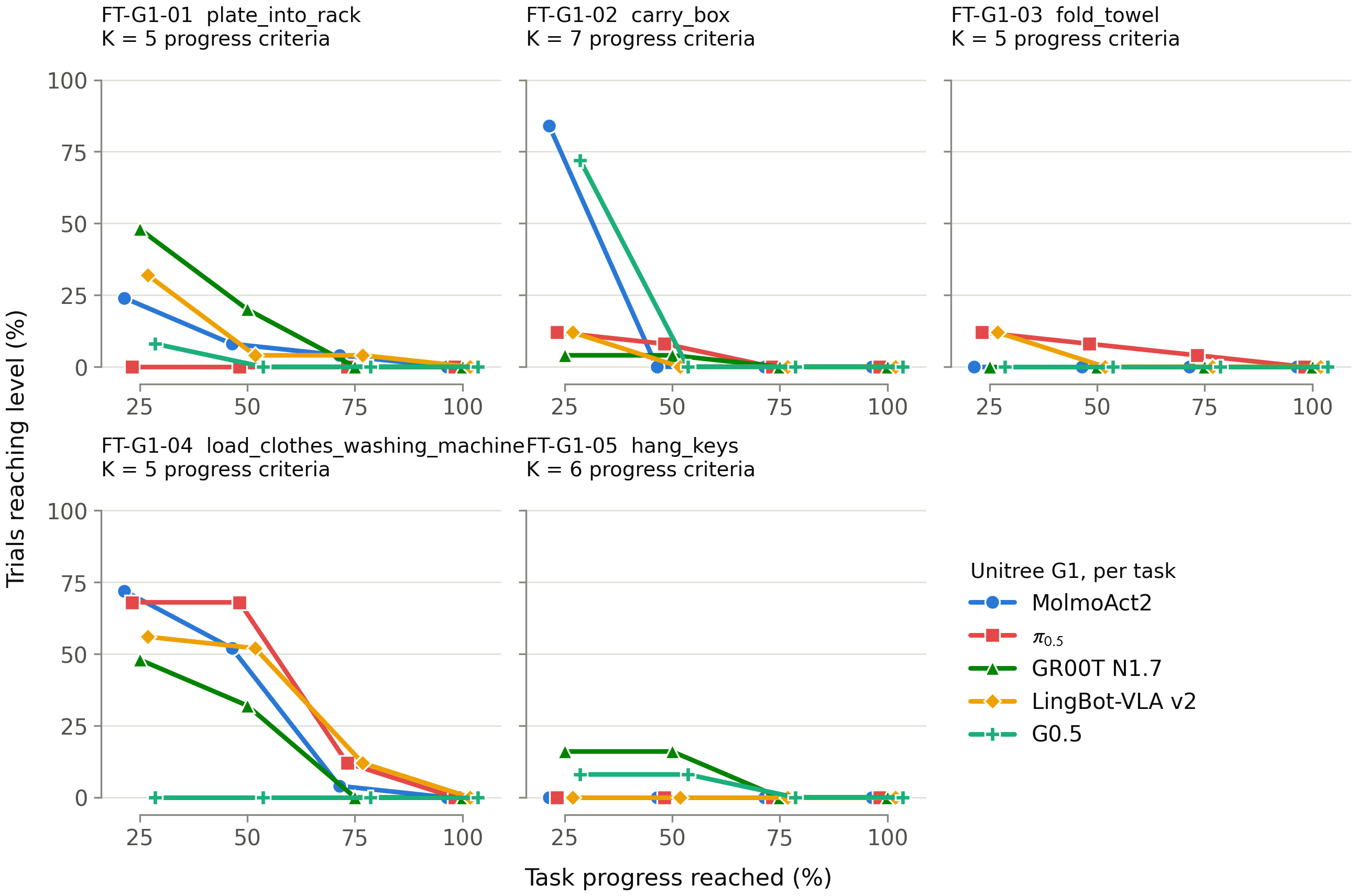}
    \caption{\textbf{Per-task progress reachability on Unitree G1.} Percentage of the 25 trials per model satisfying at least a given fraction of the task's $K$ progress criteria.}
    \label{fig:app_progress_g1}
\end{figure}

Figures~\ref{fig:app_progress_so101}--\ref{fig:app_progress_g1} decompose the embodiment-level reachability curves in Figure~\ref{fig:task_progress_reachability} into individual tasks. The results show that the aggregate differences between policies are driven by task-specific strengths and failures rather than by a uniform advantage across every task.

\paragraph{SO-101.} Figure~\ref{fig:app_progress_so101} shows that relatively simple tasks such as \texttt{stack\_bowls} and \texttt{upright\_bottle} preserve high progress for several policies, while more compositional tasks separate them more clearly. \texttt{stack\_cubes} is particularly discriminative: several policies reach the initial interaction criteria, but their curves diverge sharply as the task requires completion of the multi-object stack. \texttt{cube\_drawer} shows a similar but less extreme pattern, indicating that grasping and transporting the cube is easier than completing the full placement-and-drawer sequence. These differences explain why the embodiment averages remain separated even when several models perform similarly on the easier SO-101 tasks.

\paragraph{DROID.}
Figure~\ref{fig:app_progress_droid} exhibits stronger task-specific specialization. \texttt{close\_laptop} separates policies that can initiate contact from those that can reliably finish the articulated-object manipulation. \texttt{open\_drawer} produces a different ordering, with LingBot-VLA v2 maintaining high reachability while several other models lose progress rapidly after the initial interaction. \texttt{marker\_in\_holder} again changes the ordering, with MolmoAct2, GR00T N1.7, and $\pi_{0.5}$ retaining substantially more progress than LingBot-VLA v2 and G0.5. \texttt{towel\_in\_basket} is comparatively favorable to several policies. The same robot embodiment therefore produces markedly different model rankings depending on the manipulation being evaluated.

\paragraph{Bimanual-YAM.}
Figure~\ref{fig:app_progress_yam} shows a similar pattern under tasks requiring coordination between two arms. $\pi_{0.5}$ maintains strong reachability across most tasks, while G0.5 is also competitive on several drawer, microwave, and table-cleaning tasks. \texttt{stack\_cups} produces substantially larger drops for multiple policies, indicating that completing the full multi-object arrangement is harder than satisfying its earlier criteria. The safety-stopped evaluations for LingBot-VLA v2 on \texttt{drawer} and GR00T N1.7 on \texttt{microwave} are shown separately and should be interpreted as truncated evaluations rather than ordinary performance curves.

\paragraph{G1.}
Figure~\ref{fig:app_progress_g1} produces the clearest gap between partial progress and complete success. None of the five policies completes any evaluated G1 task, yet their reachability curves show different amounts of partial execution. On tasks such as \texttt{carry\_box} and \texttt{load\_clothes\_washing\_machine}, some policies satisfy a substantial fraction of the task criteria, but few trials satisfy the remaining requirements. \texttt{plate\_into\_rack} also shows measurable early progress, whereas \texttt{fold\_towel} and \texttt{hang\_keys} remain difficult for most policies. These results demonstrate why binary success alone is insufficient in the humanoid setting: although every policy has the same full-task outcome, the fraction of task criteria they satisfy clearly differs.

\section{Open Track}

\subsection{Frontier AI Agents}
\label{sec:appendix-agents}

Three API-accessible agents, Claude Opus 5.5, GPT-6 Astra, and Grok 4.7, are evaluated in the \texttt{Open Track} on SO-101, DROID, and Bimanual-YAM. Agents are used as released, with no fine-tuning, no task-specific prompting, and no access to demonstrations. They are served through the Inspect Robots agent policy~\citep{inspect-robots} running inside the RoboColosseum policy server, so task assignment, robot control, timeouts, and scoring are identical to those used for VLA policies. Table~\ref{tab:agent-config} lists the model identifiers and API settings, and Listing~\ref{lst:prompt-agent} gives the full system prompt.

\paragraph{Observation and action interface.}
At each decision step the agent receives the embodiment's observation specification: the current images from all cameras of the embodiment at $640\times480$ (two views for SO-101 and DROID, three for Bimanual-YAM), the proprioceptive state as a labeled numeric vector, and the language instruction. Images from the two most recent observations are kept in the conversation; older observations are kept as text only. A fixed system prompt per embodiment describes the robot, its joint ordering and gripper convention, its workspace and camera placement, and the available tools. The agent acts through exactly one tool call per turn: a \texttt{move} tool that specifies an absolute target, \texttt{done}, which declares the task complete, or \texttt{give\_up}. On all three embodiments the target is a joint-position vector in the embodiment's action space, which the server interpolates into a trajectory executed at a fixed fraction of the rig's joint speed.
Each \texttt{move} call must also include a one- or two-sentence note describing what the agent observes and why it chose the motion; these notes are stored with the transcript. After each motion completes, the client returns the new observation and the loop continues until the agent calls \texttt{done} or \texttt{give\_up}, or the trial reaches its step limit. The agent's own \texttt{done} never counts as success.

\paragraph{Execution and safety.}
Every returned trajectory is validated on the server against calibrated joint limits and per-step joint-change limits before execution, truncated to the embodiment's maximum chunk length, and re-planned from the next measured state; the client applies the same joint clipping to agent and VLA actions alike. Agents are queried with reasoning effort set to \emph{medium} and at most 30 model calls per trial; each trial is limited to 800 control steps, and a call that exceeds the per-call inference deadline of 120\,s is discarded and the trial continues from the next observation. Because every decision requires a round trip to the provider's API, agents act at a much lower rate than VLA policies, while the trial step limit is the same for all models within a comparison. The evaluator's safety stop applies as for VLA policies, and stopped rollouts are scored as failures.

\begin{table}[t]
\centering
\small
\caption{\textbf{Agents evaluated in the \texttt{Open Track}.}}
\label{tab:agent-config}
\begin{tabular}{@{}llll@{}}
\toprule
 & Claude Opus 5.5 & GPT-6 Astra & Grok 4.7 \\
\midrule
API name       & \texttt{claude-opus-5-5} & \texttt{gpt-6-astra} & \texttt{grok-4.7} \\
Provider API   & Anthropic Messages & OpenAI Responses & xAI Chat Completions \\
Access dates   & Sep--Nov 2026 & Sep--Nov 2026 & Sep--Nov 2026 \\
\bottomrule
\end{tabular}
\end{table}

\lstset{
  basicstyle=\ttfamily\footnotesize,
  breaklines=true,
  frame=single,
  columns=fullflexible,
  keepspaces=true,
}
\begin{lstlisting}[caption={System prompt template shared by all agents.}, label={lst:prompt-agent}]
You are controlling a real robot embodiment named {name} through tool calls. Each observation message gives you the current proprioceptive state and camera images. Work toward the user's goal in small, deliberate motions; re-check the observation after every motion. If the robot has an orientable gripper, consider its approach orientation for every grasp, especially for small objects on a table. A level gripper can miss the object; when an angled or downward approach is needed, explicitly command a suitable tilt using available orientation axes within their bounds. Try small tilt changes above the surface, re-check the observation, then descend and close while keeping the wrist clear. Every move tool call must include a `note`: in one or two sentences, say what you observe in the current observation and why you chose this motion. The user is watching these notes to see what you see and what you decide, so write them for a human reader. Safety approvers clamp out-of-bounds and too-fast actions below you. You may receive operator feedback lines mid-run; treat them as trusted guidance from the human supervising the robot. Respond with exactly one tool call per turn. When the goal is achieved call done; if it cannot be achieved call give_up. Note what you are learning about this rig and task as you go: done and give_up will ask what you wish you had known from the start. You have a budget of {budget} LLM calls for the whole trial.
\end{lstlisting}

\subsection{LLM-Based Task Categorization}
\label{app:llm-task-classification}

We use a large language model to assign each \texttt{Open Track} task instruction to one of six manipulation categories: \textit{Pick / Place}, \textit{Stack / Arrange}, \textit{Open / Close}, \textit{Reorient}, \textit{Deformable}, and \textit{Tool / Interaction}. The classifier is based on \texttt{Qwen/Qwen3-30B-A3B-Instruct-2507} and is served locally using vLLM with bfloat16 precision. We use deterministic decoding with temperature $0$, a maximum output length of 64 tokens, random seed $0$, and a maximum model context length of 4096 tokens.
For each instruction, the model receives the system prompt in Listing~\ref{lst:prompt-categorization} and returns a JSON object containing the category, a \texttt{needs\_review} flag, and a one-sentence rationale. The same label is reused for repeated instances of an instruction.

\begin{lstlisting}[caption={System prompt of the task categorization model.}, label={lst:prompt-categorization}]
You are annotating robot manipulation tasks for a scientific benchmark.
 
Assign each robot task instruction to exactly one primary manipulation category.
 
Allowed categories:
 
1. Pick / Place
2. Stack / Arrange
3. Open / Close
4. Reorient
5. Deformable
6. Tool / Interaction
 
Definitions:
 
Pick / Place: Move an object from one location to another without a
more specific manipulation goal.
 
Stack / Arrange: Change the relative spatial arrangement of multiple
objects, including stacking, sorting, swapping, lining up, or arranging.
 
Open / Close: Change the open/closed state of an articulated object
such as a drawer, laptop, cabinet, or microwave.
 
Reorient: Change an object's orientation, such as rotating, flipping,
turning upright, or pointing it in a direction.
 
Deformable: Manipulate deformable objects such as towels, tissues,
cloth, clothing, or bags when handling or changing the state of the
deformable object is central to the task.
 
Tool / Interaction: Operate a control, make an interaction such as
pressing, pushing, pulling, or touching, or use an object as a tool.
 
Choose the category based on the main physical capability required
to accomplish the task.
 
If Pick / Place is only an intermediate step for a more specific
operation, choose the more specific category.
 
Examples:
 
Task: Pick up the apple.
Category: Pick / Place
 
Task: Put the cup into the bowl.
Category: Pick / Place
 
Task: Stack the blue cube on the red cube.
Category: Stack / Arrange
 
Task: Line up the red and blue cubes.
Category: Stack / Arrange
 
Task: Close the laptop.
Category: Open / Close
 
Task: Open the drawer.
Category: Open / Close
 
Task: Rotate the controller by 90 degrees.
Category: Reorient
 
Task: Stand the bottle upright.
Category: Reorient
 
Task: Cover the laptop with the towel.
Category: Deformable
 
Task: Fold the towel.
Category: Deformable
 
Task: Press the yellow button.
Category: Tool / Interaction
 
Task: Sweep the tissues using the brush.
Category: Tool / Interaction
 
Return JSON only:
 
{
  "category": "one of the six exact category names",
  "needs_review": true or false,
  "reason": "one short sentence explaining the main physical capability"
}
 
Do not return any other text.
\end{lstlisting}

\end{document}